%% file: main.tex
\documentclass{article}
\usepackage[utf8]{inputenc}
\usepackage{authblk}

\title{An Empirical Study of Generative Models with Encoders}

\author[$\sharp\flat$]{Paul K. Rubenstein\thanks{Work was done while at Google AI.}}
\author[$\natural$]{Yunpeng Li}
\author[$\natural$]{Dominik Roblek}
\affil[$\sharp$]{Max Planck Institute for Intelligent Systems, T\"ubingen}
\affil[$\flat$]{Machine Learning Group, University of Cambridge}
\affil[$\natural$]{Google AI}
\usepackage{natbib}
\usepackage{rotating}
\usepackage{afterpage}
\usepackage{subcaption}
\usepackage{enumitem}
\usepackage{xcolor}
\usepackage{hyperref}
\usepackage{bbold}
\usepackage{bbm}

\usepackage{amsthm}
\newtheorem{proposition}{Proposition}
\newtheorem{theorem}{Theorem}

\usepackage{graphicx}
\usepackage{amsfonts}
\usepackage{amsmath}
\newcommand{\Z}{\mathcal{Z}}

\newcommand{\X}{\mathcal{X}}
\newcommand{\E}{\mathbbm{E}}

\newcommand{\comment}[1]{}
\newcommand{\resolved}[1]{}

\begin{document}

\maketitle

\begin{abstract}
Generative adversarial networks (GANs) are capable of producing high quality image samples.
However, unlike variational autoencoders (VAEs), GANs lack encoders that provide the inverse mapping for the generators, i.e., encode images back to the latent space.
In this work, we consider adversarially learned generative models that also have encoders.
We evaluate models based on their ability to produce high quality samples and reconstructions of real images. 
Our main contributions are twofold:
First, we find that the baseline \emph{Bidirectional GAN} (BiGAN) can be improved upon with the addition of an autoencoder loss, at the expense of an extra hyper-parameter to tune. 
Second, we show that comparable performance to BiGAN can be obtained by simply training an encoder to invert the generator of a normal GAN.
\end{abstract}

\section{Introduction}

\input{introduction}

\section{Background and Related Work}\label{section:background}

\input{background}

\section{Models}\label{section:model-zoo}

\input{model_zoo}

\section{Experiments}\label{sec:experiments}

\input{experiments}

\section{Conclusion}\label{sec:conclusion}

\input{conclusion}

\section*{Acknowledgements}
The authors are grateful to Karol Kurach and Sylvain Gelly for stimulating discussions, and to Jeff Donahue for help with an initial implementation of BiGAN.

\bibliography{references}

\appendix

\newpage

\section{Appendix}

\input{appendix}

{
\section{Qualitative Results}\label{appendix:qualitative-results}

\input{qualitative}

}

\end{document}

%% file: introduction.tex
The fundamental problem of unsupervised generative modelling is to learn the distribution of real data. 
More specifically, given a finite number of samples from a (typically high dimensional) distribution $P_X$, the goal is to learn a distribution $P_G^\theta$ with parameter $\theta$ that minimises a divergence $D(P_X, P_G^\theta)$.
The two most popular approaches currently dominating this field are Variational Auto-Encoders (VAE) \citep{kingma2013auto} and those based on Generative Adversarial Network (GAN) style adversarial training \citep{goodfellow2014generative}. 

Both approaches are instances of latent variable models, in which $P_G^\theta$ is implicitly defined by specifying a fixed \emph{prior} distribution $P_Z$ over a low-dimensional space $\Z$ and parameterising the generator $G$, a conditional distribution $P^\theta_G(X|Z)$ which may or may not be deterministic. Since most interesting divergences $D(P_X, P_G^\theta)$ are intractable to evaluate, the desired divergence is typically substituted with a tractable estimate or bound.

The derivations of the objective functions for autoencoder (AE) models---including the popular VAE and recently introduced Wasserstein AE \citep{tolstikhin2017wasserstein}---involve the introduction of an encoder $E: \X \to \Z$ mapping in the reverse direction to the generator. The fact that the such models come with encoders `for free' in spite of fundamentally being trained to minimise a divergence in $\X$-space is an oft-cited advantage of these models, since the encoder can be used as a feature extractor for downstream tasks.

In contrast, GAN models in their basic form work by simultaneously training a \emph{discriminator} to estimate the divergence $D(P_X, P_G^\theta)$ and a generator to minimise the estimated divergence. The discriminator can be interpreted as a classifier whose objective is to distinguish between genuine samples from $P_X$ and `fake' samples from the model distribution $P_G^\theta$. 
Although GANs suffer from many problems---unstable training and sensitivity to hyper-parameters due to minimax optimisation; mode collapse; lack of encoders---when trained successfully the synthetic images that they generate are typically far superior to those of AE models. 

The differing strengths and weaknesses of these model types have motivated several authors to investigate architectures that blend the best of both worlds. One line of research has involved trying to improve the sample quality of AE models by introducing some aspect of adversarial training, e.g. \citep{makhzani2015adversarial, larsen2016autoencoding}. Another line of research aims to train GAN-like generative models that are also equipped with encoders, e.g. \citep{donahue2016adversarial, dumoulin2016adversarially, ulyanov2017takes}.

In this work we compare several generative models with encoders based on the quality of the samples that they produce as well as the quality of the reconstructions they make of real images. 
While there are many ways to define the quality of an encoder, for instance with reference to some downstream task such as in a semi-supervised setting, we chose to focus on image reconstruction quality since this is the most generic criterion when modelling images.
In this setting, we are interested both in producing random samples that \emph{look good} and are \emph{diverse} (i.e. $D(P_X, P_G^\theta)$ is small) and that $G(E(X)) \approx X$ for $X \sim P_X$. Our contributions and conclusions are summarised as follows.

\begin{itemize}
    \item We consider four losses that can be used to train encoders to invert the generator, including two novel adversarial losses. These are losses that can be `dropped in' to models that already have encoders to additionally encourage inversion, or used to train encoders in models that do not have them already.
    \item We empirically test each of these losses by considering their combination with both the (usually encoder-less) GAN and the \emph{Bi-directional GAN} (BiGAN) \citep{donahue2016adversarial, dumoulin2016adversarially}, testing on several datasets with extensive hyper-parameter search.
    \item We find consistently that the reconstructive performance of BiGAN can be improved upon, but there was no single model that dominated all others in all metrics.
    \item Very naive methods to take a normal GAN and invert the generator to get an encoder work surprisingly well, and may be preferable in many cases due to greater understanding of how best to train normal GANs.
\end{itemize}

Our concise recommendations to practitioners are that the invertible property of BiGAN can be improved upon with the addition of an extra autoencoder reconstruction loss to the objective of the encoder (we refer to this as BiGAN + $\X$ autoencoder), at the expense of an additional hyper-parameter to be tuned. 
If computational resources are limited and the burden of additional hyper-parameter tuning is too great, then taking a normal GAN and training an encoder with one of our adversarial losses (GAN + $\Z$-adversarial) results in reconstructions that are comparable or better than BiGAN. 
Alternatively, taking a normal GAN and training an encoder with an autoencoder loss (GAN + $\X$ autoencoder) results in reconstructions that are perceptually more similar to the originals, but that look less realistic on average, than the BiGAN models. 
We will describe these additional losses in details in Section \ref{subsec:invert-losses}.
Training such models offers the additional advantage of being able to make use of research on which combinations of losses and regularisation are in practice best for training (encoder-less) GANs  \citep{lucic2017gans}. In contrast, comparable studies have not been performed for BiGAN models.

The rest of this paper is organised as follows:
In Section~\ref{section:background}, we introduce the necessary background material. In Section~\ref{section:model-zoo} we introduce the losses encouraging invertibility. In Section~\ref{sec:experiments} we describe our experimental method and evaluation pipeline, and present our results. We give concluding remarks in Section~\ref{sec:conclusion}.

\textbf{Notation:} Throughout, we will use $P_G$ and $P_G^\theta$ interchangeably, omitting the $\theta$ for notational convenience where unambiguous. $P_Z$ refers to the prior and $P_X$ the data distribution. $P_{X, E(X)}$ refers to the distribution of the $\X \times \Z$-valued random variable $(X, E(X))$ where $X \sim P_X$. $P_{Z, G(Z)}$ refers to the distribution of the $\Z \times \X$-valued random variable $(Z, G(Z))$ where $Z \sim P_Z$, and similarly for other distributions such as $P_{Z, G(E(G(Z)))}$. Capital letters ($X, Z$) refer to random variables. Script letters ($\X, \Z$) refer to the sets in which they take value, and lower-case letters ($x, z$) refer to elements of these sets.

%% file: background.tex
\subsection*{Generative Models}

Since the inception of generative adversarial networks (GAN) \citep{goodfellow2014generative}, there has been a flurry of activity within the machine learning community applying the concept of \emph{adversarial training}. 
This first GAN paper treats generative modelling as a game between two competing players.
The goal of the \emph{discriminator} $D: \X \to [0,1]$ is to distinguish between real samples drawn from $P_X$ and fake samples from the model distribution $P_G^\theta$, which is implemented as a binary classification problem.
The goal of the generator $G: \Z \to \X$ is to tune the model distribution $P_G^\theta$ so as to maximise the discriminator's loss. In practice $G$ and $D$ are implemented as deep neural networks, and we obtain the following minimax optimisation problem that can be solved using stochastic gradient methods:

\begin{align*}
    \max_G\min_D \ V(G,D) := \E_{X\sim P_X}\left[-\log(D(X)) \right] + \E_{Z \sim P_Z}\left[-\log(1 - D(G(Z))) \right].
\end{align*}

Consider any fixed generator $G$, and define $D^*_G$ to be the optimal discriminator $D$ minimising $V(G,D)$ for this fixed $G$.
\cite{goodfellow2014generative} showed that $D_G^*(x) := \frac{dP_X(x)}{dP_X(x) + dP_G^\theta(x)}$, and at this point $V(G, D_G^*) = \log(4) - 2 \cdot D_{JS}(P_X || P_G^\theta)$ where $D_{JS}$ is the \emph{Jensen-Shannon} divergence. In this sense, the discriminator estimates a divergence between the model distribution and true data distribution.

In practice, the above objective function for $G$ provides poor gradients for training purposes and so it is common to use the following objective functions with the same fixed point characteristics for training (this is often referred to as the \emph{non-saturating log-loss} in the literature):

\begin{equation}\label{eqn:gan_loss}
\begin{aligned}
    D: \quad &\min_D \ \E_{X\sim P_X}\left[-\log(D(X)) \right] + \E_{Z \sim P_Z}\left[-\log(1 - D(G(Z))) \right] \\
    G: \quad &\min_{G} \ \E_{Z \sim P_Z}\left[-\log(D(G(Z))) \right]. \\
\end{aligned}    
\end{equation}

While GANs do not come with encoders, it is possible to exploit the features learned by the discriminator. 
For instance, \cite{larsen2016autoencoding} jointly trains a GAN and a VAE using a reconstruction loss at the level of features from the discriminator rather than at the level of pixels. 
Taking a different approach, \cite{salimans2016improved} use GANs for semi-supervised classification by using the generator to generate unlabelled data and training a $(K+1)$-class discriminator to distinguish between each of the $K$ classes of true data and the single class of generated data. 
While both of these approaches exploit features learned by the discriminator, they do not provide a method to embed images into the latent space of the generator.

The recently proposed \emph{Bidirectional} GAN (BiGAN) \citep{donahue2016adversarial} sought to extend the GAN framework to include an encoder $E: \X \to \Z$ that maps in the reverse direction of the generator. The idea was also explored independently by \citet{dumoulin2016adversarially} in a concurrent paper, leading to essentially the same formulation.
In this setting, the discriminator $D: \X \times \Z \to [0,1]$ is tasked with distinguishing between pairs $(G(Z), Z)$, where $Z \sim P_Z$ is drawn from the prior, and $(X, E(X))$ where $X\sim P_X$ is drawn from the data distribution. 
This leads to the following objective:

\begin{align*}
    \max_{E,G}\min_D \ V(E,G,D) :=  \E_{X\sim P_X}\left[-\log(D(X, E(X))) \right] + \E_{Z \sim P_Z}\left[-\log(1 - D(G(Z), Z)) \right]. \\
\end{align*}

\begin{figure}
    \centering
    \includegraphics[width=0.95\columnwidth]{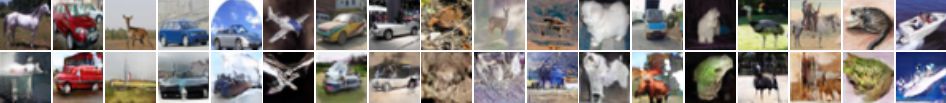}%
    
    \vspace{0.5em}
    
    \includegraphics[width=0.95\columnwidth]{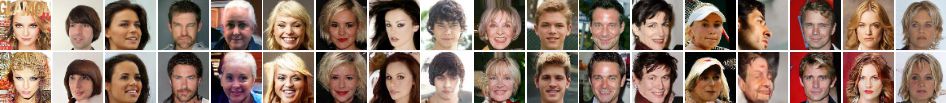}%
    
    \vspace{0.5em}
    
    \includegraphics[width=0.95\columnwidth]{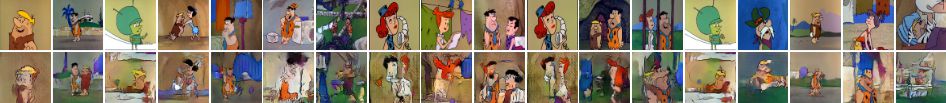}%
    
    \vspace{0.5em}
    
    \includegraphics[width=0.95\columnwidth]{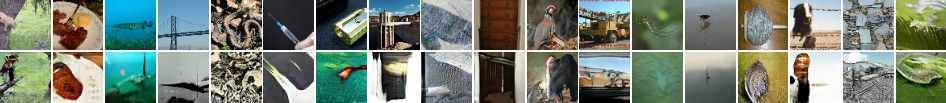}%

    \caption{Top rows: real samples from cifar10/celeba/flintstones/imagenet. Bottom rows: reconstructions from BiGAN. BiGAN characteristically produces reconstructions that are themselves plausible images, but often differ significantly from the original.}
    \label{fig:bigan_bad_reconstructions}
\end{figure}

Similar analysis to that performed by \cite{goodfellow2014generative} shows that for fixed $E$ and $G$, the loss under the optimal discriminator $D_{E,G}^*$ is $V(E, G, D_{E,G}^*) = \log(4) - 2 \cdot D_{JS}(P_{X, E(X)} || P_{G(Z),Z}^\theta)$ where $P_{X, E(X)}$ is the distribution of the $\X \times \Z$ valued random variable $(X, E(X))$ where $X \sim P_X$ and $P_{G(Z), Z}$ is the distribution of the random variable $(G(Z), Z)$ where $Z \sim P_Z$. This further implies that the global optimum occurs when $D_{JS}(P_{X, E(X)} || P_{G(Z),Z}^\theta) = 0$, which occurs when $E$ and $G$ are inverse to each other in the sense that $G\circ E = id_{\X}$ $P_X$-almost everywhere and $E \circ G = id_{\Z}$ $P_Z$-almost everywhere. In practice, this optimum may be difficult to attain, resulting in $E$ and $G$ not being inverse to one another, with BiGAN often producing reconstructions of images that look little like the original, despite often being semantically related (see Figure~\ref{fig:bigan_bad_reconstructions}).

Similarly to normal GANs, the non-saturating variant of the objective function is used in practice as it has the same fixed-point characteristics as the original but is much easier to optimise.

\begin{equation}\label{eqn:bigan_loss}
\begin{aligned}
    D: \quad &\min_D \ \E_{X\sim P_X}\left[-\log(D(X, E(X))) \right] + \E_{Z \sim P_Z}\left[-\log(1 - D(G(Z), Z)) \right] \\
    G, E: \quad &\min_{G, E} \ \E_{X\sim P_X}\left[-\log( 1 - D(X, E(X))) \right] + \E_{Z \sim P_Z}\left[-\log(D(G(Z), Z)) \right]. \\
\end{aligned}    
\end{equation}

\subsection*{Evaluating Generative Models with FID Scores}

While \cite{dumoulin2016adversarially} and \cite{donahue2016adversarial} used the encoders of their models as feature extractors to improve the sample efficiency of downstream tasks such as classification, our work was motivated by a desire to obtain good image reconstruction quality. To evaluate the quality of samples and reconstructions, we made extensive use of the \emph{Fr\'echet Inception Distance} (FID) \citep{heuselgans}. This is a distance between two probability distributions over images, based on comparing activation statistics of the top hidden layer of a pre-trained Inception (V3) network \citep{szegedy2015inception}.
The FID is computed as the Fr\'echet distance between the two corresponding distributions of Inception activations fitted to multivariate Gaussians.

More specifically, suppose that $P_{X_1}$ and $P_{X_2}$ are distributions over images and $I:\X\to \mathbb{R}^{2048}$ is a mapping taking an image to the activations of the top hidden layer. If $\mu_i$ and $\Sigma_i$ are the mean and covariance of the random variable $I(X_i)$, then the FID between $P_{X_1}$ and $P_{X_2}$ is defined to be

\begin{align}\label{eqn:fid}
    FID(P_{X_1}, P_{X_2}) = \| \mu_1 - \mu_2 \|^2 + \text{Tr}(\Sigma_1 - \Sigma_2 - 2(\Sigma_1 \Sigma_2)^{1/2}).
\end{align}

In particular, $FID(P_X, P_G)$ is a measure of how realistic model samples appear on average, with lower scores being better. Denoting by $P_{G\circ E(X)}$ the distribution of reconstructed images $G(E(X))$ where $X \sim P_X$, $FID(P_X, P_{G\circ E (X)})$ is then a measure of how \emph{realistic} reconstructed images are on average. 

The above quantities measure similarity at the level of distributions, but say nothing about how \emph{faithful} each reconstruction is to the original. 
Since the pixel-wise L2 distance between two images may not agree with human perception of similarity, we leverage the features learned by the Inception network and compute L2 distances in this feature space, giving us an Inception L2 distance. 
Averaging over the data distribution gives us an average measure of similarity between real images and their reconstructions:

\begin{align}\label{eqn:inceptionl2}
    \text{Inception L2}(P_X, P_{G\circ E(X)}) = \E_{X \sim P_X} l_2\left(I(X), I(G(E(X))) \right).
\end{align}

Note that this can be related to FID scores as follows. 
Let $W$ be the random variable taking value $w\in\X$ almost surely and denote the prbability measure of this random variable by $\delta_w$.
The mean and covariance of $I(W)$ are $I(w)$ and $0$ respectively. 
It follows from Equation \ref{eqn:fid} that $\text{FID}(\delta_{w_1}, \delta_{w_2}) = l_2(I(w_1), I(w_2))$, and therefore we can equivalently write Equation \ref{eqn:inceptionl2} as

\begin{align*}
    \text{Inception L2}(P_X, P_{G\circ E(X)}) = \E_{X \sim P_X} \text{FID}(\delta_X, \delta_{G(E(X))}).
\end{align*}

%% file: model_zoo.tex
We are interested in improving on the reconstructive performance of BiGAN. To this end, we consider modifying the BiGAN objective (\ref{eqn:bigan_loss}) with four additional losses to encourage $E$ and $G$ to be inverse. We additionally consider modifying the GAN objective (\ref{eqn:gan_loss}) with each loss in order to train encoders for models that are otherwise without an encoder. We introduce each of these losses next.

The motivation behind each of them is the observation that $E$ and $G$ being inverse is a statement about the compositions of functions $E\circ G$ and $G \circ E$. 
However, by inspecting the objective of BiGAN it is clear that neither of these compositions are ever evaluated. 
Although $E$ and $G$ trained according to the BiGAN objective should theoretically be inverse, in practice we find that this is not the case.
We therefore seek to explore losses that directly encourage invertibility, where $E$ is explicitly trained on the output of $G$ and $G$ is explicitly trained on the output of $E$.

\subsection{Losses Encouraging Invertibility}
\label{subsec:invert-losses}

\subsubsection*{$\Z$ autoencoder loss}

The simplest losses to encourage invertibility are autoencoder style reconstruction losses, for instance:

\begin{equation}\label{eqn:z_ae_loss}
    L_{\text{AE-Z}}(E,G) = \E_{Z\sim P_Z} l_2\left(Z, E(G(Z)) \right), \\
\end{equation}

where $l_2(z, z')$ where $l_2$ is the Euclidean distance.
We henceforth refer to this as the $\Z$ autoencoder loss.
This is zero if and only if $E\circ G = id_{\Z}$ $P_{Z}$-almost everywhere (a.e.), which itself implies that $G\circ E = id_{\X}$ $P_{G}$-a.e.
Note, however, that $G$ may be non-injective (as occurs when GANs exhibit mode collapsing behaviour), which would make the minimum of Equation~\ref{eqn:z_ae_loss} greater than zero.
In practice, we found that using such a loss to train $G$ resulted in blurry samples typical of using pixel-wise reconstruction losses. 
As such, we only minimise this loss with respect to the encoder.

\subsubsection*{$\X$ autoencoder loss}

We will additionally consider the following loss:

\begin{equation}\label{eqn:x_ae_loss}
\begin{aligned}
    L_{\text{AE-X}}(E,G) &= \E_{X\sim P_G} l_2\left(X, G(E(X)) \right)  \\
     &= \E_{Z\sim P_Z} l_2\left(G(Z), G(E(G(Z))) \right).  \\
\end{aligned}    
\end{equation}

The equality follows from the definition of $P_G$ as the distribution of the variable $X=G(Z)$ where $Z \sim P_Z$.
This is zero if and only if $G\circ E = id_{\X}$ $P_{G}$-a.e., and does not suffer from the possible problem of $G$ being non-injective as previously mentioned for $L_{\text{AE-Z}}$. 
While this instead suffers if $E$ is non-injective, this is less of an issue in practice if $E$ is not trained with an adversarial loss.

Note that this loss encourages small $l_2$ reconstruction error on \emph{fake} images, not real images. 
After experimenting with the analogous loss using real images, $E_{X\sim P_X} l_2\left(X, G(E(X)) \right)$, we found that training $G$ with such a loss led to the blurry images characteristic of autoencoder models. When instead using this loss to train $E$ \emph{without} $G$, we observed pathological behaviour for images $x$ not in the support of $P_G$. In this case, $E$ would typically map $x$ to extreme parts of the latent space far away from the support of the prior, and the reconstruction $G(E(x))$ would be a faded and blurry image.

\subsubsection*{Adversarial $\Z$ loss}\label{subsubsec:adv-z}

We now generalise the concept of the $\Z$ autoencoder loss to an adversarial setting, since specifying an $l_2$ loss may not be suitable. We do this by training a discriminator to distinguish between pairs $(Z, G(Z))$  and $(E(G(Z)), G(Z))$. Treating this as a problem of binary classification, we obtain the following objective:

\begin{align*}
    \max_{E,G} \min_D L_{\Z}(E, G, D) &= \E_{Z\sim P_Z}\left[-\log\left(D(G(Z), Z)\right)  - \log\left(1 - D(G(Z), E(G(Z)))\right)\right]. \\
\end{align*}

Following essentially the same reasoning as \cite{donahue2016adversarial}, we can observe the following properties of this objective (for proofs, see Appendix \ref{appendix:proofs}).

\begin{proposition}\label{prop:adv-z-optimal-disc}
For any fixed $E$ and $G$, the optimal discriminator minimising the objective $D^*_{EG} = \arg\min_D L_{\Z}(E, G, D)$ is the Radon-Nikodym derivative $f_{EG} := \frac{dP_{Z, G(Z)}}{d(P_{Z, G(Z)} + P_{E(G(Z)), G(Z)})}$.
\end{proposition}

\begin{proposition}\label{prop:adv-z-js}
The final objective with respect to $E$ and $G$, $C(E,G) := L_{\Z}(E, G, D^*_{EG})$ is a function of the Jensen-Shannon divergence $C(E,G) =  \log4 - 2 D_{\text{JS}}(P_{Z, G(Z)} || P_{E(G(Z)), G(Z)})$.
\end{proposition}

\begin{theorem}\label{thm:adv-z-e-equal-g}
For any given $G$, the optimal $E^*$ maximising $C(E,G)$ satisfies $E\circ G = id_Z$ $P_Z$-a.e.
\end{theorem}

In spite of these desirable properties, we found that using this as an additional loss to train the generator led to poor quality samples.
This is less surprising after observing that under this loss the generator is not `anchored' to real data. 
Thus we use this only to train the encoder. 
In practice, we use the non-saturating log-loss so that the actual objective (with the same fixed-point characteristics) is:

\begin{align*}
    D: \quad &\min_D \ \E_{Z\sim P_Z}\left[-\log\left(D(G(Z), Z)\right)  - \log\left(1 - D(G(Z), E(G(Z)))\right)\right] \\
    E: \quad &\min_{E} \ \E_{Z\sim P_Z}\left[- \log\left(D(G(Z), E(G(Z)))\right)\right]. \\
\end{align*}

\subsubsection*{Adversarial $\X$ loss}\label{subsubsec:adv-x}

The idea is similar to the adversarial $\Z$ loss, except with the following key difference: We are not demanding $E$ to map $G(Z)$ back to $Z$, but rather to any $Z'$ such that $G(Z') = G(Z)$. The discriminator tries distinguish between pairs $(Z, G(Z))$ and $(Z, G(E(G(Z))))$. Treating this as binary classification, we obtain

\begin{align*}
    \max_{E,G} \min_D L_{\X}(E, G, D) &= \E_{Z\sim P_Z}\left[-\log\left(D(Z, G(Z))\right)  - \log\left(1 - D(Z, G(E(G(Z))))\right)\right]. \\
\end{align*}

Similar to the adversarial $\Z$ loss, the objective has the following properties (for proofs, see Appendix \ref{appendix:proofs}).

\begin{proposition}\label{prop:adv-x-optimal-disc}
For any fixed $E$ and $G$, the optimal discriminator minimising the objective $D^*_{EG} = \arg\min_D L(D,G,E)$ is the Raydon-Nikodym derivative $f_{EG} := \frac{dP_{Z, G(Z)}}{d(P_{Z, G(Z)} + P_{Z, G(E(G(Z)))})}$.
\end{proposition}

\begin{proposition}\label{prop:adv-x-js}
Given an optimal discriminator, the remaining objective $C(E,G) := L(D^*_{EG},G,E)$ is a function of the Jensen-Shannon divergence $C(E,G) = \log4 - 2 D_{\text{JS}}(P_{Z, G(Z)} || P_{Z, G(E(G(Z)))})$.
\end{proposition}
 
\begin{theorem}\label{thm:adv-x-e-equal-g}
Given a fixed $G$, the optimal $E^*$ maximising $C(E,G)$ satisfies $G\circ E = id_{\X}$ $P_G$-almost everywhere.
\end{theorem}

For the same reason as the adversarial $\Z$ loss, we found this objective unsuitable for the generator and thus use it only for training the encoder. 
Again, we use the non-saturating log-loss in practice:

\begin{align*}
    D: \quad &\min_D \ \E_{Z\sim P_Z}\left[-\log\left(D(Z, G(Z))\right)  - \log\left(1 - D(Z, G(E(G(Z))))\right)\right] \\
    E: \quad &\min_{E} \ \E_{Z\sim P_Z}\left[ - \log\left(D(Z, G(E(G(Z))))\right)\right]. \\
\end{align*}

\subsection{Models Considered}

We experimented with adding each of the aforementioned four losses to the BiGAN and GAN losses, resulting in 8 combinations in total. 
For the two adversarial losses, this involves training a new discriminator.
In all cases, we found that modifying the loss of the generator led to poor performance---for the adversarial losses, greater instability in training; for the autoencoder losses, blurry and/or faded generator output---and consider henceforth only cases for which we modified the loss of the encoder. 

For brevity, we state here the objectives of only two of the eight models.
For the objectives of all models we considered, see Appendix \ref{appendix:models}.

Under BiGAN the encoder already has a loss, so we added an additional hyper-parameter $\lambda$ to balance this with the new part of the loss in each case.
For instance, the losses for each component for \textbf{BiGAN + adversarial $\X$} are:

\begin{align*}
    D_{1}: \quad &\min_{D_{1}} \ \E_{X\sim P_X}\left[-\log(D_{1}(X, E(X))) \right] + \E_{Z \sim P_Z}\left[-\log(1 - D_{1}(G(Z), Z)) \right] \\
    D_{2}: \quad &\min_{D_{2}} \ \E_{Z\sim P_Z}\left[-\log\left(D_{2}(Z, G(Z))\right)  - \log\left(1 - D_{2}(Z, G(E(G(Z))))\right)\right] \\
    G: \quad &\min_{G} \ \E_{Z \sim P_Z}\left[-\log(D_{1}(G(Z), Z)) \right] \\
    E: \quad &\min_{E} \ \E_{X\sim P_X}\left[-\log( 1 - D_{1}(X, E(X))) \right] + \lambda \E_{Z\sim P_Z}\left[ - \log\left(D_{2}(Z, G(E(G(Z))))\right)\right]. \\
\end{align*}

Since the normal GAN has no encoder, it was not necessary to add additional hyper-parameters when adding the losses in this case. For instance, the losses for each component for \textbf{GAN + adversarial $\Z$} are:
\begin{align*}
    D_{1}: \quad &\min_{D_{1}} \ \E_{X\sim P_X}\left[-\log(D_{1}(X)) \right] + \E_{Z \sim P_Z}\left[-\log(1 - D_{1}(G(Z))) \right] \\
    G: \quad &\min_{G} \ \E_{Z \sim P_Z}\left[-\log(D_{1}(G(Z))) \right] \\
    D_{2}: \quad &\min_{D_{2}} \ \E_{Z\sim P_Z}\left[-\log\left(D_{2}(G(Z), Z)\right)  - \log\left(1 - D_{2}(G(Z), E(G(Z)))\right)\right] \\
    E: \quad &\min_{E} \ \E_{Z\sim P_Z}\left[- \log\left(D_{2}(G(Z), E(G(Z)))\right)\right]. \\
\end{align*}

We additionally compared against BiGAN without any additional losses and a VAE.

%% file: experiments.tex
We implemented each model in such a way that they could be fairly compared, using the same architectures wherever applicable. Of course this is not always possible, since some models make fundamentally different architectural assumptions. Further, varying numbers of hyper-parameters lead to the questions of how to distribute a limited computational budget when searching over the hyper-parameter spaces and whether to allow more computational budget for those models with more hyper-parameters.
We decided to search over the same set of optimisation hyper-parameters for each model, and to perform a fixed grid search over the $\lambda$ hyper-parameter for each of the models combining BiGAN with an additional loss.

Our evaluation pipeline consisted of training each model several times with different hyper-parameters on $4$ datasets (see Table \ref{table:datasets}). 
Each single run was trained with a batch size of $64$ for $500$K training steps. Every $20$K training steps, a checkpoint was saved and the following quantities were calculated:

\begin{itemize}
    \item $FID(P_X, P_G)$: the average quality of generated images.
    \item $FID(P_X, P_{G(E(X))})$: the average quality of reconstructed images.
    \item Inception $L_2(P_X, P_{G(E(X))})$: the average perceptual faithfulness of reconstructed images. 
\end{itemize}

The architectures used for all components were based on those used by \cite{miyato2018spectral}, which we refer to as the SN-DCGAN architectures. 
All encoders were constructed by taking the SN-DCGAN discriminator architecture and replacing the last output layer with a mapping to $\Z$ instead of $\mathbbm{R}$. 
For cases that the discriminator is defined on $\Z \times \X$ (instead of $\X$) we fed $Z$ into each convolutional layer by convolving with a $1\times 1$ kernel after spatial replication and adding the result as a (spatially uniform) pre-activation bias. 
In cases that the model has two discriminators both defined on $\Z \times \X$, we shared all parameters except the final layer.
Further details of architectures, optimisation and hyper-parameter search can be found in Appendix \ref{appendix:architectures}.

\begin{table}
\centering
\begin{tabular}{c|c|c|c}
    Dataset & Resolution & Latent dim. &  Pre-processing\\ \hline
    CIFAR-10 \citep{krizhevsky2009learning} & $32\times 32 \times 3$ & $64$ &  None \\ 
    CelebA \citep{liu2015faceattributes} & $64\times 64 \times 3$ & $128$ &  Centre-cropped, area downsampled \\ 
    ImageNet \citep{deng2009imagenet} & $64\times 64 \times 3$ & $128$ &  Area downsampled \\ 
    Flintstones \citep{gupta2018imagine} & $64\times 64 \times 3$ & $256$ &  Area downsampled \\ 
\end{tabular}
\caption{\label{table:datasets}Datasets used in our evaluation.}
\end{table}

\subsection{Evaluation}\label{subsection:evaluation}

\begin{figure}
    \centering
    \begin{minipage}{0.03\columnwidth}
        \begin{turn}{90}
                Imagenet \hspace{10em} Flintstones \hspace{10em} CelebA \hspace{10em} Cifar10
        \end{turn}
    \end{minipage}%
    \begin{minipage}{0.97\columnwidth}
    \includegraphics[width=0.28\columnwidth]{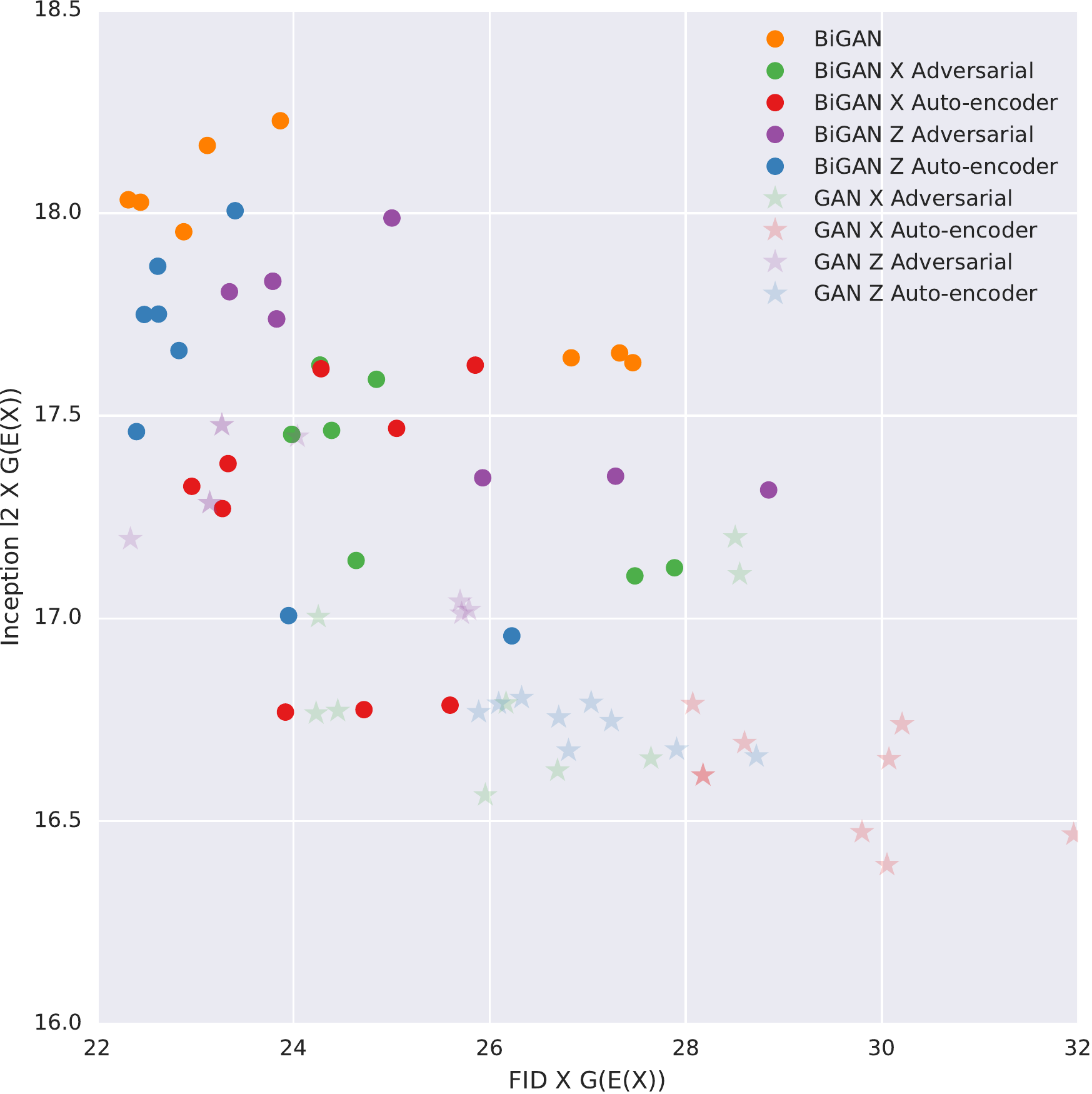}%
    \hspace{0.05\columnwidth}%
    \includegraphics[width=0.28\columnwidth]{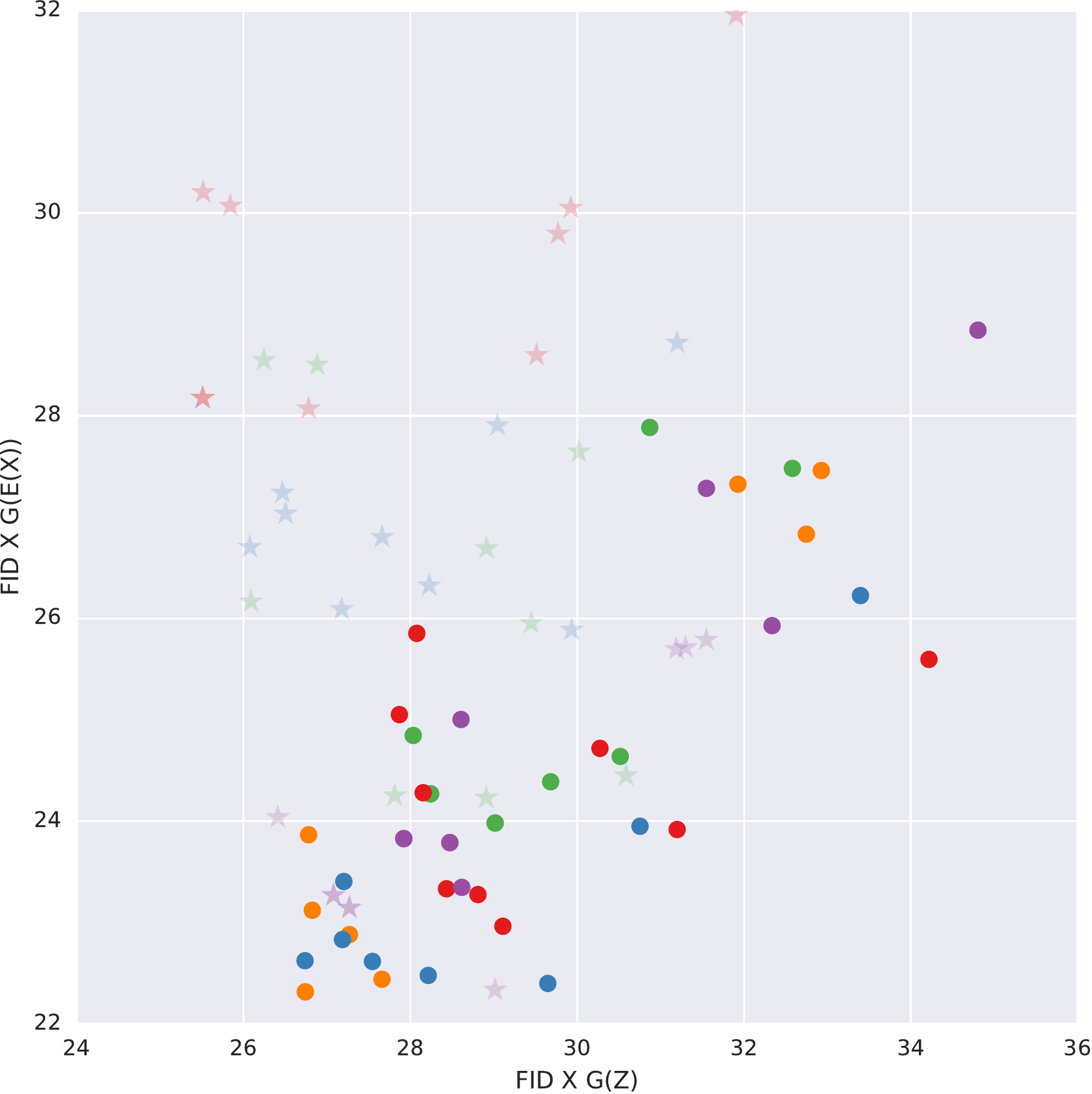}%
    \hspace{0.05\columnwidth}%
    \includegraphics[width=0.28\columnwidth]{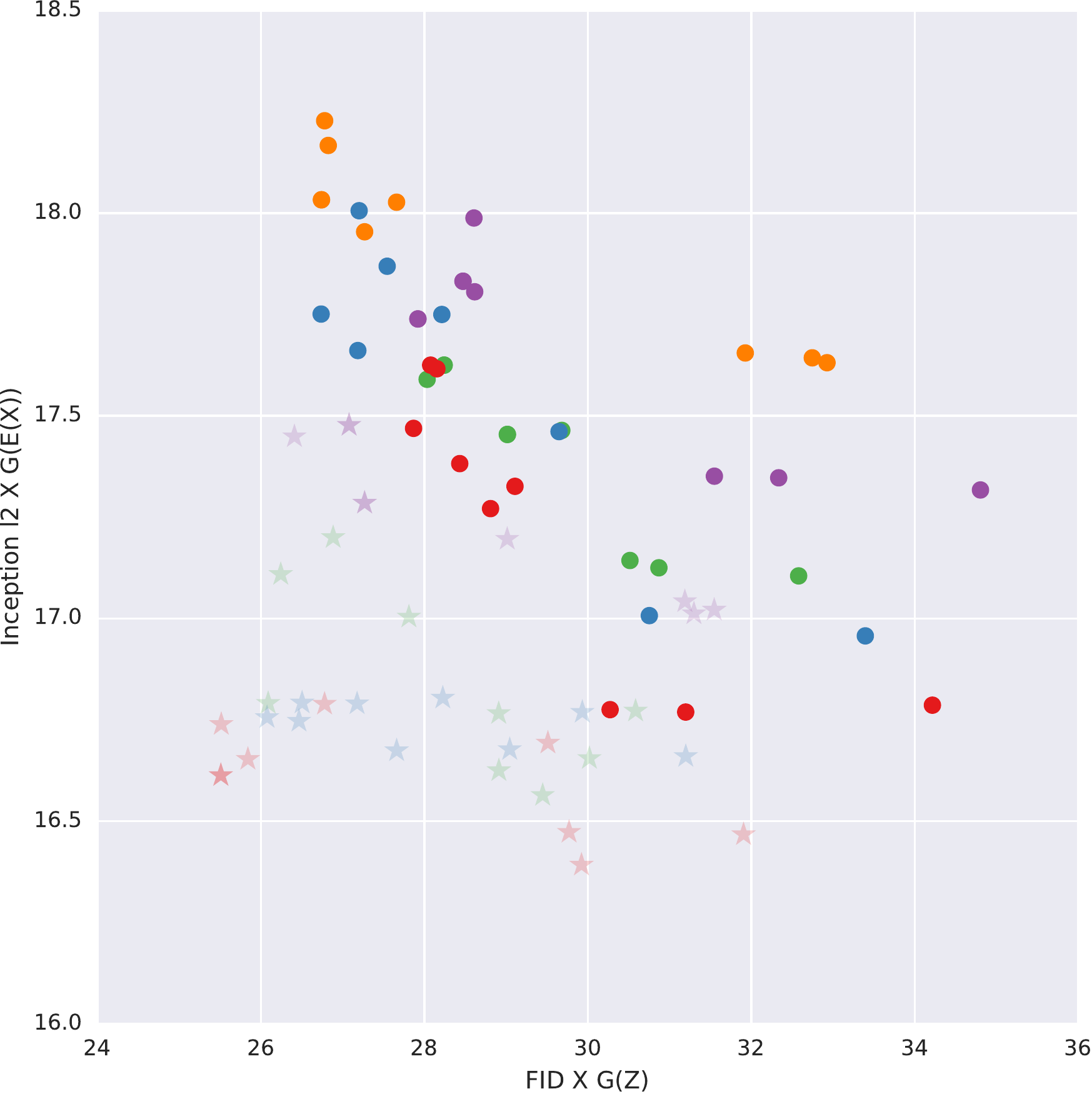}%

    \vspace{0.05\columnwidth}%
    
    \includegraphics[width=0.28\columnwidth]{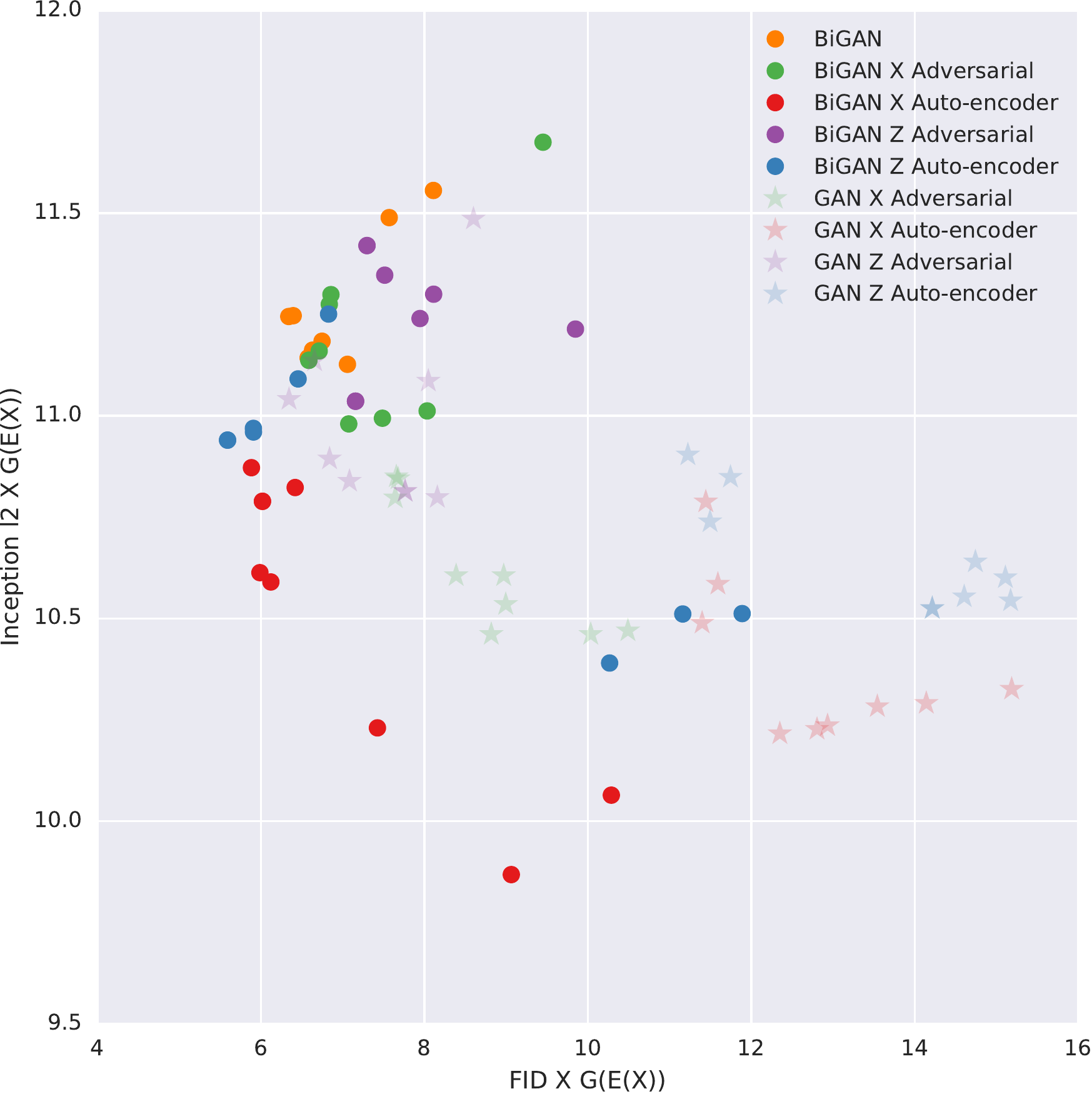}%
    \hspace{0.05\columnwidth}%
    \includegraphics[width=0.28\columnwidth]{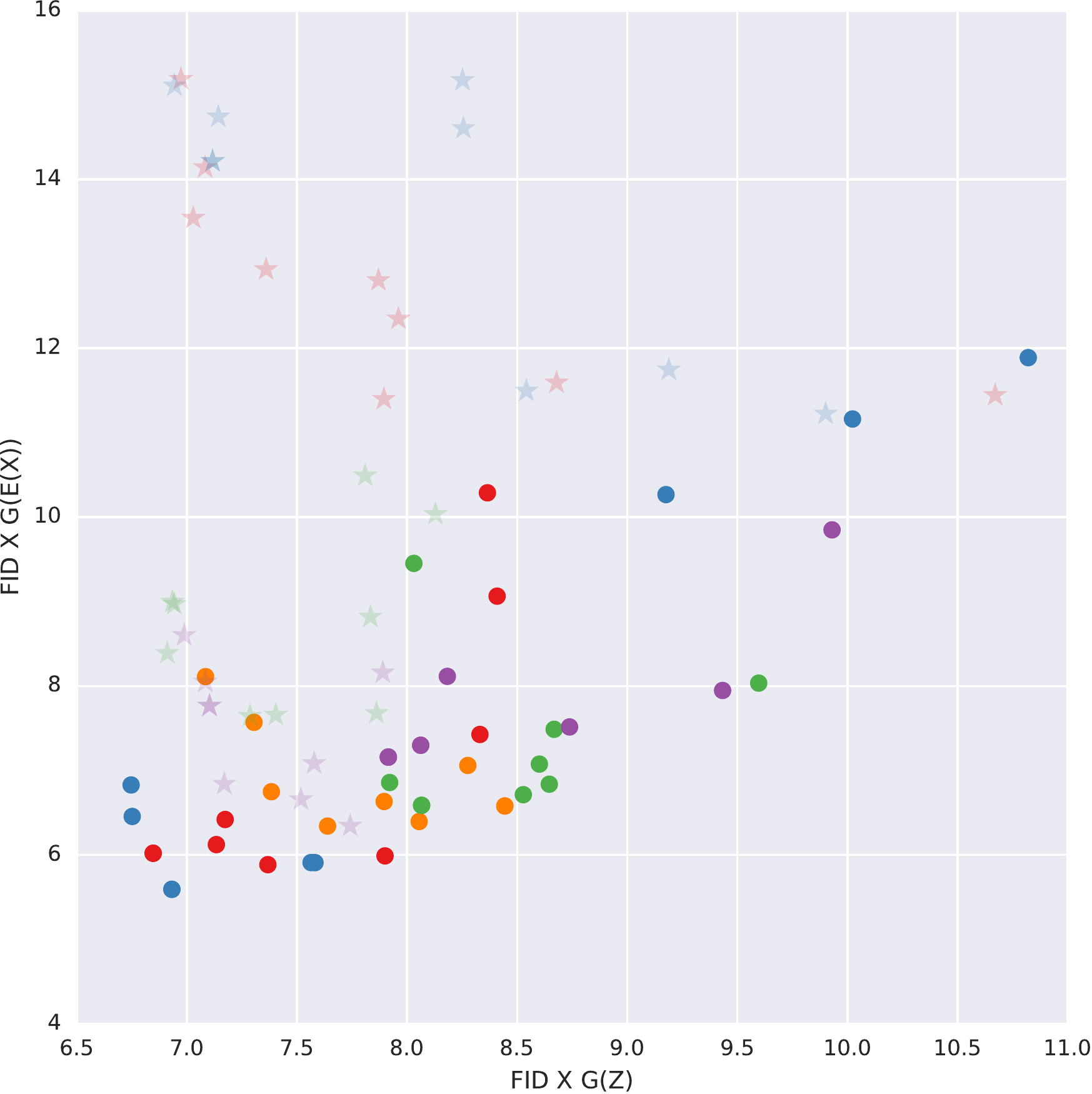}%
    \hspace{0.05\columnwidth}%
    \includegraphics[width=0.28\columnwidth]{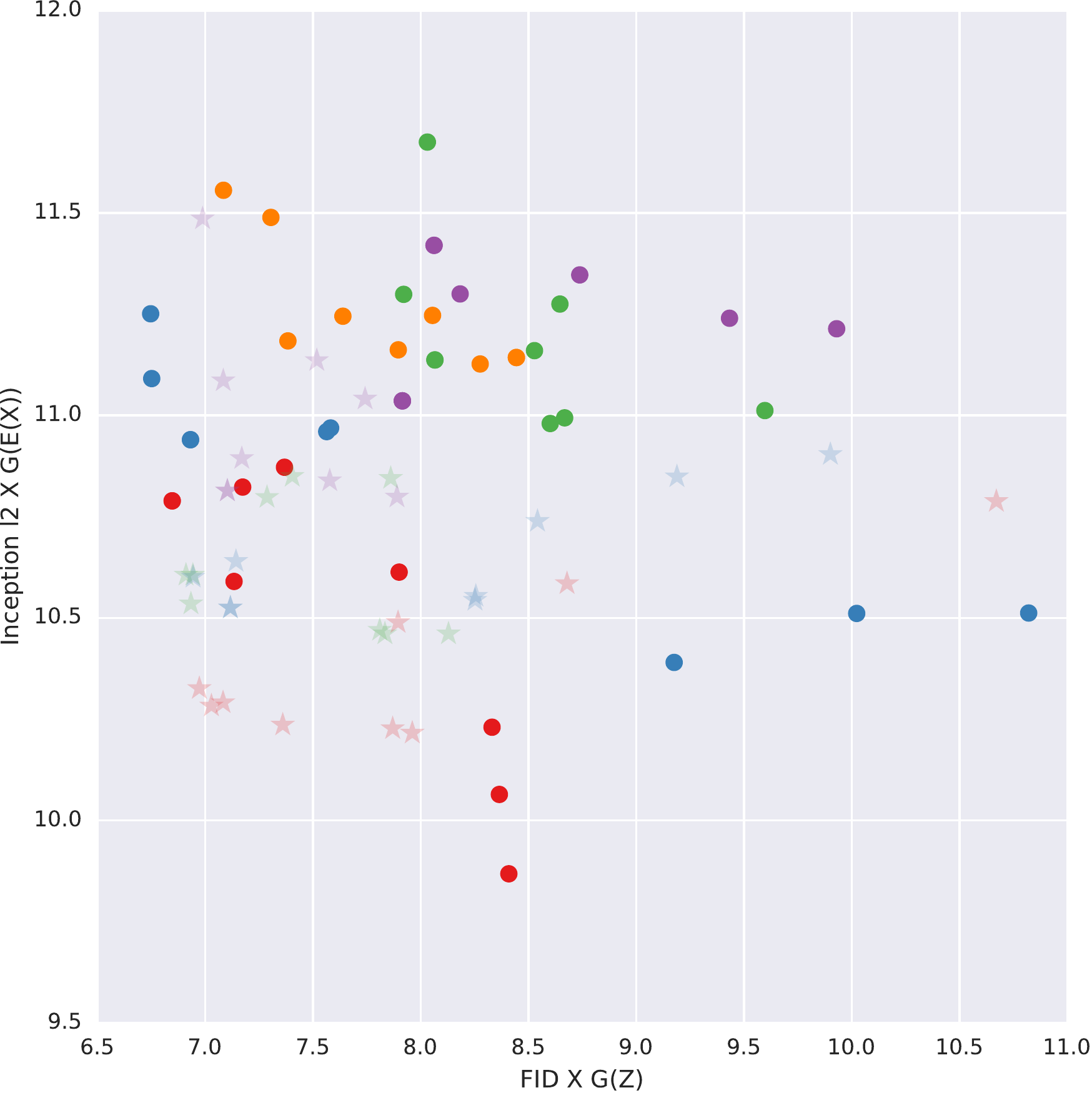}%

    \vspace{0.05\columnwidth}%
    
    \includegraphics[width=0.28\columnwidth]{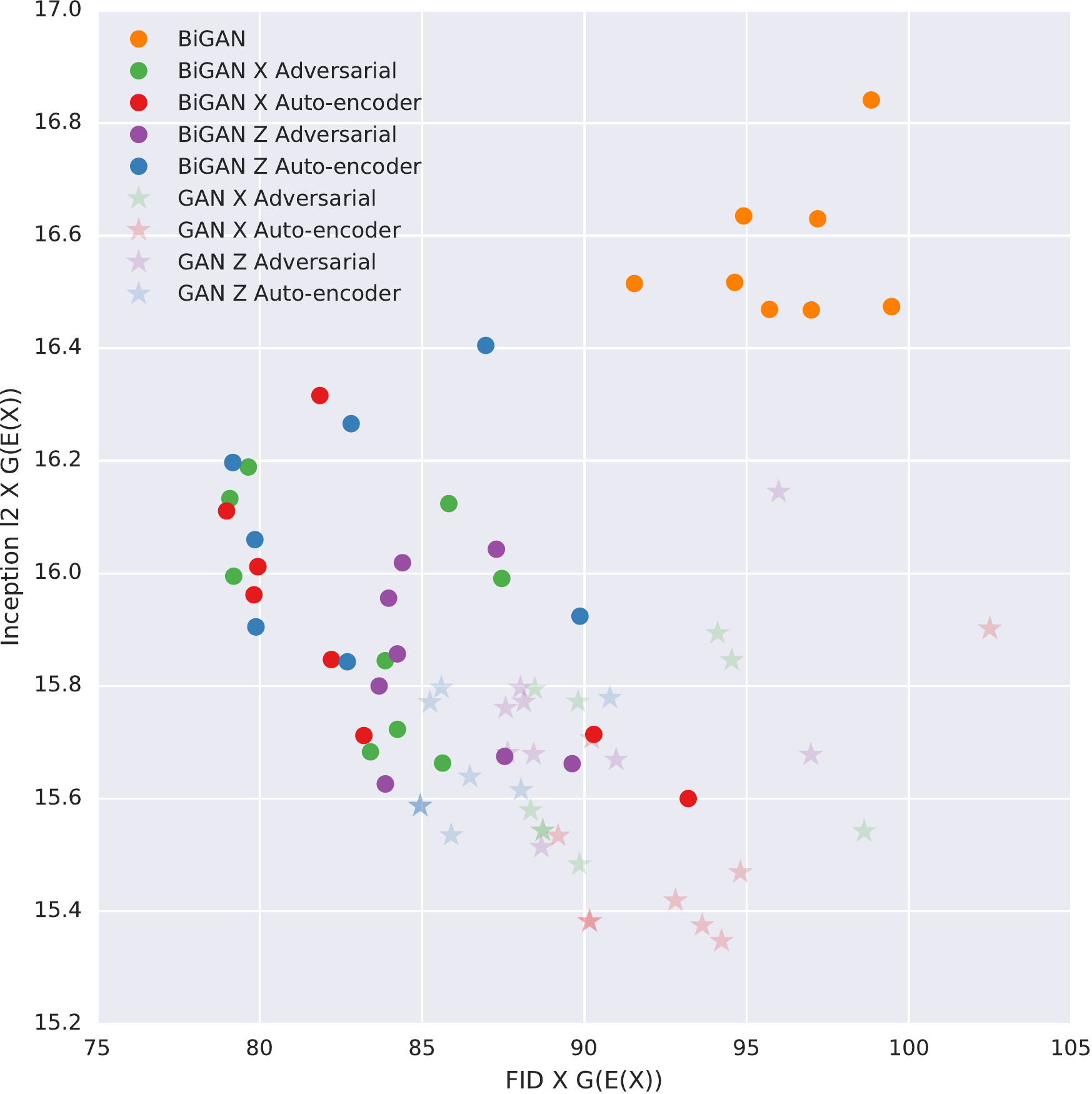}%
    \hspace{0.05\columnwidth}%
    \includegraphics[width=0.28\columnwidth]{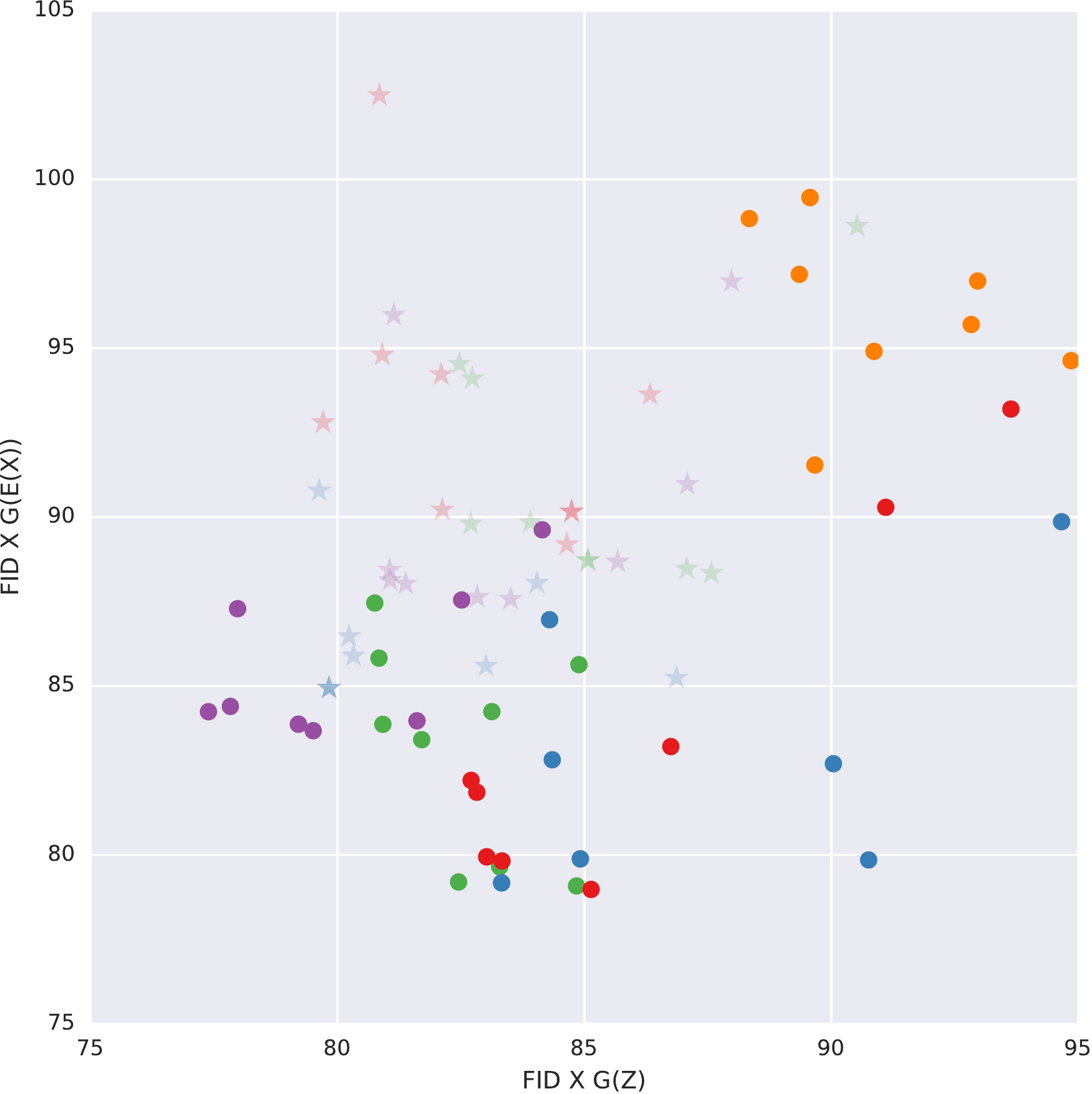}%
    \hspace{0.05\columnwidth}%
    \includegraphics[width=0.28\columnwidth]{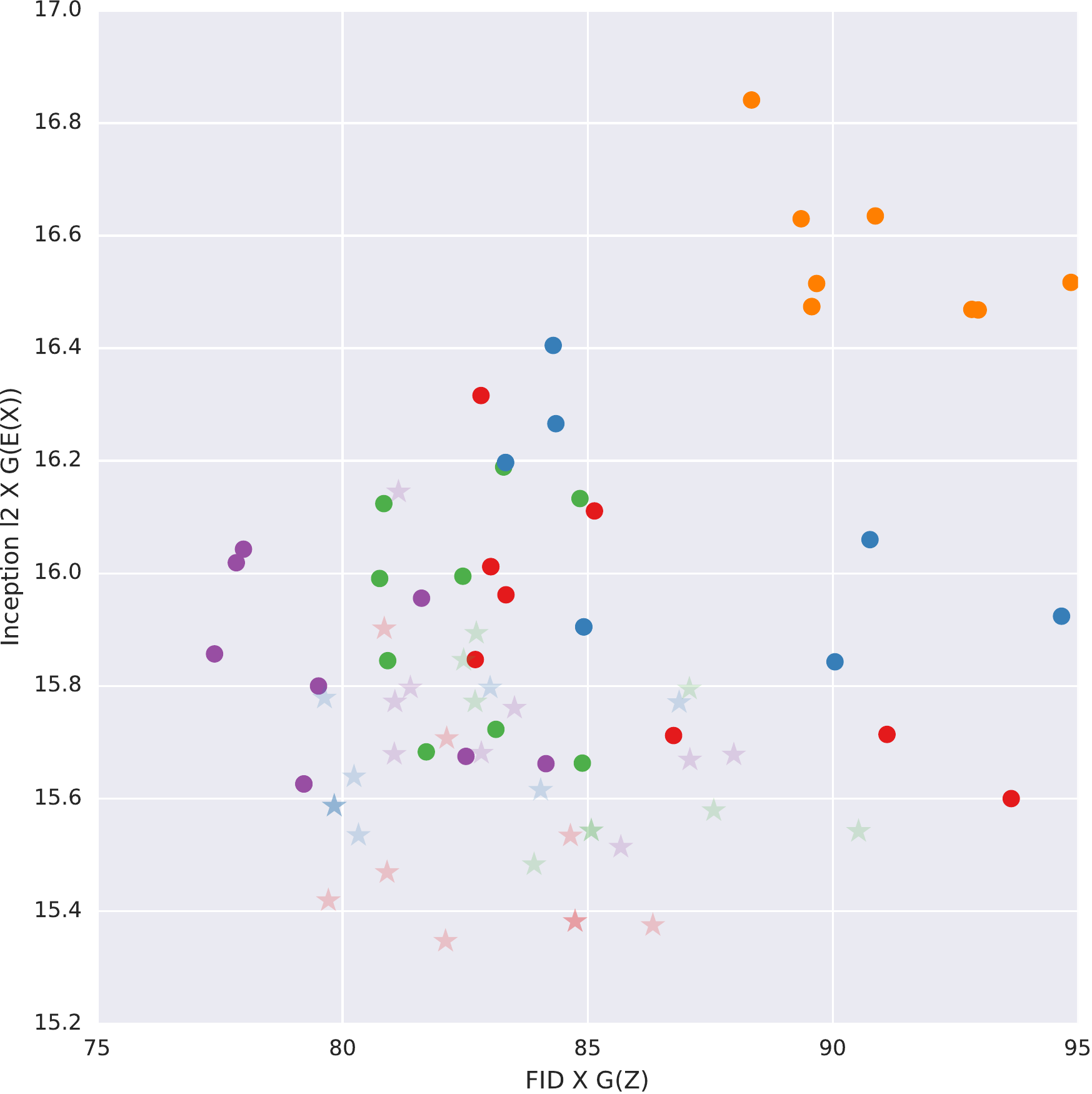}%

    \vspace{0.05\columnwidth}%

    \includegraphics[width=0.28\columnwidth]{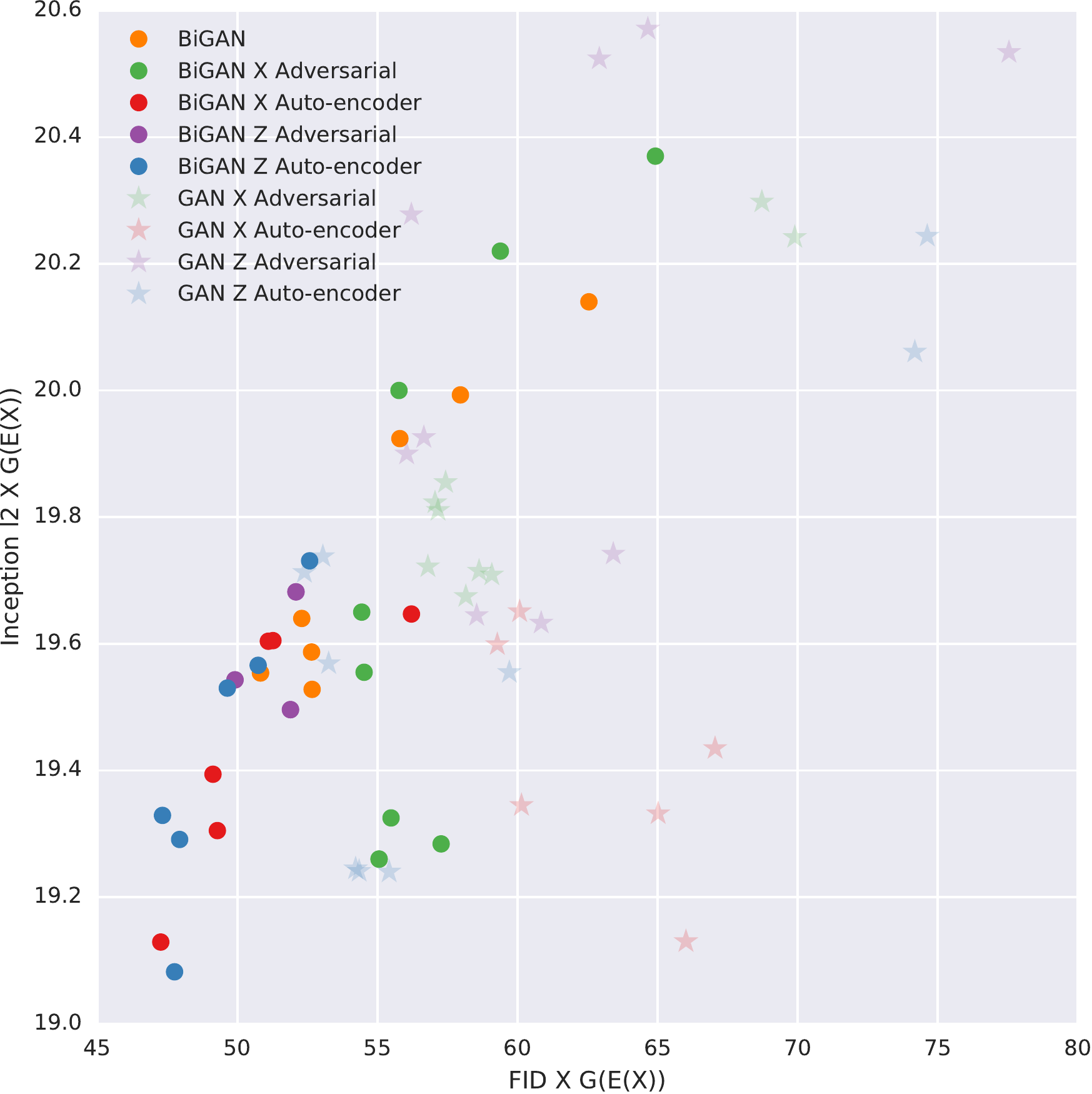}%
    \hspace{0.05\columnwidth}%
    \includegraphics[width=0.28\columnwidth]{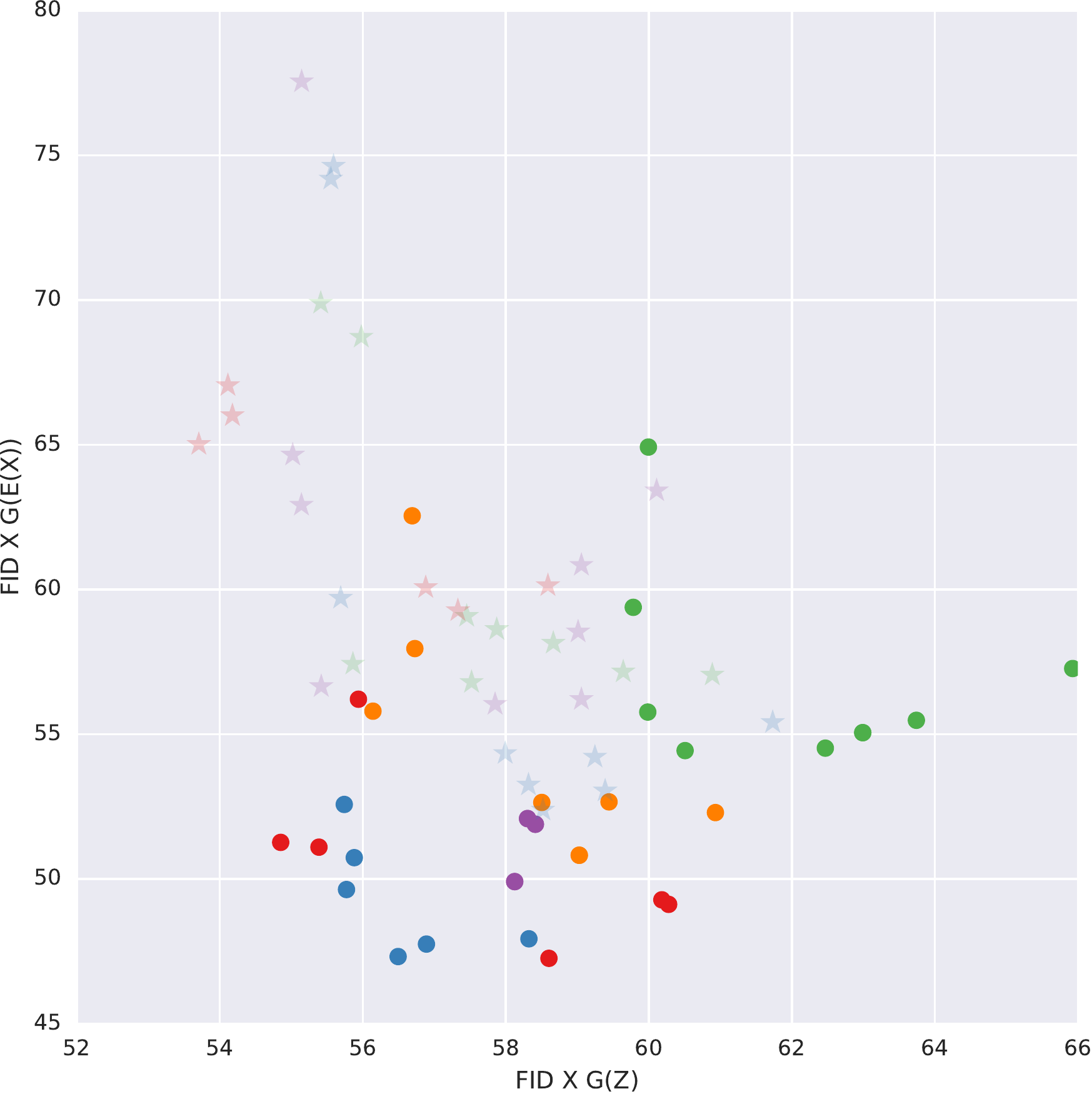}%
    \hspace{0.05\columnwidth}%
    \includegraphics[width=0.28\columnwidth]{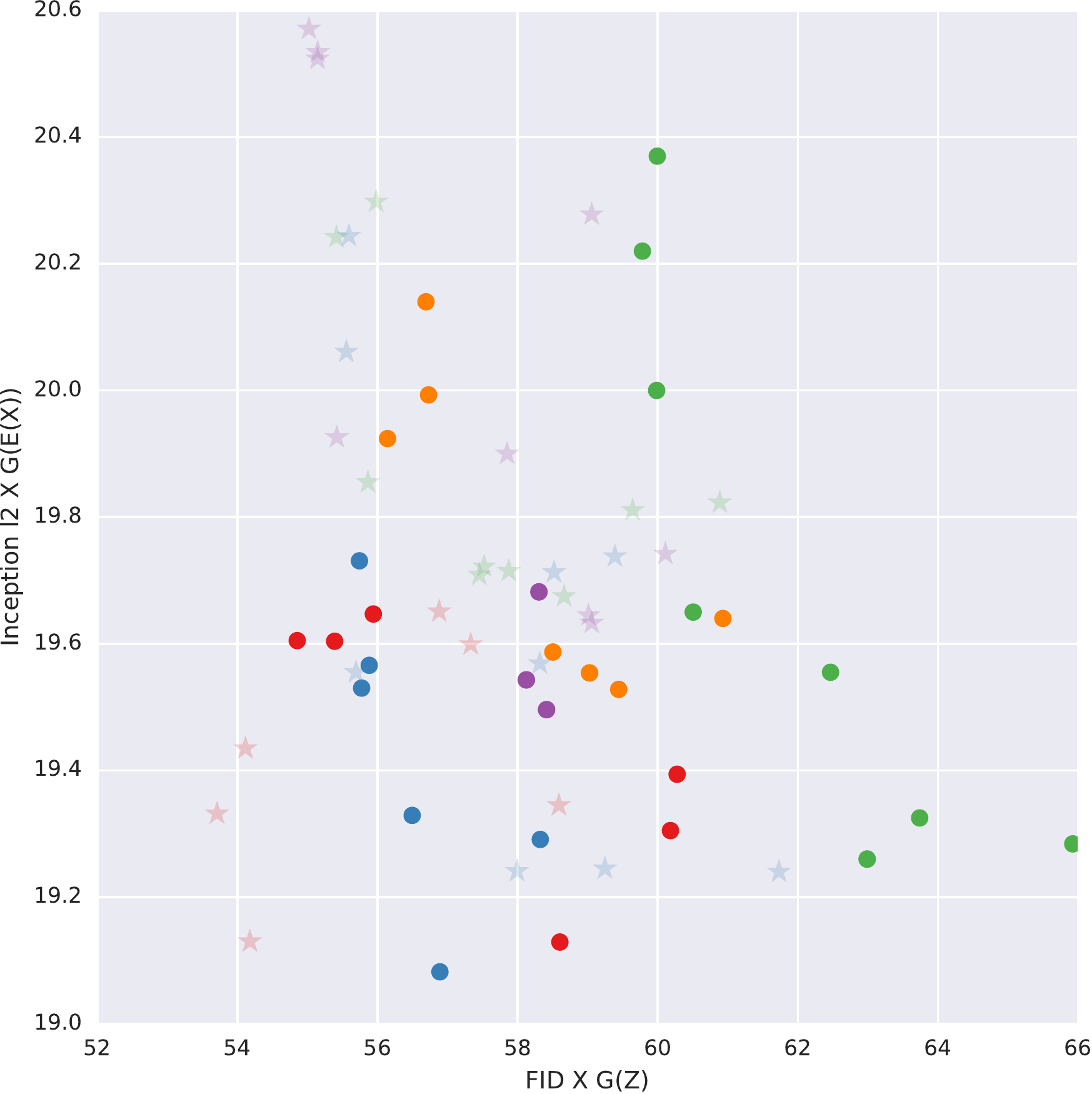}%
    \end{minipage}%
    \caption{Results of evaluation described in Section \ref{subsection:evaluation}. BiGAN results emphasised.}
    \label{fig:quant_results_gan_faded}
\end{figure}

\begin{figure}
    \centering
    \begin{minipage}{0.03\columnwidth}
        \begin{turn}{90}
                Imagenet \hspace{10em} Flintstones \hspace{10em} CelebA \hspace{10em} Cifar10
        \end{turn}
    \end{minipage}%
    \begin{minipage}{0.97\columnwidth}
    \includegraphics[width=0.28\columnwidth]{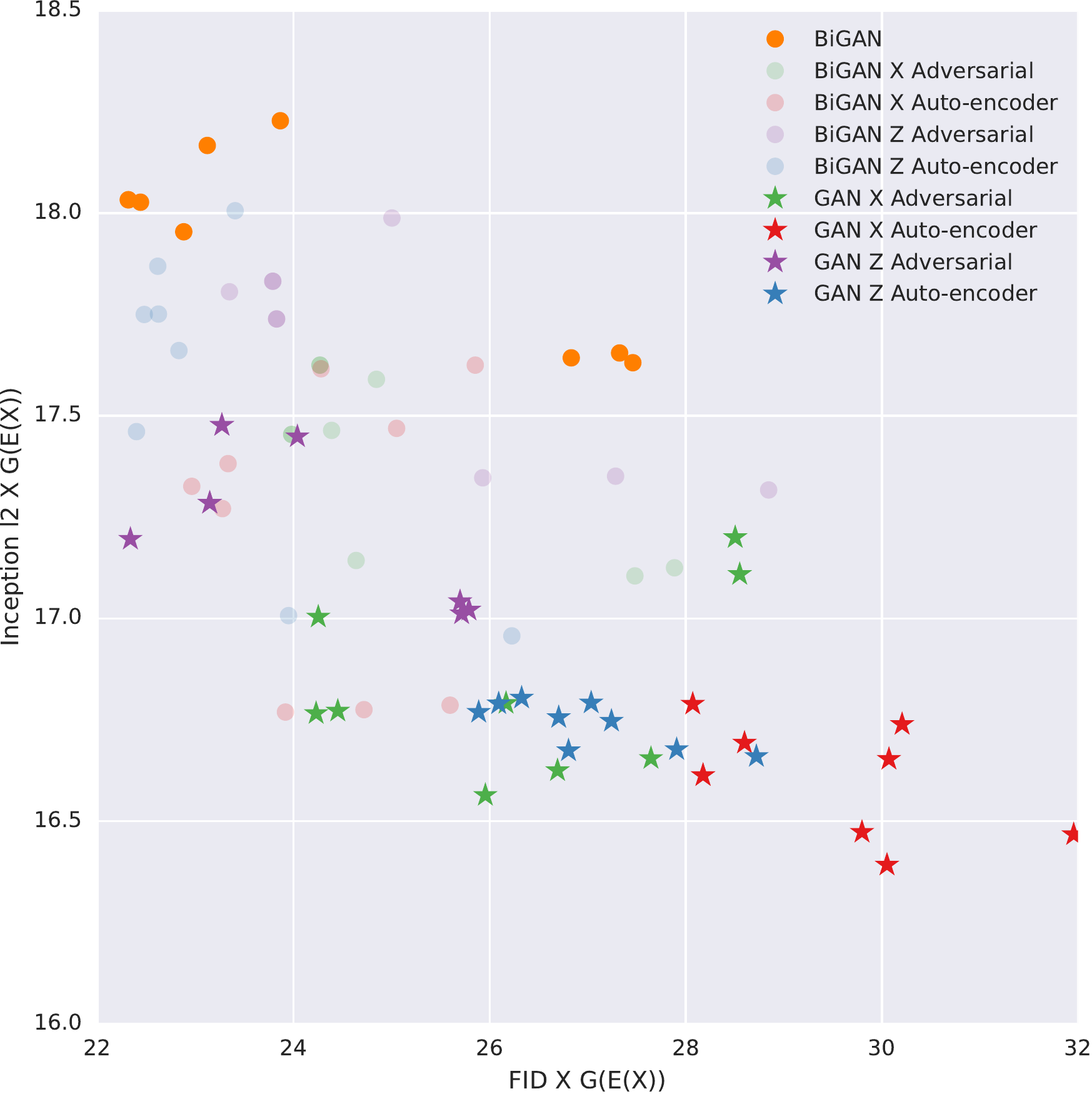}%
    \hspace{0.05\columnwidth}%
    \includegraphics[width=0.28\columnwidth]{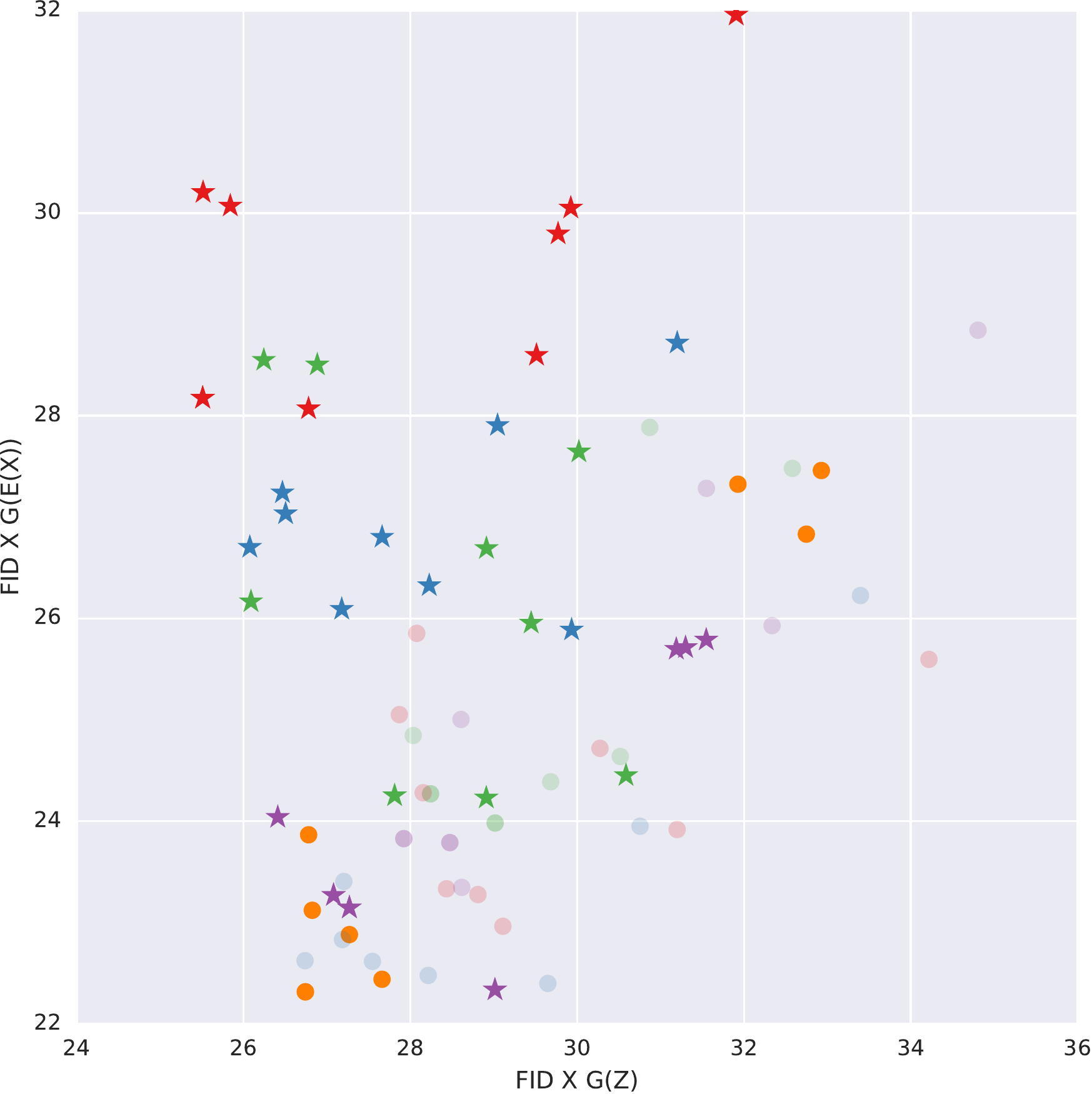}%
    \hspace{0.05\columnwidth}%
    \includegraphics[width=0.28\columnwidth]{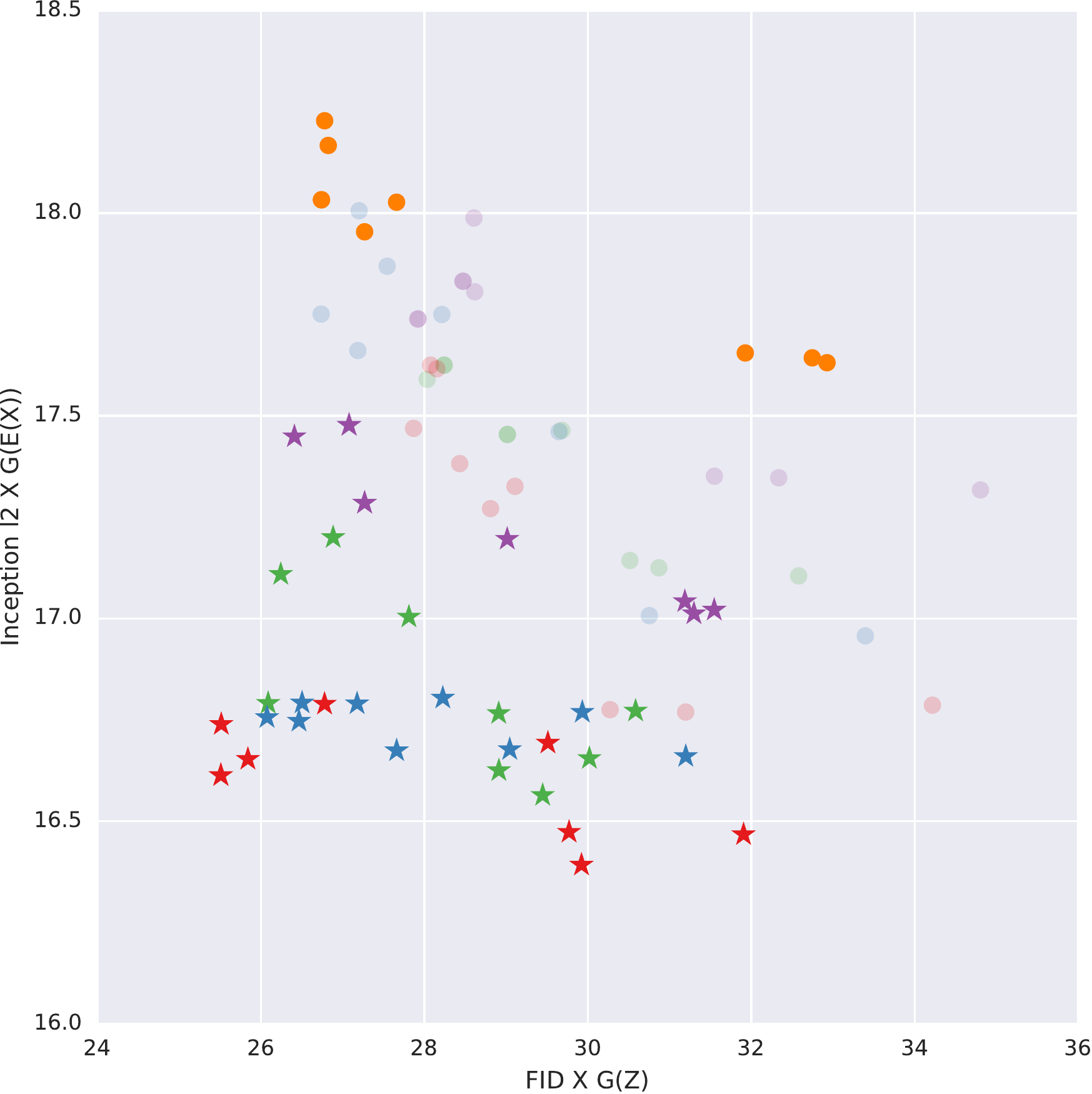}%

    \vspace{0.05\columnwidth}%

    \includegraphics[width=0.28\columnwidth]{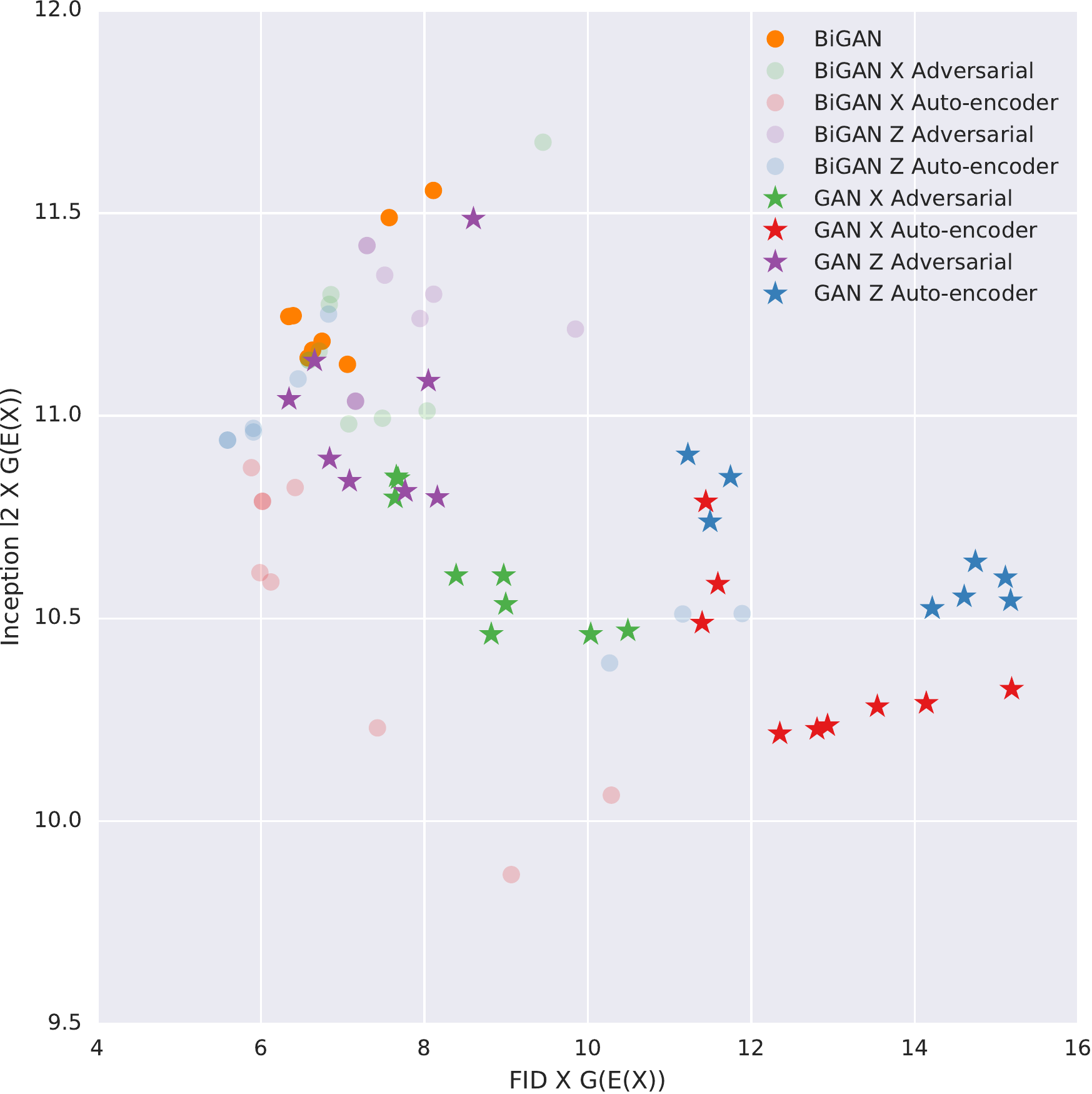}%
    \hspace{0.05\columnwidth}%
    \includegraphics[width=0.28\columnwidth]{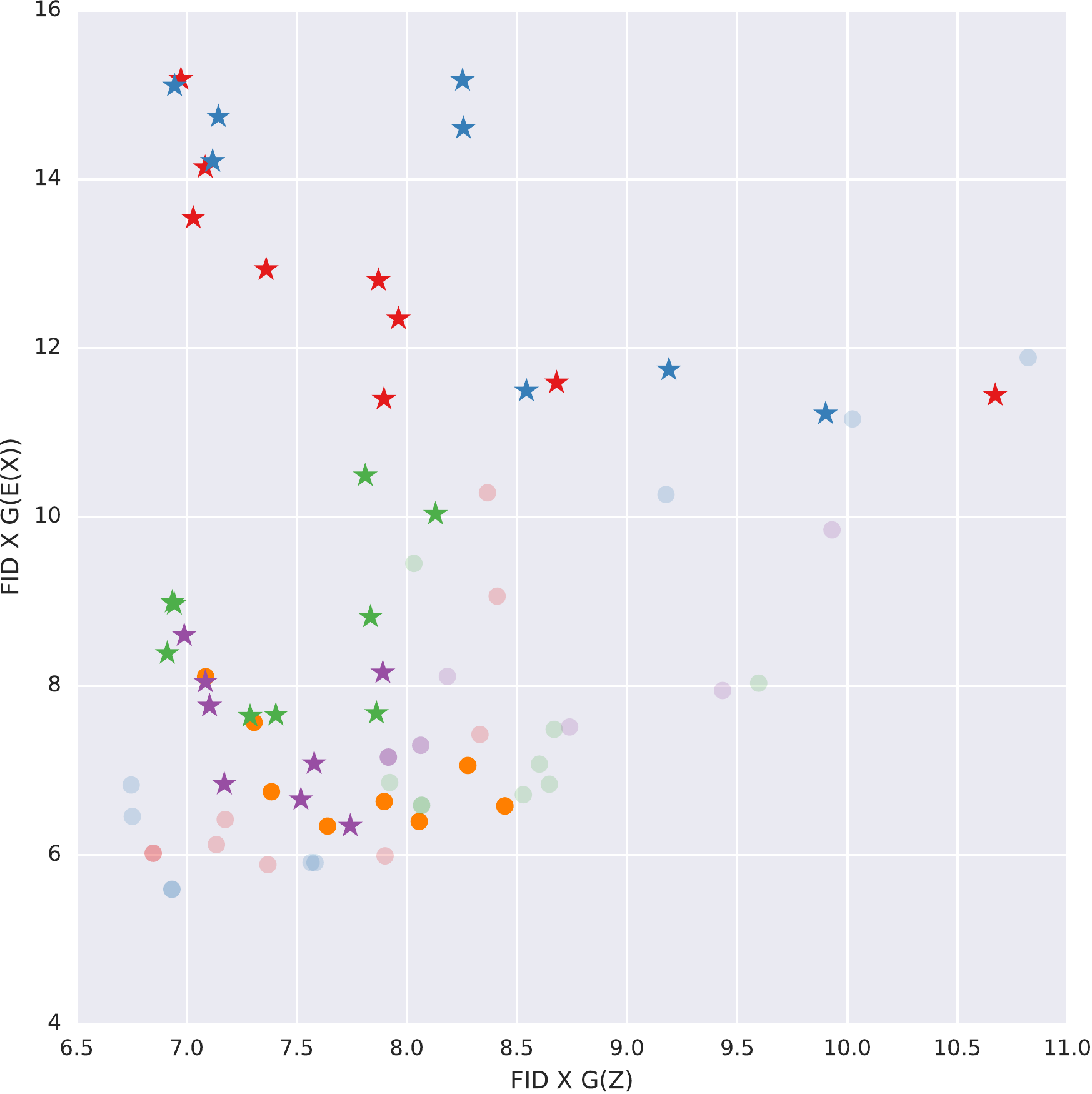}%
    \hspace{0.05\columnwidth}%
    \includegraphics[width=0.28\columnwidth]{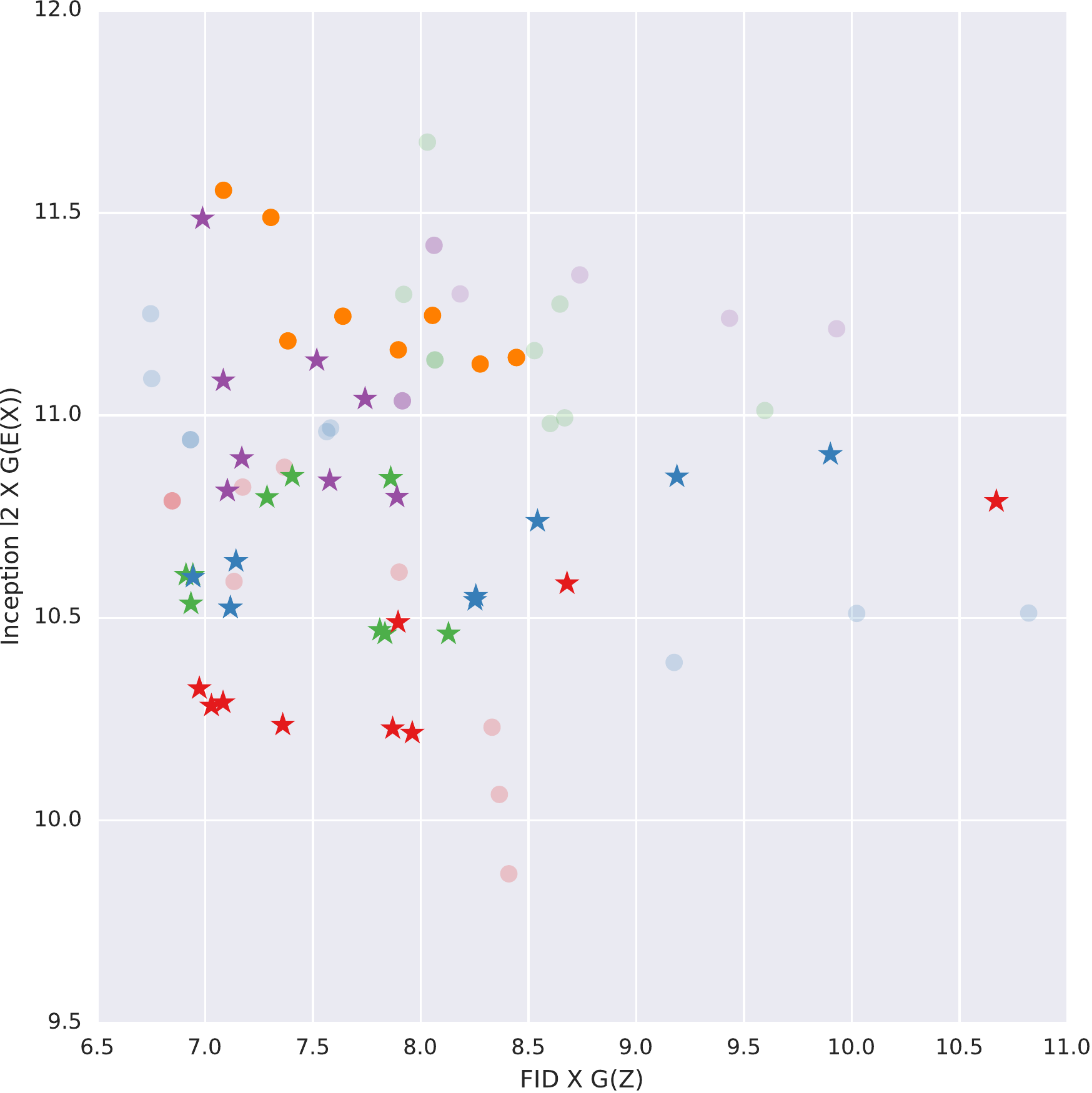}%

    \vspace{0.05\columnwidth}%

    \includegraphics[width=0.28\columnwidth]{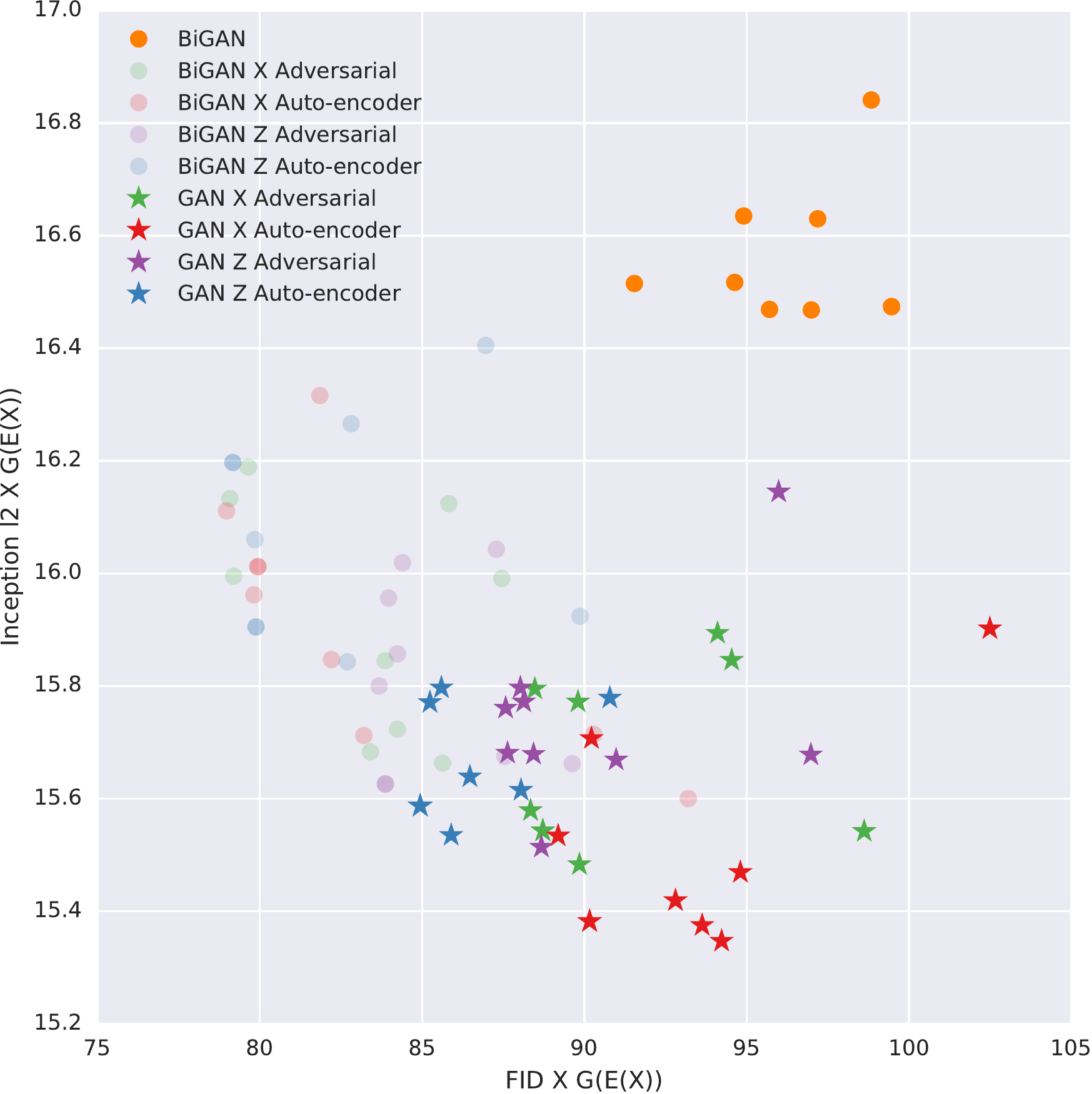}%
    \hspace{0.05\columnwidth}%
    \includegraphics[width=0.28\columnwidth]{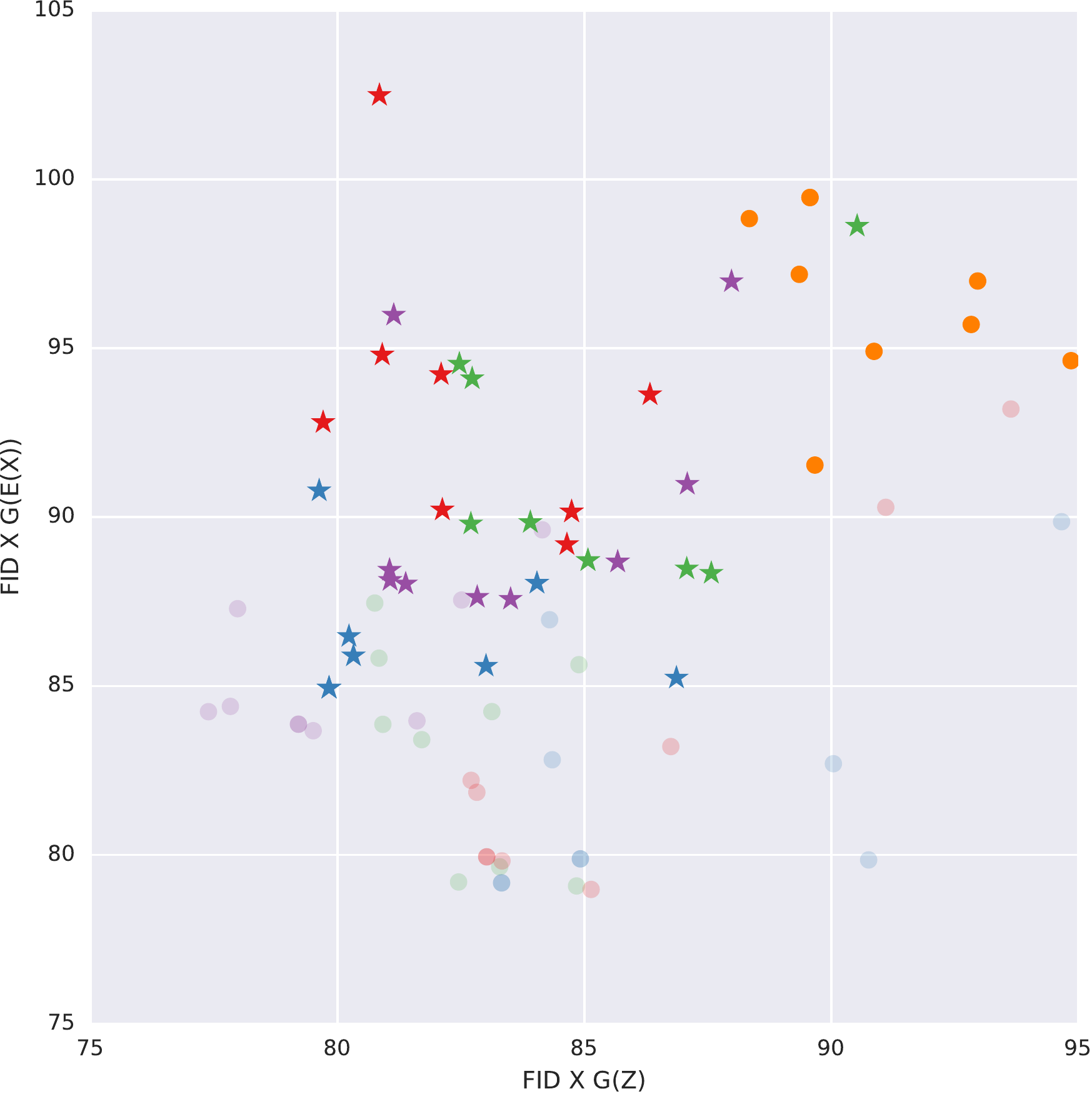}%
    \hspace{0.05\columnwidth}%
    \includegraphics[width=0.28\columnwidth]{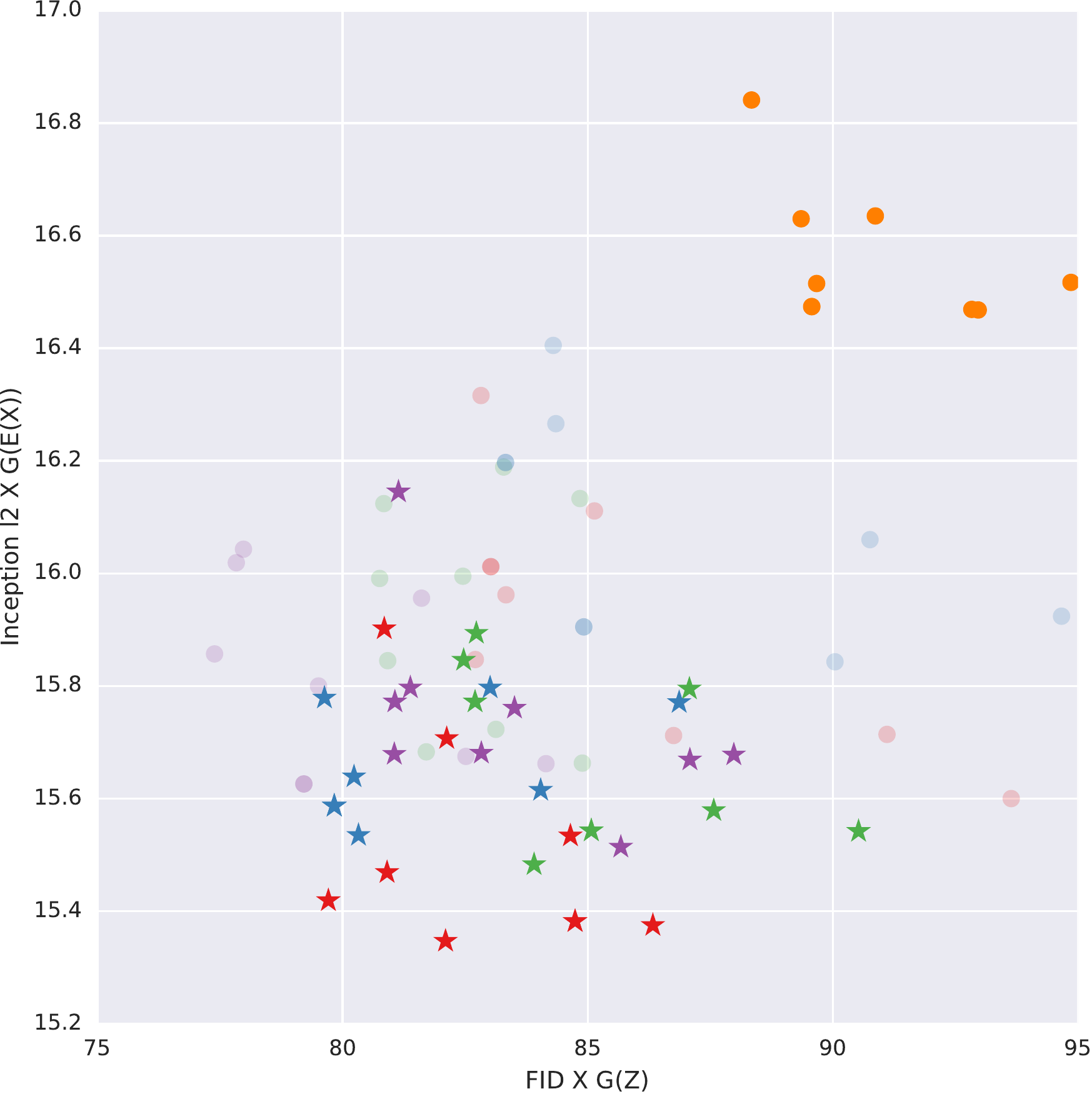}%

    \vspace{0.05\columnwidth}%

    \includegraphics[width=0.28\columnwidth]{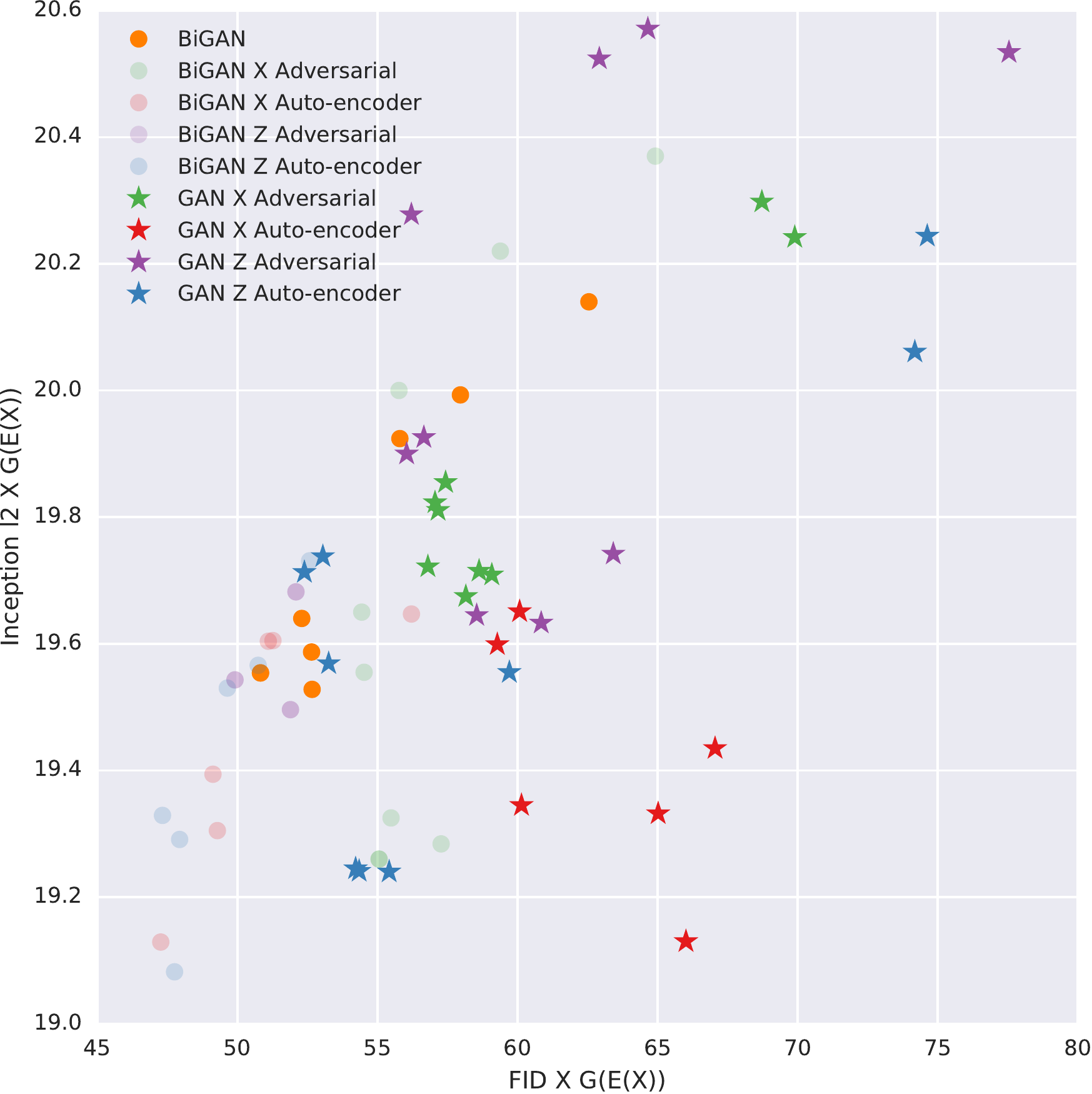}%
    \hspace{0.05\columnwidth}%
    \includegraphics[width=0.28\columnwidth]{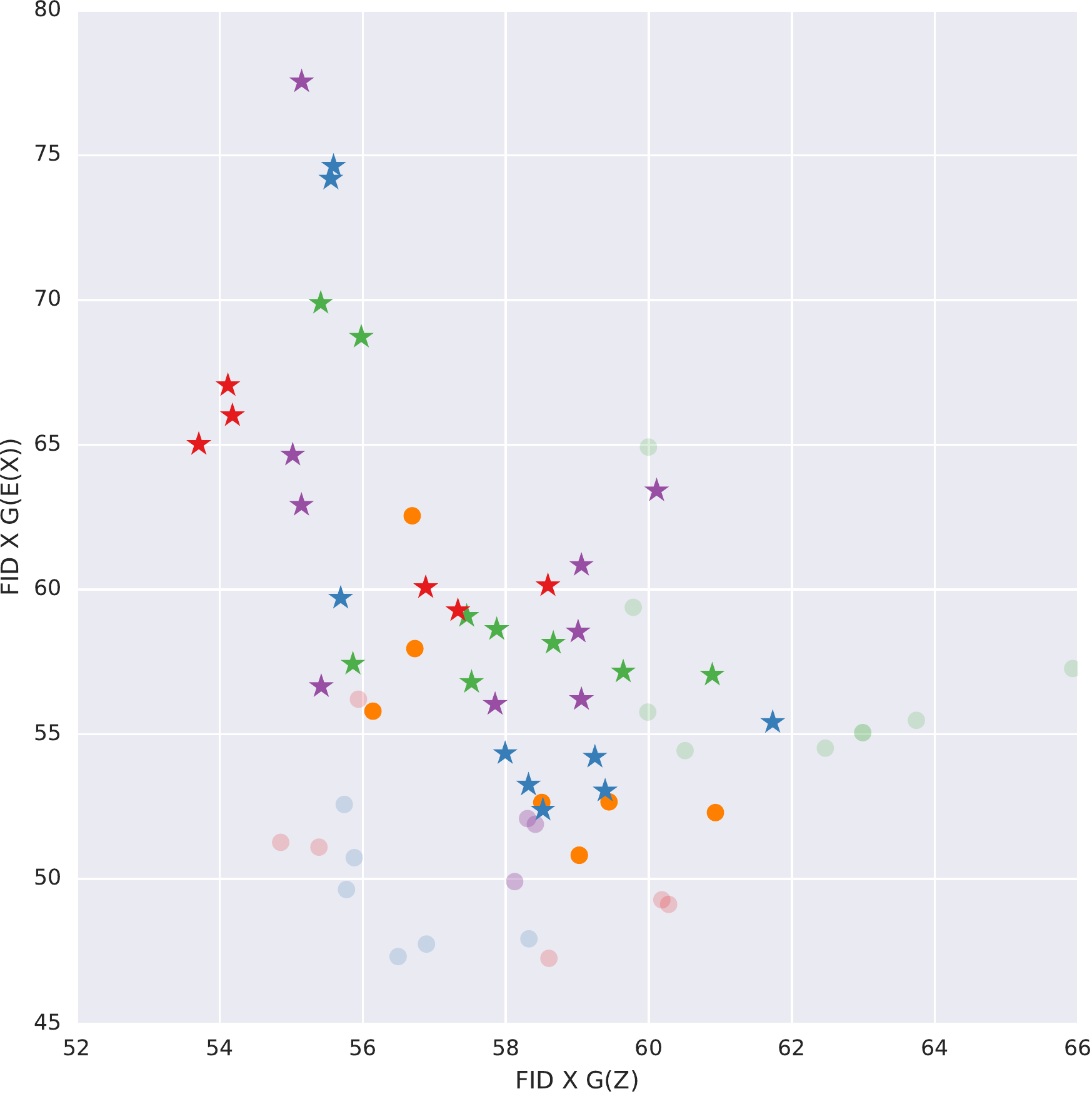}%
    \hspace{0.05\columnwidth}%
    \includegraphics[width=0.28\columnwidth]{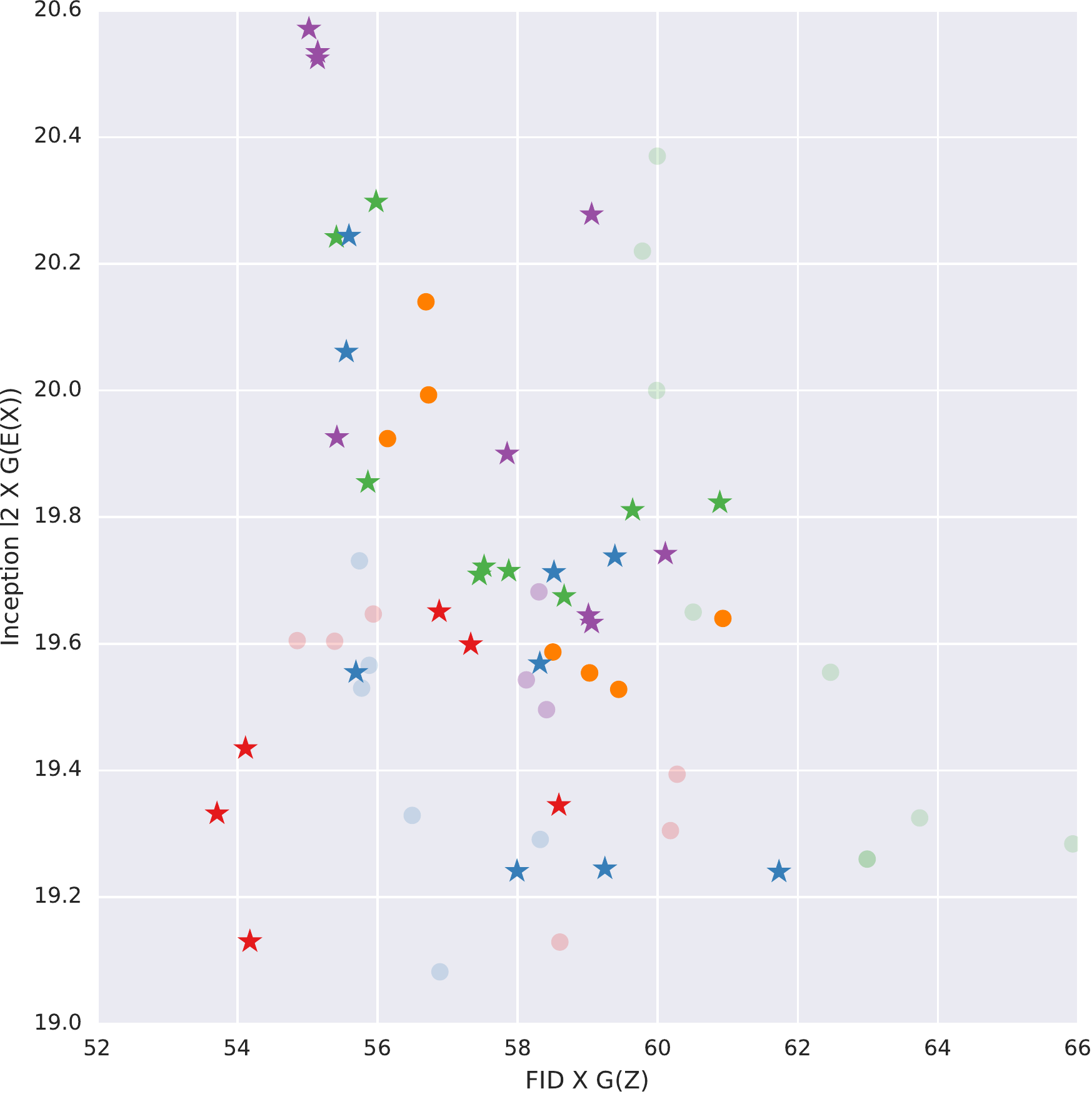}%

    \end{minipage}%
    \caption{Results of evaluation described in Section \ref{subsection:evaluation}. GAN results emphasised.}
    \label{fig:quant_results_bigan_faded}
\end{figure}

The results of our evaluation can be seen in Figures~\ref{fig:quant_results_gan_faded} and~\ref{fig:quant_results_bigan_faded}. 
These plots were generated by the following procedure.
For each model and each dataset, we took all saved checkpoints and performed the evaluation pipeline. 
We then sorted by each of the three evaluation metrics and considered the best three checkpoints under each metric. 
Thus, we end up with at most $9$ checkpoints for each dataset and model (we could have fewer than $9$ if the same checkpoint ranks amongst the best three under more than one metric).
Each row of Figures~\ref{fig:quant_results_gan_faded} and~\ref{fig:quant_results_bigan_faded} corresponds to one dataset, and each column corresponds to one pair of metrics.
Within each sub-figure, we plot all models together.
To avoid cluttered plots, we duplicated this into two Figures.
In Figure~\ref{fig:quant_results_gan_faded}, the results corresponding to `GAN + additional loss' models are faded out. 
In Figure~\ref{fig:quant_results_bigan_faded}, the results corresponding to `BiGAN + additional loss' models are faded out.

On \emph{Cifar10} there is a frontier of optimality reached collectively by all of the models, with a tradeoff between the realism of reconstructed images $FID(P_X, P_{G(E(X))})$ and the perceptual similarity between real images and their reconstructions (Inception $L_2$). In comparison to the GAN+ models, the BiGAN+ models produce reconstructions that are higher quality images but that resemble the originals less.

On \emph{CelebA}, excluding BiGAN + $\mathcal{X}$-AE we see a similar pattern to \emph{Cifar10}. On this particular dataset, BiGAN + $\mathcal{X}$-adversarial is arguably optimal. 

On \emph{Flintstones}, BiGAN is dominated by all other models, with these models exhibiting a similar optimal-frontier pattern: the BiGAN+ models produce reconstructions that are higher quality than the GAN+ models but that resemble the originals less.
It should be noted, however, that none of the models perform particularly well on this dataset, with no model acheiving $FID(P_X, P_G)<75$.

On \emph{ImageNet}, BiGAN + $\mathcal{X}$-AE and BiGAN + $\mathcal{Z}$-AE perform best out of all of the models.

If the goal is to produce reconstructions that look realistic as a first priority and resemble the originals as a second priority, then BiGAN + $\mathcal{X}$-AE seems to be the best model. 
Additionally, in two of the four datasets (\emph{ImageNet} and \emph{CelebA}) BiGAN + $\mathcal{X}$-AE also achieves amongst the best Inception $L_2$.
Training such a model may be expensive though, as the performance of BiGAN + $\mathcal{X}$-AE is sensitive to hyper-parameter choice (see Figure~\ref{fig:training_stability_bigan+_models_celeba}).

If, on the other hand, the main goal is reconstruction faithfulness and computational budget is limited, GAN + $\mathcal{X}$-AE has better performance compared with the BiGAN+ models and is more stable to train. 
It has the additional advantage of not having the extra weighting hyper-parameter to tune.
Although $FID(P_X, P_{G(E(X))})$ for GAN + $\mathcal{X}$-AE was always amongst the worst of all models, for all datasets the $FID(P_X, P_G)$ and Inception $L_2$ were amongst the best of all models.

\subsection{Training Stability}

\begin{figure}
    \vspace{-2em}
    \centering
    \begin{minipage}{0.03\columnwidth}
        \begin{turn}{90}
                GAN + $\Z$-AE \hspace{4em} GAN + $\X$-AE \hspace{4em} GAN + $\Z$-adversarial \hspace{3em} GAN + $\X$-adversarial \hspace{5em} BiGAN
        \end{turn}
    \end{minipage}%
    \begin{minipage}{0.97\columnwidth}
    
    \centering

    \includegraphics[width=0.8\columnwidth]{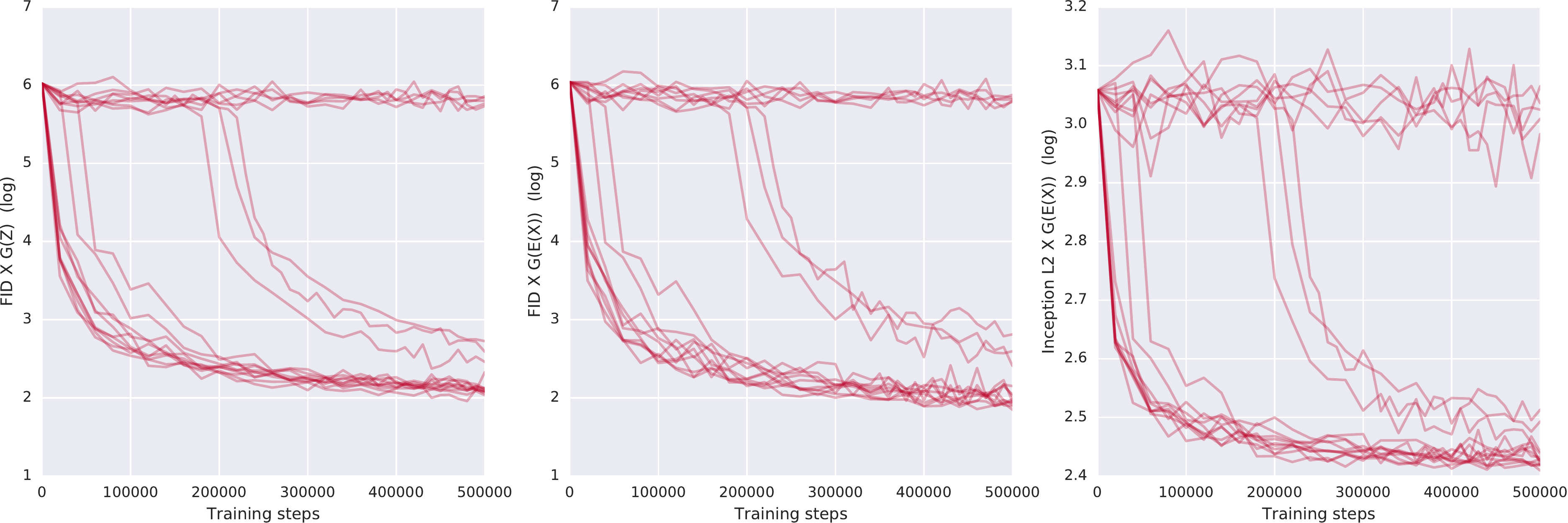}%

    \vspace{0.01\columnwidth}%
    
    \includegraphics[width=0.8\columnwidth]{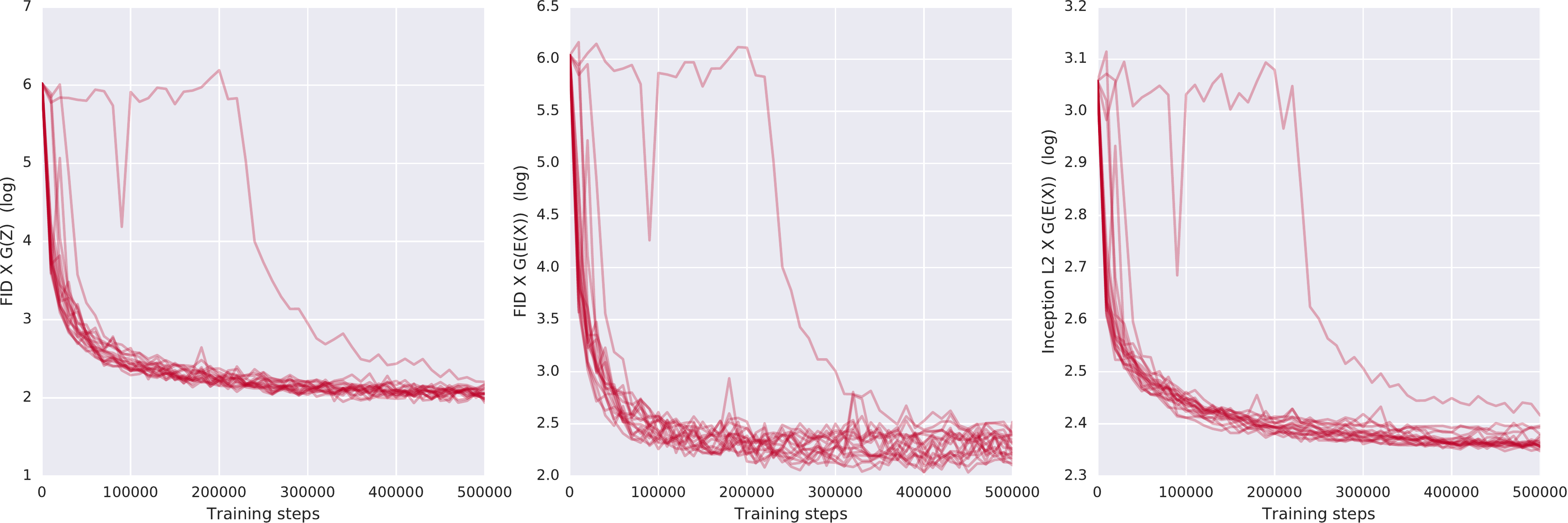}%

    \vspace{0.01\columnwidth}%

    \includegraphics[width=0.8\columnwidth]{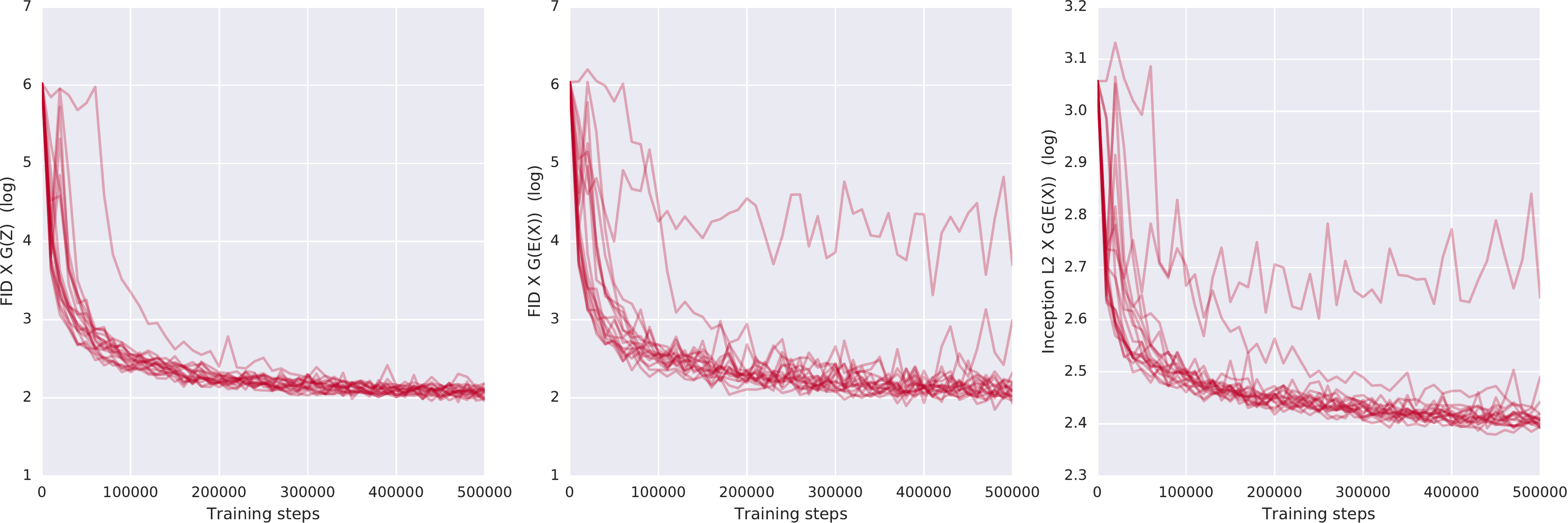}%

    \vspace{0.01\columnwidth}%

    \includegraphics[width=0.8\columnwidth]{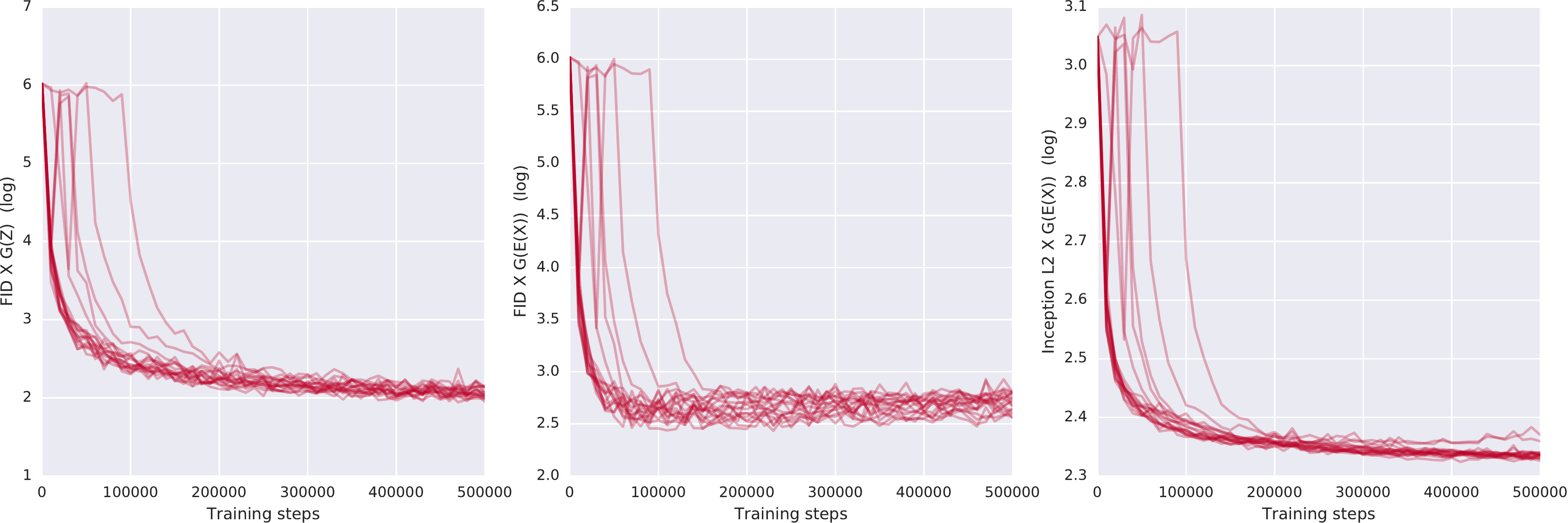}%

    \vspace{0.01\columnwidth}%

    \includegraphics[width=0.8\columnwidth]{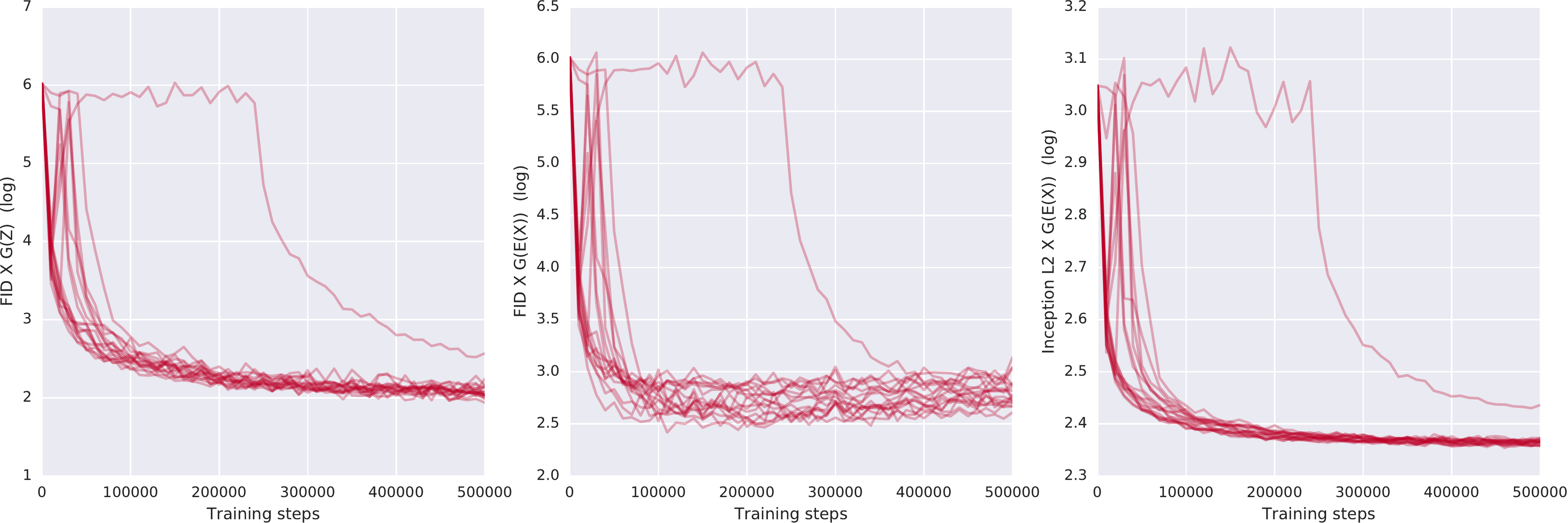}%

    \end{minipage}%
    \caption{Stability of training of BiGAN and GAN+ models on CelebA. Each line represents how evaluation statistics change throughout changing for one combination of hyper-parameters. In comparison to BiGAN, the GAN+ models are stable to train, with most combinations of hyper-parameters leading to similar levels of performance. Behaviour was similar for the other datasets.}
    \label{fig:training_stability_gan+_models_celeba}
\end{figure}

\begin{figure}
    \vspace{-2em}
    \centering
    \begin{minipage}{0.03\columnwidth}
        \begin{turn}{90}
                \hspace{3em} BiGAN + $\Z$-AE \hspace{8em} BiGAN + $\X$-AE \hspace{7em} BiGAN + $\Z$-adversarial \hspace{4em} BiGAN + $\X$-adversarial
        \end{turn}
    \end{minipage}%
    \begin{minipage}{0.97\columnwidth}
    \includegraphics[width=0.9\columnwidth]{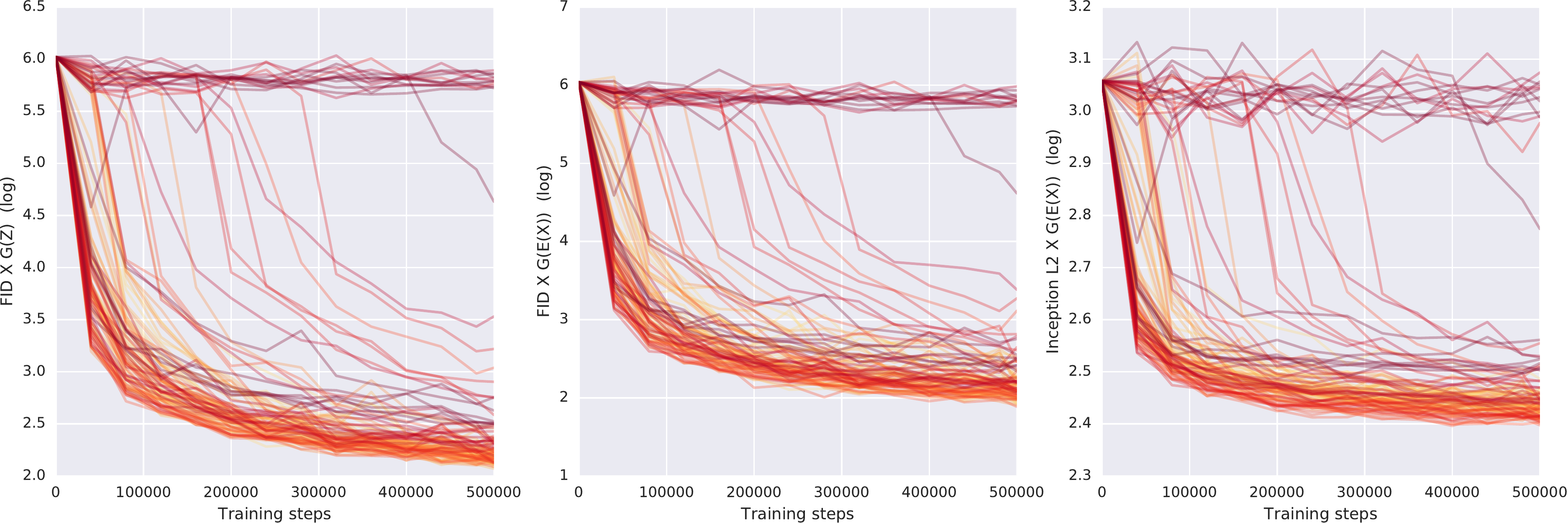}%

    \vspace{0.05\columnwidth}%

    \includegraphics[width=0.9\columnwidth]{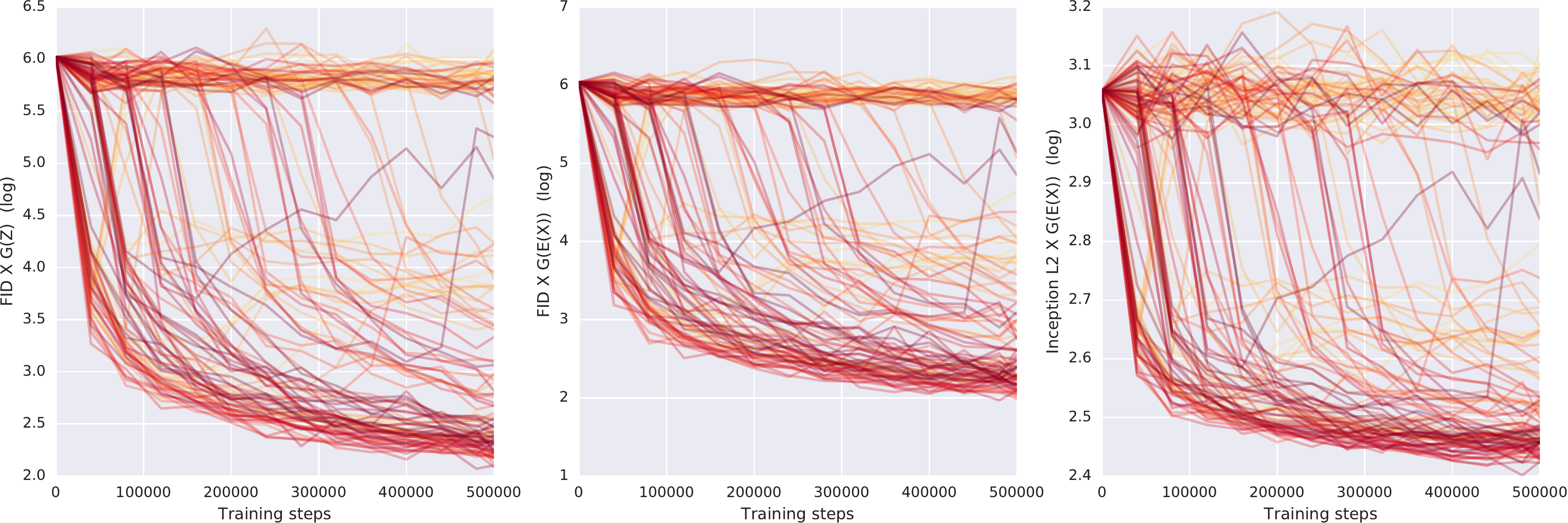}%

    \vspace{0.05\columnwidth}%

    \includegraphics[width=0.9\columnwidth]{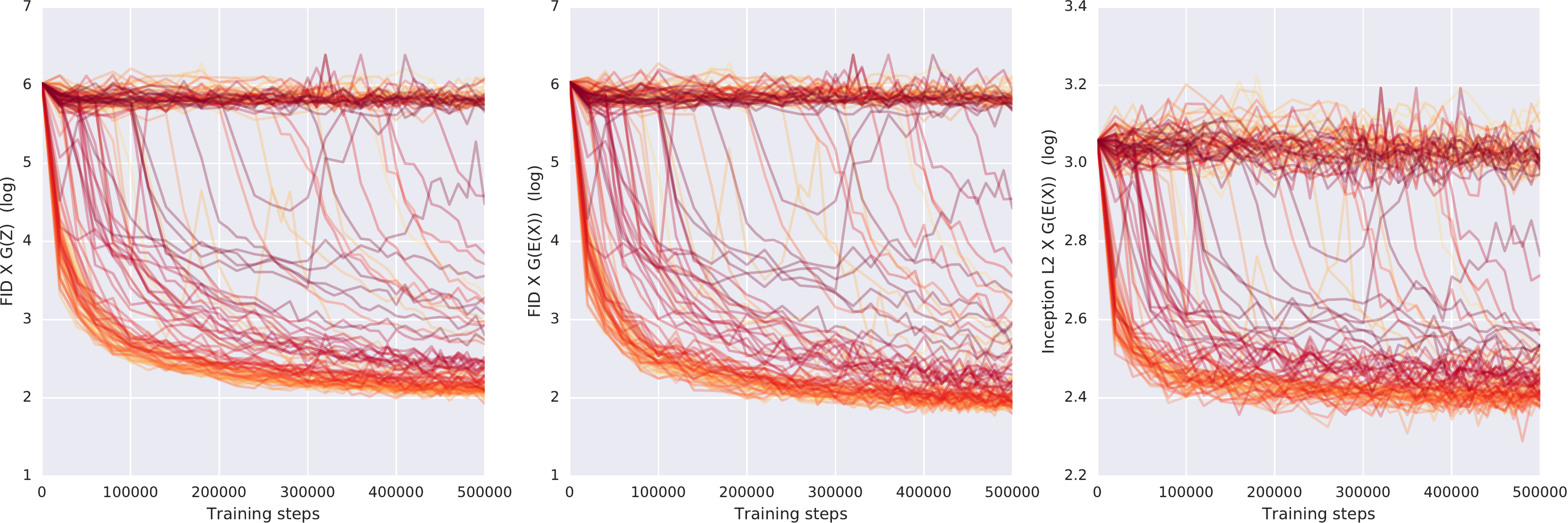}%

    \vspace{0.05\columnwidth}%

    \includegraphics[width=0.9\columnwidth]{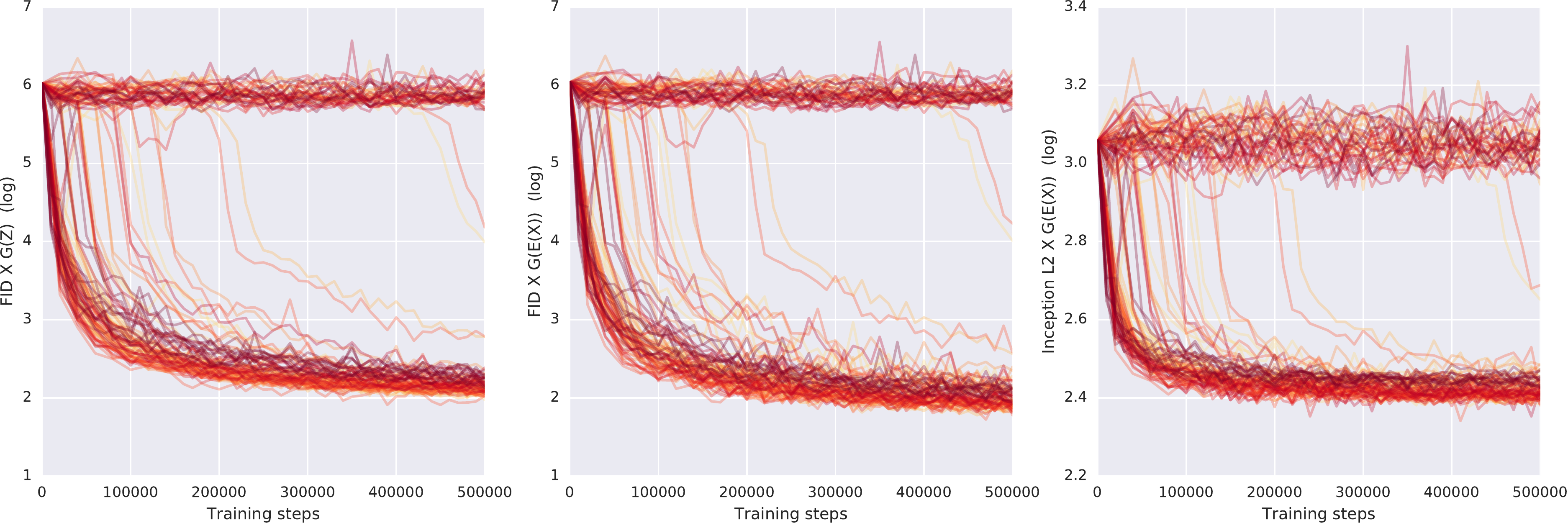}%

    \end{minipage}%
    \caption{Stability of training of BiGAN+ models on CelebA. Each line represents how evaluation statistics change throughout changing for one combination of hyper-parameters, with different colours representing different $\lambda$ weighting the additional loss. These models are unstable to train, requiring extensive hyper-parameter search to obtain good performance. Behaviour was similar for the other datasets. There are more lines per plot compared to Figure~\ref{fig:training_stability_gan+_models_celeba} due to search over both $\lambda$ and optimisation hyper-parameters.} 
    \label{fig:training_stability_bigan+_models_celeba}
\end{figure}

Figures~\ref{fig:training_stability_gan+_models_celeba} and~\ref{fig:training_stability_bigan+_models_celeba} show the evolution of each of the metrics throughout training for each experiment performed using the \emph{CelebA} dataset. 
Since the results for the other datasets are similar, we omit them for brevity.

Figure~\ref{fig:training_stability_gan+_models_celeba} displays the results for BiGAN and GAN+ models. 
In these plots, different lines correspond to different optimisation hyper-parameters.
They show that BiGAN is unstable to train relative to the GAN+ models, with the GAN+AE models being the most stable to train. 
Note that since the training objective of the generator $G$ is the same as in the usual GAN loss, the $FID(P_X, P_G)$ scores (left column) in each of the GAN+ models should all be comparable to training usual GANs.
Although GANs are notoriously unstable to train, we found that with the regularisation and normalisation we used, the $FID(P_X, P_G)$ scores were remarkably robust to different optimisation hyper-parameter choices accross all of the datasets, with almost all runs converging to similar FID scores within each dataset.

Figure~\ref{fig:training_stability_bigan+_models_celeba} displays the results for BiGAN+ models. 
In these plots, different colours correspond to different settings for the hyper-parameter $\lambda$ determining the weighting of the additional loss, with different lines of the same colour having the same $\lambda$ but different optimisation hyper-parameters.
They show that these models are sensitive to the choice of both $\lambda$ and the optimisation hyper-parameters.
In particular, for all models many training runs resulted in very poor performance, not significantly different from that of randomly initialised networks.
Unlike the GAN+ models, the training of the BiGAN generator is affected directly by the encoder. 
Hence, altering the loss function of the encoder is expected to have some impact on the training of the generator.

\subsection{Qualitative Results}

We include in Appendix \ref{appendix:qualitative-results} examples of images from one model trained on each dataset,
along with the hyper-parameters used for each training run and the statistics $FID(P_X, P_G)$, $FID(P_X, P_{G(E(X))})$ and Inception $L_2(P_X, P_{G(E(X))})$. 
These are displayed in Figures~\ref{fig:qualitative_bigan_vanilla_cifar10}--\ref{fig:qualitative_bigan_z_adv_imagenet}. 
For each combination of model and dataset, we picked one training run and plotted the following:
\begin{itemize}[itemsep=1pt]
    \item Random model samples.
    \item Reconstructions of real images.
    \item Reconstructions of fake images. 
    \item Interpolations between embeddings of real images.
    \item Interpolations between embeddings of fake images.
\end{itemize}
Although it is difficult to make any definitive assessment based on a comparison of a small sample of images, we include these results for interested readers.

%% file: conclusion.tex
In this study, we find that the performance of BiGAN can be improved upon with the addition of an auto-encoding loss (BiGAN + $\X$-AE) as measured by the quality of samples and real image reconstruction. 
These improvements, however, come at the expense of an additional hyper-parameter that must be tuned. 
Thus this approach may only be viable when there are sufficient computational resources.

In addition, the BiGAN-like formulation requires a non-standard discriminator architecture which takes as input the tuple $(X, Z)$ rather than just $X$ in the typical GAN setting.
This deviation from the norm means that it may not be possible to take advantage of existing research into architectures proven to be useful in other areas of image modelling.
For instance, we were unable to find a modification to ResNet that could produce meaningful results with BiGAN or with any of the proposed BiGAN+ models.

The second finding of our study is, interestingly, that the simple GAN + $\X$-AE approach is a highly viable alternative for adding encoders to GANs, especially when a convolutional architecture for the discriminator is not suitable or when computational resources are limited.
Although the FID scores of reconstructed samples are inferior to those of other models, the sample quality and Inception L2 were amongst the best of all models on all datasets.
Furthermore, these models are no more difficult to train than normal GANs since the encoder can be trained separately \emph{after} the generator. This means that it is possible to take advantage of a large body of research into how to best train GANs.

We further find that that the additional adversarial losses perform no better than the autoencoder losses.
Indeed in this case simplicity works best.
Moreover BiGAN does not perform better than GAN + $\Z$-AE in terms of the metrics we considered. In fact GAN + $\Z$-AE dominated BiGAN under all metrics and on all datasets except \emph{Imagenet}.
This is a somewhat surprising finding, since the $\Z$-AE loss is the most basic way to derive an encoder for given a generator.

Finally we note that our evaluation focuses on the image domain, in particular the visual quality of image samples and the perceptual faithfulness of reconstructions. An alternative is to compare the proposed methods and their baselines by looking into the meaningfulness of the learned \emph{representation} given by the encoder, such as metrical properties and usefulness for downstream tasks. This is beyond the scope of this study, but may be revisited in future work.

%% file: appendix.tex
\subsection{Proofs of Propositions and Theorems.}\label{appendix:proofs}

The proofs of the propositions and theorems in Sections~\ref{subsubsec:adv-z} are essentially the same as the proofs presented in \cite{donahue2016adversarial}, but are reproduced here for completeness.

\subsubsection{Proof of Proposition~\ref{prop:adv-z-optimal-disc}}

\begin{proof}
For any measures $P$ and $Q$, the Radon-Nikodym derivatives $\frac{dP}{d(P+Q)}$ and $\frac{dQ}{d(P+Q)}$ exist and sum to 1.
Taking $P=P_{Z, G(Z)}$ and $Q=P_{E(G(Z)), G(Z)}$, we have that
\begin{align*}
    & L_{\Z}(D, G, E)  \\
    &= \E_{Z\sim P_Z}\left[-\log\left(D(G(Z), Z)\right)  - \log\left(1 - D(G(Z), E(G(Z)))\right)\right] \\
    &= -\E_{(Z,X)\sim P}\left[\log\left(D(X, Z)\right)\right] -   \E_{(Z,X)\sim Q}\left[\log\left(1 - D(X, Z)\right)\right] \\
    &= -2\E_{(Z,X)\sim (P+Q)/2}\left[\frac{dP}{d(P+Q)}(Z,X)\log\left(D(X, Z)\right)\right] \\
    & \qquad \qquad - 2\E_{(Z,X)\sim (P+Q)/2}\left[\frac{dQ}{d(P+Q)}(Z,X)\log\left(1 - D(X, Z)\right)\right] \\
    &= -2\E_{(Z,X)\sim (P+Q)/2}\left[\frac{dP}{d(P+Q)}(Z,X)\log\left(D(X, Z)\right) +   \left(1 - \frac{dP}{d(P+Q)}(Z,X) \right)\log\left(1 - D(X, Z)\right)\right]. \\
\end{align*}

The objective of $D$ is to maximise this quantity, which occurs when $D(X, Z) = \frac{dP}{d(P+Q)}(X,Z)$ since $\arg\max_y \{a \log(y) + (1-a)\log(1-y)\} = a$ for any $a\in[0,1]$.
\end{proof}

\subsubsection{Proof of Proposition~\ref{prop:adv-z-js}}

\begin{proof}
Again write $P=P_{Z, G(Z)}$ and $Q=P_{E(G(Z)), G(Z)}$, and let $f=\frac{dP}{d(P+Q)}$ and $g=\frac{dQ}{d(P+Q)}$.
Given an optimal discriminator, the remaining objective for the encoder and generator is

\begin{align*}
    C(G, E) &= -\E_{(Z,X)\sim P}\left[\log f(Z, X)\right] -   \E_{(Z,X)\sim Q}\left[\log g(Z, X)\right] \\
    &=-\E_{(Z,X)\sim P}\left[\log \frac{1}{2}\frac{dP}{d(P+Q)/2}(Z, X)\right] -   \E_{(Z,X)\sim Q}\left[\log \frac{1}{2}\frac{dQ}{d(P+Q)/2}(Z, X)\right] \\
    &=-D_{\text{KL}}[P || (P+Q)/2] -D_{\text{KL}}[Q || (P+Q)/2] + \log 4\\
    &=\log 4 - D_{\text{JS}}(P||Q).
\end{align*}
\end{proof}

\subsubsection{Proof of Theorem~\ref{thm:adv-z-e-equal-g}}

\begin{proof}
Given a fixed $G$, the optimal $E^*$ maximising $C(E,G)$ satisfies $C(E^*, G)=\log 4$, in which case $D_{\text{JS}}(P_{Z, G(Z)}|| P_{E(G(Z)), G(Z)}) = 0$. Since the Jensen-Shannon divergence is a distance between probability distributions, $D_{\text{JS}}$ being zero implies that $P_{Z, G(Z)} = P_{E(G(Z)), G(Z)}$ as measures.

The claim is that $E(G(z)) = z$ with probability one under $z\sim P_Z$. 
Consider the set $R = \{z\in\Z \ : \ E(G(z)) \not = z \}$. We will show that this set has measure zero under $P_Z$. 

Let $R_X = \{(z,x) \in \Z\times\X \ : \ x = G(z) \text{ and } z \in R \} $.

\begin{align*}
    P_Z(R) &= \int_{\Z} p_Z(z) \mathbb{1}_{\{z \in R\}} dz \\
    &= \int_{\Z} p_Z(z) \mathbb{1}_{\{z \in R\}} dz \\
    &= \int_{\Z} p_Z(z) \mathbb{1}_{\{(z,G(z)) \in R_X\}} dz \\
    &= P_{Z, G(Z)} \left( R_X\right) \\
    &= P_{E(G(Z)), G(Z)} \left( R_X\right) \\
    &= \int_{\X} p_{G(Z)}(x) \mathbb{1}_{\{(E(x),x) \in R_X\}} dx \\
    &= 0.
\end{align*}

The last line follows since $(E(x),x) \in R_X$ implies that $G(E(x)) = x$ and $G(E(x)) \not= x$, hence the set $\{(E(x),x) \in R_X\}$ is empty.
\end{proof}

\subsubsection{Proofs of Propositions~\ref{prop:adv-x-optimal-disc} and \ref{prop:adv-x-js}}

The proofs of Propositions~\ref{prop:adv-x-optimal-disc} and \ref{prop:adv-x-js} follow directly from the proofs of Propositions~\ref{prop:adv-z-optimal-disc} and \ref{prop:adv-z-js}, substituting $P=P_{Z, G(Z)}$ and $Q=P_{Z, G(E(G(Z))}$

\subsubsection{Proof of Theorem~\ref{thm:adv-x-e-equal-g}}

\begin{proof}
Given a fixed $G$, the optimal $E^*$ maximising $C(E,G)$ satisfies $C(E^*, G)=\log 4$, in which case $D_{\text{JS}}(P_{Z, G(Z)}|P_{Z, G(E(G(Z)))}) = 0$. Since the Jensen-Shannon divergence is distance on probability distributions, $D_{\text{JS}}$ being zero implies that $P_{Z, G(Z)} = P_{Z, G(E(G(Z)))}$ as measures. 

The claim is that $G(E(x)) = x$ with probability one under $x\sim P_G$. This is equivalent to the claim that $G(E(G(z))) = z$ with probability one under $z\sim P_Z$
Consider the set $R = \{z\in\Z \ : \ G(E(G(z))) \not = z \}$. We will show that this set has measure zero under $P_Z$. 

Let $R_X = \{(z,x) \in \Z\times\X \ : \ x = G(z) \text{ and } z \in R \} $.

\begin{align*}
    P_Z(R) &= \int_{\Z} p_Z(z) \mathbb{1}_{\{z \in R\}} dz \\
    &= \int_{\Z} p_Z(z) \mathbb{1}_{\{z \in R\}} dz \\
    &= \int_{\Z} p_Z(z) \mathbb{1}_{\{(z,G(z)) \in R_X\}} dz \\
    &= P_{Z, G(Z)} \left( R_X\right) \\
    &= P_{Z, G(E(G(Z)))} \left( R_X\right) \\
    &= \int_{\Z} p_{Z}(z) \mathbb{1}_{\{(z,G(E(G(z)))) \in R_X\}} dx \\
    &= 0.
\end{align*}

The last line follows since $(z,G(E(G(z)))) \in R_X$ implies that $G(E(G(z))) = G(z)$ and $G(E(G(z))) \not= G(z)$, hence the set $\{(z,G(E(G(z)))) \in R_X\}$ is empty.
\end{proof}

\subsection{All Models Tested}\label{appendix:models}
\textbf{GAN + $\X$ auto-encoder loss}

\begin{align*}
    D: \quad &\min_D \ \E_{X\sim P_X}\left[-\log(D(X)) \right] + \E_{Z \sim P_Z}\left[-\log(1 - D(G(Z))) \right] \\
    G: \quad &\min_{G} \ \E_{Z \sim P_Z}\left[-\log(D(G(Z))) \right] \\
    E: \quad &\min_{E} \ \E_{Z\sim P_Z} l_2\left(G(Z), G(E(G(Z))) \right)  \\
\end{align*}

\textbf{GAN + $\Z$ auto-encoder loss}
\begin{align*}
    D: \quad &\min_D \ \E_{X\sim P_X}\left[-\log(D(X)) \right] + \E_{Z \sim P_Z}\left[-\log(1 - D(G(Z))) \right] \\
    G: \quad &\min_{G} \ \E_{Z \sim P_Z}\left[-\log(D(G(Z))) \right] \\
    E: \quad &\min_{E} \ \E_{Z\sim P_Z} l_2\left(Z, E(G(Z)) \right)  \\
\end{align*}

\textbf{GAN + adversarial $\Z$ loss}
\begin{align*}
    D_{1}: \quad &\min_{D_{1}} \ \E_{X\sim P_X}\left[-\log(D_{1}(X)) \right] + \E_{Z \sim P_Z}\left[-\log(1 - D_{1}(G(Z))) \right] \\
    G: \quad &\min_{G} \ \E_{Z \sim P_Z}\left[-\log(D_{1}(G(Z))) \right] \\
    D_{2}: \quad &\min_{D_{2}} \ \E_{Z\sim P_Z}\left[-\log\left(D_{2}(G(Z), Z)\right)  - \log\left(1 - D_{2}(G(Z), E(G(Z)))\right)\right] \\
    E: \quad &\min_{E} \ \E_{Z\sim P_Z}\left[- \log\left(D_{2}(G(Z), E(G(Z)))\right)\right] \\
\end{align*}

\textbf{GAN + adversarial $\X$ loss}
\begin{align*}
    D_{1}: \quad &\min_{D_{1}} \ \E_{X\sim P_X}\left[-\log(D_{1}(X)) \right] + \E_{Z \sim P_Z}\left[-\log(1 - D_{1}(G(Z))) \right] \\
    G: \quad &\min_{G} \ \E_{Z \sim P_Z}\left[-\log(D_{1}(G(Z))) \right] \\
    D_{2}: \quad &\min_{D_{2}} \ \E_{Z\sim P_Z}\left[-\log\left(D_{2}(Z, G(Z))\right)  - \log\left(1 - D_{2}(Z, G(E(G(Z))))\right)\right] \\
    E: \quad &\min_{E} \ \E_{Z\sim P_Z}\left[ - \log\left(D_{2}(Z, G(E(G(Z))))\right)\right] \\
\end{align*}

\textbf{BiGAN}
\begin{align*}
    D: \quad &\min_D \ \E_{X\sim P_X}\left[-\log(D(X, E(X))) \right] + \E_{Z \sim P_Z}\left[-\log(1 - D(G(Z), Z)) \right] \\
    G: \quad &\min_{G} \ \E_{Z \sim P_Z}\left[-\log(D(G(Z), Z)) \right] \\
    E: \quad &\min_{E} \ \E_{X\sim P_X}\left[-\log( 1 - D(X, E(X))) \right] \\
\end{align*}

\textbf{BiGAN + $P_G$-$\X$ auto-encoder loss}
\begin{align*}
    D: \quad &\min_D \ \E_{X\sim P_X}\left[-\log(D(X, E(X))) \right] + \E_{Z \sim P_Z}\left[-\log(1 - D(G(Z), Z)) \right] \\
    G: \quad &\min_{G} \ \E_{Z \sim P_Z}\left[-\log(D(G(Z), Z)) \right] \\
    E: \quad &\min_{E} \ \E_{X\sim P_X}\left[-\log( 1 - D(X, E(X))) \right] + \lambda \E_{Z\sim P_Z} l_2\left(G(Z), G(E(G(Z))) \right) \\
\end{align*}

\textbf{BiGAN + $P_Z$-$\Z$ auto-encoder loss}
\begin{align*}
    D: \quad &\min_D \ \E_{X\sim P_X}\left[-\log(D(X, E(X))) \right] + \E_{Z \sim P_Z}\left[-\log(1 - D(G(Z), Z)) \right] \\
    G: \quad &\min_{G} \ \E_{Z \sim P_Z}\left[-\log(D(G(Z), Z)) \right] \\
    E: \quad &\min_{E} \ \E_{X\sim P_X}\left[-\log( 1 - D(X, E(X))) \right] + \lambda \E_{Z\sim P_Z} l_2\left(Z, E(G(Z)) \right) \\
\end{align*}

\textbf{BiGAN + adversarial $\Z$ loss}
\begin{align*}
    D_{1}: \quad &\min_{D_{1}} \ \E_{X\sim P_X}\left[-\log(D_{1}(X, E(X))) \right] + \E_{Z \sim P_Z}\left[-\log(1 - D_{1}(G(Z), Z)) \right] \\
    D_{2}: \quad &\min_{D_{2}} \ \E_{Z\sim P_Z}\left[-\log\left(D_{2}(G(Z), Z)\right)  - \log\left(1 - D_{2}(G(Z), E(G(Z)))\right)\right] \\
    G: \quad &\min_{G} \ \E_{Z \sim P_Z}\left[-\log(D_{1}(G(Z), Z)) \right] \\
    E: \quad &\min_{E} \ \E_{X\sim P_X}\left[-\log( 1 - D_{1}(X, E(X))) \right] + \lambda \E_{Z\sim P_Z}\left[- \log\left(D_{2}(G(Z), E(G(Z)))\right)\right] \\
\end{align*}

\textbf{BiGAN + adversarial $\X$ loss}
\begin{align*}
    D_{1}: \quad &\min_{D_{1}} \ \E_{X\sim P_X}\left[-\log(D_{1}(X, E(X))) \right] + \E_{Z \sim P_Z}\left[-\log(1 - D_{1}(G(Z), Z)) \right] \\
    D_{2}: \quad &\min_{D_{2}} \ \E_{Z\sim P_Z}\left[-\log\left(D_{2}(Z, G(Z))\right)  - \log\left(1 - D_{2}(Z, G(E(G(Z))))\right)\right] \\
    G: \quad &\min_{G} \ \E_{Z \sim P_Z}\left[-\log(D_{1}(G(Z), Z)) \right] \\
    E: \quad &\min_{E} \ \E_{X\sim P_X}\left[-\log( 1 - D_{1}(X, E(X))) \right] + \lambda \E_{Z\sim P_Z}\left[ - \log\left(D_{2}(Z, G(E(G(Z))))\right)\right] \\
\end{align*}

\textbf{VAE}

The VAEs we trained had a unit Gaussian prior $P_Z$, factored Gaussian posteriors and Gaussian decoders with covariance $\sigma^2 I$ where $\sigma$ is a learned parameter. In the following, the encoder maps $X$ to the mean and covariance of a Gaussian $\mu(X)$ and $\Sigma(X)$. Qualitative results from VAE models can be found in Appendix~\ref{appendix:qualitative-results}, but quantitative results are not present in Figures~\ref{fig:quant_results_gan_faded} and~\ref{fig:quant_results_bigan_faded} because the FID scores were significantly worse than all GAN and BiGAN models, obscuring the differences between other models when plotted on the same figure.

\begin{align*}
    G, E: \min_{G, E} \E_{X \sim P_X} \left[ - \E_{Z\sim \mathcal{N}(\mu(X), \Sigma(X))} \frac{1}{2\sigma^2} \left(X - G(Z)\right)^2 - d_Z \log(\sigma) + KL[\mathcal{N}(\mu(X), \Sigma(X)) || P_Z] \right].
\end{align*}

\subsection{Experimentation Details}\label{appendix:architectures}

In this section we list experimental details omitted for brevity from the main text.

All models were trained with batch size $64$ and for $500$K training steps. Checkpoints were saved every $10$K training steps for evaluation.

\subsubsection{Datasets}

The datasets we evaluated on were:

\medskip

\textbf{Cifar10:} Image dimensions are $32\times32\times3$. Latent dimensionality $d_{\mathcal{Z}} = 64$. 

\textbf{CelebA:} We used the centre-cropped and rescaled version of the dataset, so that all faces are aligned in the centre of the image. Image dimensions are $64\times64\times3$. Latent dimensionality $d_{\mathcal{Z}} = 128$. 

\textbf{Flintstones 64x64:} Image dimensions are $64\times64\times3$. Latent dimensionality $d_{\mathcal{Z}} = 128$. 

\textbf{ImageNet 64x64:} Image dimensions are $64\times64\times3$, obtained from original by area downsampling. Latent dimensionality $d_{\mathcal{Z}} = 256$. 

\medskip

\subsubsection{Architectures}

For all models we used architectures derived from the SN-DCGAN~\citep{miyato2018spectral} generator and discriminator architectures. For the encoders, we used the discriminator architecture with the final linear layer mapping to $d_{\mathcal{Z}}$ outputs rather than a single output. The discriminator was normalised with spectral normalisation \citep{miyato2018spectral} and its loss included the WGAN-GP penalty \citep{gulrajani2017improved}. The generator and encoder were normalised with layer normalisation \citep{ba2016layer}. 

In the following descriptions, $d_R$ is the resolution of the dataset and $d_Z$ is the size of the latent space. All architectures used were the same except for these two variables. 
Conv refers to a normal convolution. $\text{Conv}_{\text{SN}}$ refers to convolution with a spectral normalised filter \citep{miyato2018spectral}. FC stands for Fully Connected. LN stands for Layer Norm.

\textbf{Encoder.}
\begin{align*}
    x \in \mathbb{R}^{d_R \times d_R \times 3} &\to \text{Conv}(3\times 3\text{ kernel}, 64 \text{  filters}, 1\times1 \text{ stride})  \to \text{LN} \to \text{LeakyReLU}(\alpha=0.1) \\
    &\to  \text{Conv}(4\times 4\text{ kernel}, 128 \text{ filters}, 2\times2 \text{ stride}) \to \text{LN} \to \text{LeakyReLU}(\alpha=0.1) \\
    &\to  \text{Conv}(3\times 3\text{ kernel}, 128 \text{ filters}, 1\times1 \text{ stride})\to \text{LN}  \to \text{LeakyReLU}(\alpha=0.1)\\
    &\to  \text{Conv}(4\times 4\text{ kernel}, 256 \text{ filters}, 2\times2 \text{ stride})\to \text{LN}  \to \text{LeakyReLU}(\alpha=0.1)\\
    &\to  \text{Conv}(3\times 3\text{ kernel}, 256 \text{ filters}, 1\times1 \text{ stride})\to \text{LN}  \to \text{LeakyReLU}(\alpha=0.1)\\
    &\to  \text{Conv}(4\times 4\text{ kernel}, 512 \text{ filters}, 2\times2 \text{ stride})\to \text{LN}  \to \text{LeakyReLU}(\alpha=0.1)\\
    &\to  \text{Conv}(3\times 3\text{ kernel}, 512 \text{ filters}, 1\times1 \text{ stride})\to \text{LN}  \to \text{LeakyReLU}(\alpha=0.1)\\
    &\to \text{FC}(d_Z \text{ units})
\end{align*}

\textbf{Generator.} Notation: $d_k = d_R/k$. Outputs images with pixels in range $[0,1]$.
\begin{align*}
    \mathbb{R}^{d_Z} &\to \text{FC}\left(512 \times d_8 \times d_8 \text{ units}\right) \to \text{LN} \to \text{ReLU} \to \text{Reshape}(d_8 \times d_8 \times 512) \\
    & \to \text{TransposeConv}(d_4 \times d_4 \times 256 \text{ output shape}, 4\times4 \text{ kernel}, 2\times 2 \text{ stride}) \to \text{LN} \to \text{ReLU} \\
    & \to \text{TransposeConv}(d_2 \times d_2 \times 128 \text{ output shape}, 4\times4 \text{ kernel}, 2\times 2 \text{ stride}) \to \text{LN} \to \text{ReLU} \\
    & \to \text{TransposeConv}(d_1 \times d_1 \times 64 \text{ output shape}, 4\times4 \text{ kernel}, 2\times 2 \text{ stride}) \to \text{LN} \to \text{ReLU} \\
    & \to \text{TransposeConv}(d_1 \times d_1 \times 3 \text{ output shape}, 3\times3 \text{ kernel}, 1\times 1 \text{ stride}) \to (\text{tanh} + 1) / 2\\
\end{align*}

\textbf{Discriminator (GAN models).} Outputs logits.
\begin{align*}
    x \in \mathbb{R}^{d_R \times d_R \times 3} &\to \text{Conv}_{\text{SN}}(3\times 3\text{ kernel}, 64 \text{  filters}, 1\times1 \text{ stride}) \to \text{LeakyReLU}(\alpha=0.1) \\
    &\to \text{Conv}_{\text{SN}}(4\times 4\text{ kernel}, 128 \text{ filters}, 2\times2 \text{ stride}) \to \text{LeakyReLU}(\alpha=0.1)   \\
    &\to \text{Conv}_{\text{SN}}(3\times 3\text{ kernel}, 128 \text{ filters}, 1\times1 \text{ stride}) \to \text{LeakyReLU}(\alpha=0.1)  \\
    &\to \text{Conv}_{\text{SN}}(4\times 4\text{ kernel}, 256 \text{ filters}, 2\times2 \text{ stride}) \to \text{LeakyReLU}(\alpha=0.1)  \\
    &\to \text{Conv}_{\text{SN}}(3\times 3\text{ kernel}, 256 \text{ filters}, 1\times1 \text{ stride}) \to \text{LeakyReLU}(\alpha=0.1)  \\
    &\to \text{Conv}_{\text{SN}}(4\times 4\text{ kernel}, 512 \text{ filters}, 2\times2 \text{ stride}) \to \text{LeakyReLU}(\alpha=0.1)  \\
    &\to \text{Conv}_{\text{SN}}(3\times 3\text{ kernel}, 512 \text{ filters}, 1\times1 \text{ stride}) \to \text{LeakyReLU}(\alpha=0.1)  \\
    &\to \text{FC}(1 \text{ unit})
\end{align*}

\textbf{Discriminator (BiGAN models, $\X$- and $\Z$-adversarial losses).} Outputs logits. At each layer, we treat the latent code $Z$ as a $1 \times 1 \times d_Z$ tensor and convolve it with a (spectrally normalised) $1\times 1$ convolution and add it as a per-feature-bias, broadcasted across height and width dimensions. We denote by $\text{Conv}_{\text{SN}}(Z, n \text{ features})$ this convolution operation, where $n$ is the number of filters of the convolutional kernel. 
\begin{align*}
    &x \in \mathbb{R}^{d_R \times d_R \times 3} \\
    &\to \text{Conv}_{\text{SN}}(3\times 3\text{ kernel}, 64 \text{  filters}, 1\times1 \text{ stride}) + \text{Conv}_{\text{SN}}(Z, 64 \text{ features}) \to \text{LeakyReLU}(\alpha=0.1) \\
    &\to \text{Conv}_{\text{SN}}(4\times 4\text{ kernel}, 128 \text{ filters}, 2\times2 \text{ stride}) + \text{Conv}_{\text{SN}}(Z, 128 \text{ features}) \to \text{LeakyReLU}(\alpha=0.1)   \\
    &\to \text{Conv}_{\text{SN}}(3\times 3\text{ kernel}, 128 \text{ filters}, 1\times1 \text{ stride}) + \text{Conv}_{\text{SN}}(Z, 128 \text{ features}) \to \text{LeakyReLU}(\alpha=0.1)  \\
    &\to \text{Conv}_{\text{SN}}(4\times 4\text{ kernel}, 256 \text{ filters}, 2\times2 \text{ stride}) + \text{Conv}_{\text{SN}}(Z, 256 \text{ features}) \to \text{LeakyReLU}(\alpha=0.1)  \\
    &\to \text{Conv}_{\text{SN}}(3\times 3\text{ kernel}, 256 \text{ filters}, 1\times1 \text{ stride}) + \text{Conv}_{\text{SN}}(Z, 256 \text{ features}) \to \text{LeakyReLU}(\alpha=0.1)  \\
    &\to \text{Conv}_{\text{SN}}(4\times 4\text{ kernel}, 512 \text{ filters}, 2\times2 \text{ stride}) + \text{Conv}_{\text{SN}}(Z, 512 \text{ features}) \to \text{LeakyReLU}(\alpha=0.1)  \\
    &\to \text{Conv}_{\text{SN}}(3\times 3\text{ kernel}, 512 \text{ filters}, 1\times1 \text{ stride}) + \text{Conv}_{\text{SN}}(Z, 512 \text{ features}) \to \text{LeakyReLU}(\alpha=0.1)  \\
    &\to \text{FC}(1 \text{ unit})
\end{align*}

For GAN+$\X$-adversarial and GAN+$\Z$-adversarial we used two separate discriminators, one being the usual GAN discriminator and the other being for the additional adversarial loss. 
For BiGAN+$\X$-adversarial and BiGAN+$\Z$-adversarial, the two discriminators shared their parameters, since both act on the tuple $(X, Z)$. All parameters were shared except for the final fully connected layer.

\subsubsection{Optimisation and Hyper-parameters}

All optimisation was performed using the Adam optimiser \citep{kingma2014adam}. In all cases either two optimisers were used (one for the discriminator, one for the generator and encoder) or three were used (one each for the discriminator, generator and encoder).  
In all cases the same hyper-parameters were used for all optimisers. 

\medskip

The hyper-parameters common to all models that we searched over were: 

\begin{itemize}
    \item \textbf{Adam learning rate} in $\lbrace1\texttt{e}{-4}, 3\texttt{e}{-4}, 1\texttt{e}{-3}\rbrace$
    \item \textbf{WGAN-GP penalty weighting lambda} in $\lbrace 1, 3, 10\rbrace$
    \item \textbf{Discriminator updates per generator/encoder update} in $\lbrace 1, 2\rbrace$
\end{itemize}

\newpage

%% file: qualitative.tex
In the remaining pages we provide sample images from each model trained on each dataset.
We give samples of:

\begin{enumerate}[label=\alph*)]
    \item Real images.
    \item Random model samples.
    \item Real images and their corresponding reconstructions.
    \item Random model samples and their corresponding reconstructions.
    \item Interpolations between real images.
    \item Interpolations between random model samples.
\end{enumerate}

We additionally report $FID(P_X, P_G)$, $FID(P_X, P_{G(E(X))})$,  Inception $L_2(P_X, P_{G(E(X))})$ and optimisation hyper-parameters.

For each model and dataset, the particular training run displayed was chosen among all optimisation hyper-parameters by choosing the one with lowest $FID(P_X, P_{G(E(X))})$, except for cases in which the resulting model had very poor $FID(P_X, P_G)$. In such cases, we subjectively picked a model with a slightly worse $FID(P_X, P_{G(E(X))})$ but significantly better $FID(P_X, P_G)$. 


\begin{figure}
    \centering
    \textbf{Model: BiGAN. \  Dataset: Cifar10.}
    
    \vspace{2em}
    
    \begin{subfigure}{0.5\columnwidth}
        \centering
        \includegraphics[width=\columnwidth]{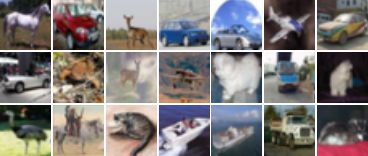}
        \caption{Real images.}
    \end{subfigure}%
    ~
    \begin{subfigure}{0.5\columnwidth}
        \centering
        \includegraphics[width=\columnwidth]{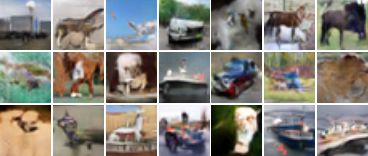}
        \caption{Random model samples.}
    \end{subfigure}%
    
    \vspace{2em}
    
    \begin{subfigure}{\columnwidth}
        \centering
        \includegraphics[width=\columnwidth]{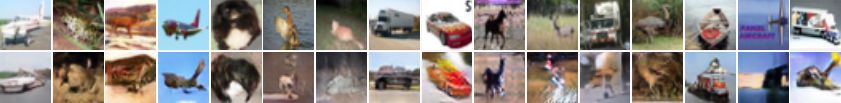}
        \caption{Real images (top row) and reconstructions (bottom row).}
    \end{subfigure}%
    
    \vspace{2em}
    
    \begin{subfigure}{\columnwidth}
        \centering
        \includegraphics[width=\columnwidth]{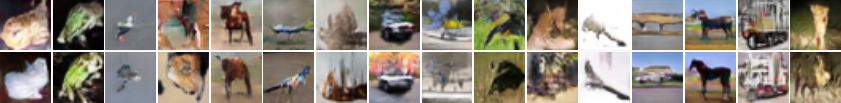}
        \caption{Random model samples (top row) and reconstructions (bottom row).}
    \end{subfigure}%
    
    \vspace{2em}
    
    \begin{subfigure}{0.5\columnwidth}
        \centering
        \includegraphics[width=\columnwidth]{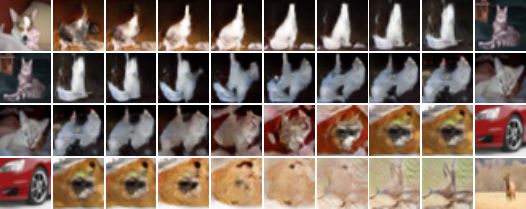}%
        \caption{Real sample interpolations.}
    \end{subfigure}%
    ~
    \begin{subfigure}{0.5\columnwidth}
        \centering
        \includegraphics[width=\columnwidth]{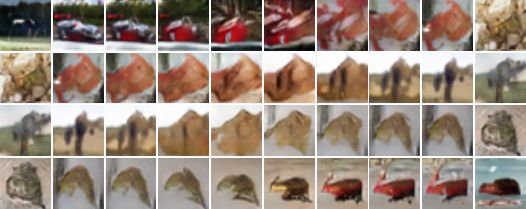}
        \caption{Random model sample interpolations.}
    \end{subfigure}%
    \caption{Qualitative results of training BiGAN on Cifar10. Interpolations are obtain by embedding two real/fake images $x_1$ and $x_2$ into the latent space with $E$, then decoding images uniformly along a line between these two embeddings $G\left(\alpha E(x_1) + (1-\alpha) E(x_2)\right)$. Left- and right-most images in each interpolation row are originals. \newline \newline
Statistics: $FID(P_X, P_G)=26.74$, \ $FID(P_X, P_{G(E(X))})=22.32$, \ Inception $L_2(P_X, P_{G(E(X))}) = 18.03$. \newline
Hyper-parameters: Adam learning rate 0.0003, $\beta_1=0.5$, $\beta_2=0.999$, 1 discriminator update(s) per generator update, WGAN-GP penalty weight $1.0$.
} 
    \label{fig:qualitative_bigan_vanilla_cifar10}
\end{figure}

\begin{figure}
    \centering
    \textbf{Model: BiGAN. \  Dataset: CelebA.}
    
    \vspace{2em}
    
    \begin{subfigure}{0.5\columnwidth}
        \centering
        \includegraphics[width=\columnwidth]{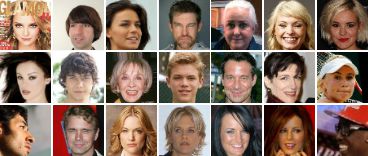}
        \caption{Real images.}
    \end{subfigure}%
    ~
    \begin{subfigure}{0.5\columnwidth}
        \centering
        \includegraphics[width=\columnwidth]{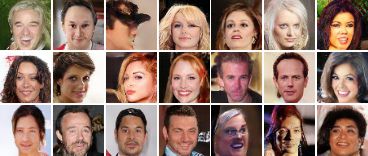}
        \caption{Random model samples.}
    \end{subfigure}%
    
    \vspace{2em}
    
    \begin{subfigure}{\columnwidth}
        \centering
        \includegraphics[width=\columnwidth]{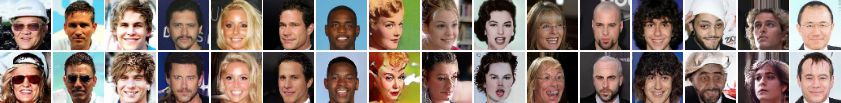}
        \caption{Real images (top row) and reconstructions (bottom row).}
    \end{subfigure}%
    
    \vspace{2em}
    
    \begin{subfigure}{\columnwidth}
        \centering
        \includegraphics[width=\columnwidth]{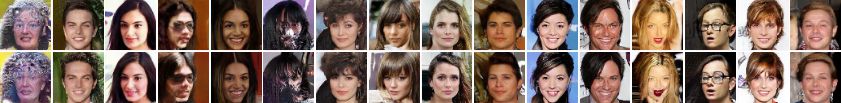}
        \caption{Random model samples (top row) and reconstructions (bottom row).}
    \end{subfigure}%
    
    \vspace{2em}
    
    \begin{subfigure}{0.5\columnwidth}
        \centering
        \includegraphics[width=\columnwidth]{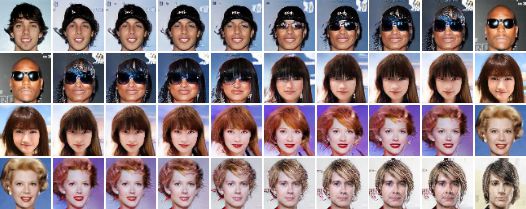}%
        \caption{Real sample interpolations.}
    \end{subfigure}%
    ~
    \begin{subfigure}{0.5\columnwidth}
        \centering
        \includegraphics[width=\columnwidth]{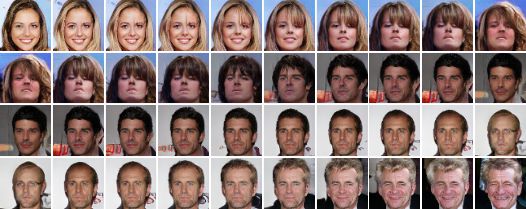}
        \caption{Random model sample interpolations.}
    \end{subfigure}%
    \caption{Qualitative results of training BiGAN on CelebA. Interpolations are obtain by embedding two real/fake images $x_1$ and $x_2$ into the latent space with $E$, then decoding images uniformly along a line between these two embeddings $G\left(\alpha E(x_1) + (1-\alpha) E(x_2)\right)$. Left- and right-most images in each interpolation row are originals. \newline \newline
Statistics: $FID(P_X, P_G)=7.64$, \ $FID(P_X, P_{G(E(X))})=6.34$, \ Inception $L_2(P_X, P_{G(E(X))}) = 11.25$. \newline
Hyper-parameters: Adam learning rate 0.0003, $\beta_1=0.5$, $\beta_2=0.999$, 2 discriminator update(s) per generator update, WGAN-GP penalty weight $1.0$.
} 
    \label{fig:qualitative_bigan_vanilla_celeba}
\end{figure}

\begin{figure}
    \centering
    \textbf{Model: BiGAN. \  Dataset: Flintstones.}
    
    \vspace{2em}
    
    \begin{subfigure}{0.5\columnwidth}
        \centering
        \includegraphics[width=\columnwidth]{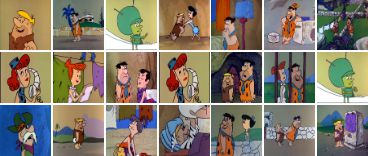}
        \caption{Real images.}
    \end{subfigure}%
    ~
    \begin{subfigure}{0.5\columnwidth}
        \centering
        \includegraphics[width=\columnwidth]{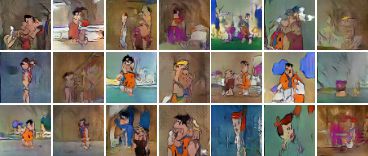}
        \caption{Random model samples.}
    \end{subfigure}%
    
    \vspace{2em}
    
    \begin{subfigure}{\columnwidth}
        \centering
        \includegraphics[width=\columnwidth]{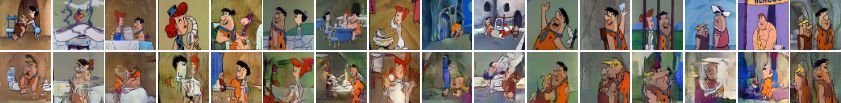}
        \caption{Real images (top row) and reconstructions (bottom row).}
    \end{subfigure}%
    
    \vspace{2em}
    
    \begin{subfigure}{\columnwidth}
        \centering
        \includegraphics[width=\columnwidth]{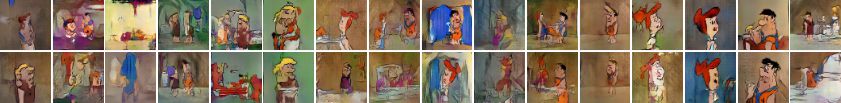}
        \caption{Random model samples (top row) and reconstructions (bottom row).}
    \end{subfigure}%
    
    \vspace{2em}
    
    \begin{subfigure}{0.5\columnwidth}
        \centering
        \includegraphics[width=\columnwidth]{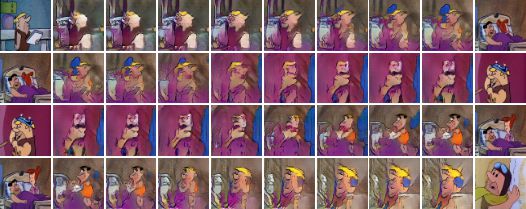}%
        \caption{Real sample interpolations.}
    \end{subfigure}%
    ~
    \begin{subfigure}{0.5\columnwidth}
        \centering
        \includegraphics[width=\columnwidth]{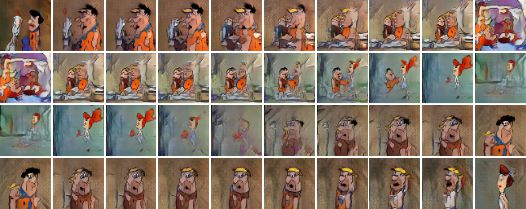}
        \caption{Random model sample interpolations.}
    \end{subfigure}%
    \caption{Qualitative results of training BiGAN on Flintstones. Interpolations are obtain by embedding two real/fake images $x_1$ and $x_2$ into the latent space with $E$, then decoding images uniformly along a line between these two embeddings $G\left(\alpha E(x_1) + (1-\alpha) E(x_2)\right)$. Left- and right-most images in each interpolation row are originals.\newline \newline
Statistics: $FID(P_X, P_G)=97.53$, \ $FID(P_X, P_{G(E(X))})=102.91$, \ Inception $L_2(P_X, P_{G(E(X))}) = 16.96$. \newline
Hyper-parameters: Adam learning rate 0.0003, $\beta_1=0.5$, $\beta_2=0.999$, 1 discriminator update(s) per generator update, WGAN-GP penalty weight $10.0$.} 
    \label{fig:qualitative_bigan_vanilla_flintstones}
\end{figure}

\begin{figure}
    \centering
    \textbf{Model: BiGAN. \  Dataset: ImageNet.}
    
    \vspace{2em}
    
    \begin{subfigure}{0.5\columnwidth}
        \centering
        \includegraphics[width=\columnwidth]{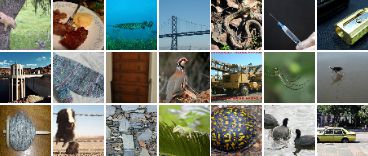}
        \caption{Real images.}
    \end{subfigure}%
    ~
    \begin{subfigure}{0.5\columnwidth}
        \centering
        \includegraphics[width=\columnwidth]{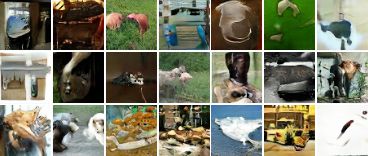}
        \caption{Random model samples.}
    \end{subfigure}%
    
    \vspace{2em}
    
    \begin{subfigure}{\columnwidth}
        \centering
        \includegraphics[width=\columnwidth]{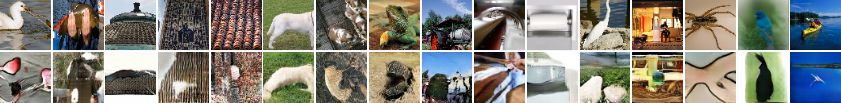}
        \caption{Real images (top row) and reconstructions (bottom row).}
    \end{subfigure}%
    
    \vspace{2em}
    
    \begin{subfigure}{\columnwidth}
        \centering
        \includegraphics[width=\columnwidth]{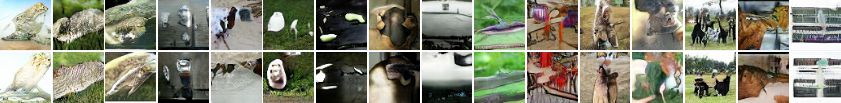}
        \caption{Random model samples (top row) and reconstructions (bottom row).}
    \end{subfigure}%
    
    \vspace{2em}
    
    \begin{subfigure}{0.5\columnwidth}
        \centering
        \includegraphics[width=\columnwidth]{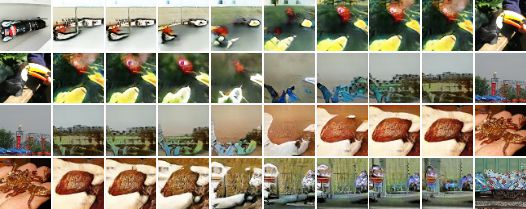}%
        \caption{Real sample interpolations.}
    \end{subfigure}%
    ~
    \begin{subfigure}{0.5\columnwidth}
        \centering
        \includegraphics[width=\columnwidth]{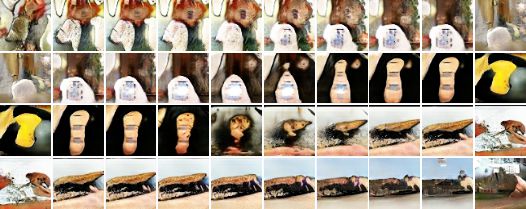}
        \caption{Random model sample interpolations.}
    \end{subfigure}%
    \caption{Qualitative results of training BiGAN on ImageNet. Interpolations are obtain by embedding two real/fake images $x_1$ and $x_2$ into the latent space with $E$, then decoding images uniformly along a line between these two embeddings $G\left(\alpha E(x_1) + (1-\alpha) E(x_2)\right)$. Left- and right-most images in each interpolation row are originals. \newline \newline
Statistics: $FID(P_X, P_G)=59.95$, \ $FID(P_X, P_{G(E(X))})=52.70$, \ Inception $L_2(P_X, P_{G(E(X))}) = 19.58$. \newline
Hyper-parameters: Adam learning rate 0.0001, $\beta_1=0.5$, $\beta_2=0.999$, 1 discriminator update(s) per generator update, WGAN-GP penalty weight $3.0$.} 
    \label{fig:qualitative_bigan_vanilla_imagenet}
\end{figure}


\begin{figure}
    \centering
    \textbf{Model: GAN + X Auto-encoder. \  Dataset: Cifar10.}
    
    \vspace{2em}
    
    \begin{subfigure}{0.5\columnwidth}
        \centering
        \includegraphics[width=\columnwidth]{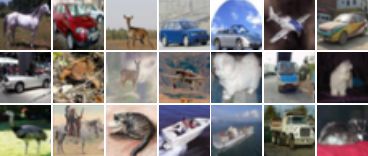}
        \caption{Real images.}
    \end{subfigure}%
    ~
    \begin{subfigure}{0.5\columnwidth}
        \centering
        \includegraphics[width=\columnwidth]{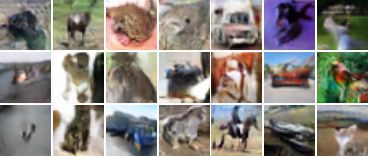}
        \caption{Random model samples.}
    \end{subfigure}%
    
    \vspace{2em}
    
    \begin{subfigure}{\columnwidth}
        \centering
        \includegraphics[width=\columnwidth]{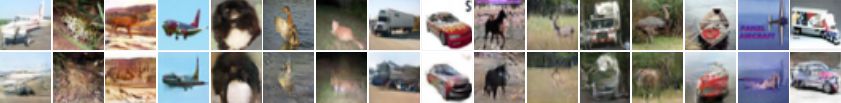}
        \caption{Real images (top row) and reconstructions (bottom row).}
    \end{subfigure}%
    
    \vspace{2em}
    
    \begin{subfigure}{\columnwidth}
        \centering
        \includegraphics[width=\columnwidth]{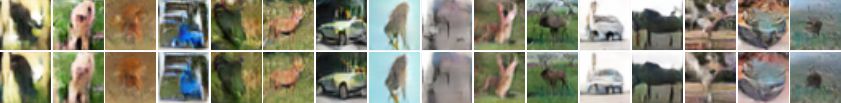}
        \caption{Random model samples (top row) and reconstructions (bottom row).}
    \end{subfigure}%
    
    \vspace{2em}
    
    \begin{subfigure}{0.5\columnwidth}
        \centering
        \includegraphics[width=\columnwidth]{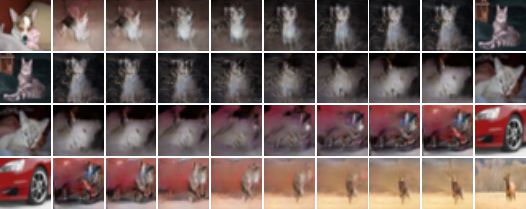}%
        \caption{Real sample interpolations.}
    \end{subfigure}%
    ~
    \begin{subfigure}{0.5\columnwidth}
        \centering
        \includegraphics[width=\columnwidth]{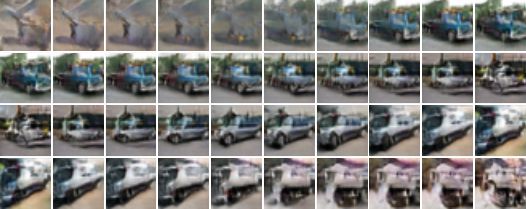}
        \caption{Random model sample interpolations.}
    \end{subfigure}%
    \caption{Qualitative results of training BiGAN on Cifar10. Interpolations are obtain by embedding two real/fake images $x_1$ and $x_2$ into the latent space with $E$, then decoding images uniformly along a line between these two embeddings $G\left(\alpha E(x_1) + (1-\alpha) E(x_2)\right)$. Left- and right-most images in each interpolation row are originals. \newline \newline
Statistics: $FID(P_X, P_G)=29.92$, \ $FID(P_X, P_{G(E(X))})=30.05$, \ Inception $L_2(P_X, P_{G(E(X))}) = 16.39$. \newline
Hyper-parameters: Adam learning rate 0.0001, $\beta_1=0.5$, $\beta_2=0.999$, 1 discriminator update(s) per generator update, WGAN-GP penalty weight $3.0$.
} 
    \label{fig:qualitative_gan_x_ae_cifar10}
\end{figure}

\begin{figure}
    \centering
    \textbf{Model: GAN + X Auto-encoder. \  Dataset: CelebA.}
    
    \vspace{2em}
    
    \begin{subfigure}{0.5\columnwidth}
        \centering
        \includegraphics[width=\columnwidth]{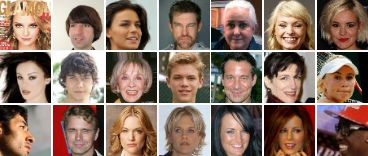}
        \caption{Real images.}
    \end{subfigure}%
    ~
    \begin{subfigure}{0.5\columnwidth}
        \centering
        \includegraphics[width=\columnwidth]{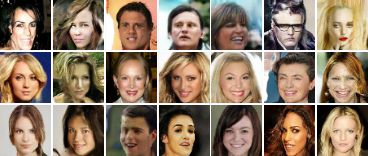}
        \caption{Random model samples.}
    \end{subfigure}%
    
    \vspace{2em}
    
    \begin{subfigure}{\columnwidth}
        \centering
        \includegraphics[width=\columnwidth]{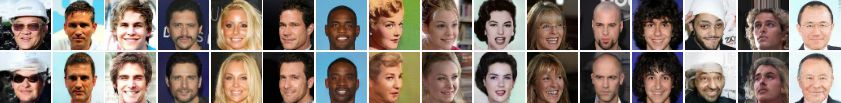}
        \caption{Real images (top row) and reconstructions (bottom row).}
    \end{subfigure}%
    
    \vspace{2em}
    
    \begin{subfigure}{\columnwidth}
        \centering
        \includegraphics[width=\columnwidth]{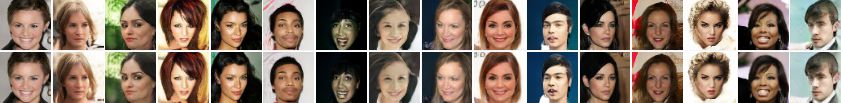}
        \caption{Random model samples (top row) and reconstructions (bottom row).}
    \end{subfigure}%
    
    \vspace{2em}
    
    \begin{subfigure}{0.5\columnwidth}
        \centering
        \includegraphics[width=\columnwidth]{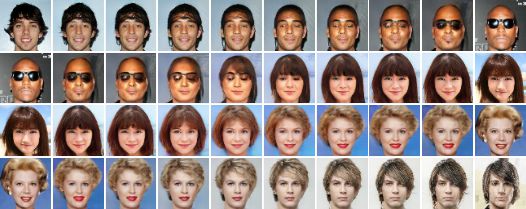}%
        \caption{Real sample interpolations.}
    \end{subfigure}%
    ~
    \begin{subfigure}{0.5\columnwidth}
        \centering
        \includegraphics[width=\columnwidth]{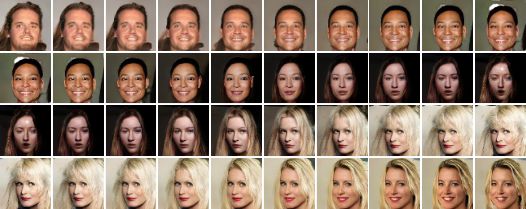}
        \caption{Random model sample interpolations.}
    \end{subfigure}%
    \caption{Qualitative results of training BiGAN on CelebA. Interpolations are obtain by embedding two real/fake images $x_1$ and $x_2$ into the latent space with $E$, then decoding images uniformly along a line between these two embeddings $G\left(\alpha E(x_1) + (1-\alpha) E(x_2)\right)$. Left- and right-most images in each interpolation row are originals. \newline \newline
Statistics: $FID(P_X, P_G)=7.96$, \ $FID(P_X, P_{G(E(X))})=12.35$, \ Inception $L_2(P_X, P_{G(E(X))}) = 10.22$. \newline
Hyper-parameters: Adam learning rate 0.0003, $\beta_1=0.5$, $\beta_2=0.999$, 2 discriminator update(s) per generator update, WGAN-GP penalty weight $1.0$.
} 
    \label{fig:qualitative_gan_x_ae_celeba}
\end{figure}

\begin{figure}
    \centering
    \textbf{Model: GAN + X Auto-encoder. \  Dataset: Flintstones.}
    
    \vspace{2em}
    
    \begin{subfigure}{0.5\columnwidth}
        \centering
        \includegraphics[width=\columnwidth]{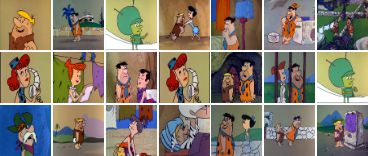}
        \caption{Real images.}
    \end{subfigure}%
    ~
    \begin{subfigure}{0.5\columnwidth}
        \centering
        \includegraphics[width=\columnwidth]{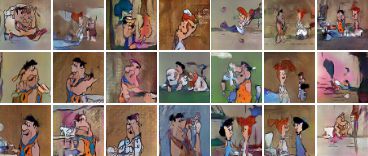}
        \caption{Random model samples.}
    \end{subfigure}%
    
    \vspace{2em}
    
    \begin{subfigure}{\columnwidth}
        \centering
        \includegraphics[width=\columnwidth]{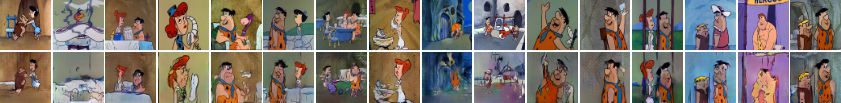}
        \caption{Real images (top row) and reconstructions (bottom row).}
    \end{subfigure}%
    
    \vspace{2em}
    
    \begin{subfigure}{\columnwidth}
        \centering
        \includegraphics[width=\columnwidth]{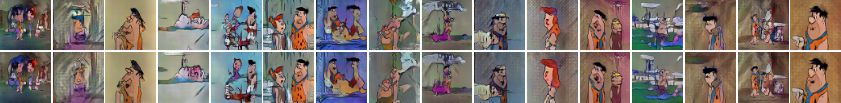}
        \caption{Random model samples (top row) and reconstructions (bottom row).}
    \end{subfigure}%
    
    \vspace{2em}
    
    \begin{subfigure}{0.5\columnwidth}
        \centering
        \includegraphics[width=\columnwidth]{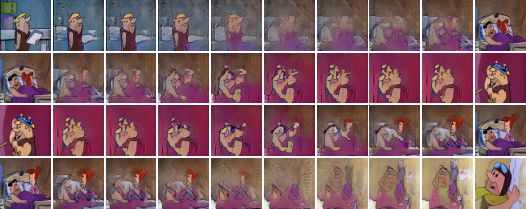}%
        \caption{Real sample interpolations.}
    \end{subfigure}%
    ~
    \begin{subfigure}{0.5\columnwidth}
        \centering
        \includegraphics[width=\columnwidth]{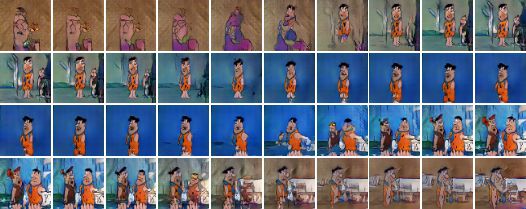}
        \caption{Random model sample interpolations.}
    \end{subfigure}%
    \caption{Qualitative results of training BiGAN on Flintstones. Interpolations are obtain by embedding two real/fake images $x_1$ and $x_2$ into the latent space with $E$, then decoding images uniformly along a line between these two embeddings $G\left(\alpha E(x_1) + (1-\alpha) E(x_2)\right)$. Left- and right-most images in each interpolation row are originals.\newline \newline
Statistics: $FID(P_X, P_G)=84.65$, \ $FID(P_X, P_{G(E(X))})=89.19$, \ Inception $L_2(P_X, P_{G(E(X))}) = 15.53$. \newline
Hyper-parameters: Adam learning rate 0.0001, $\beta_1=0.5$, $\beta_2=0.999$, 2 discriminator update(s) per generator update, WGAN-GP penalty weight $3.0$.} 
    \label{fig:qualitative_gan_x_ae_flintstones}
\end{figure}

\begin{figure}
    \centering
    \textbf{Model: GAN + X Auto-encoder. \  Dataset: ImageNet.}
    
    \vspace{2em}
    
    \begin{subfigure}{0.5\columnwidth}
        \centering
        \includegraphics[width=\columnwidth]{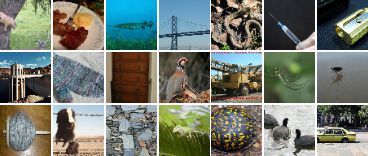}
        \caption{Real images.}
    \end{subfigure}%
    ~
    \begin{subfigure}{0.5\columnwidth}
        \centering
        \includegraphics[width=\columnwidth]{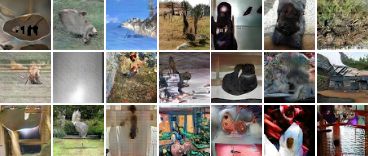}
        \caption{Random model samples.}
    \end{subfigure}%
    
    \vspace{2em}
    
    \begin{subfigure}{\columnwidth}
        \centering
        \includegraphics[width=\columnwidth]{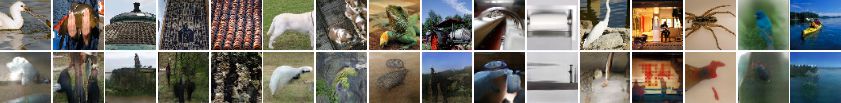}
        \caption{Real images (top row) and reconstructions (bottom row).}
    \end{subfigure}%
    
    \vspace{2em}
    
    \begin{subfigure}{\columnwidth}
        \centering
        \includegraphics[width=\columnwidth]{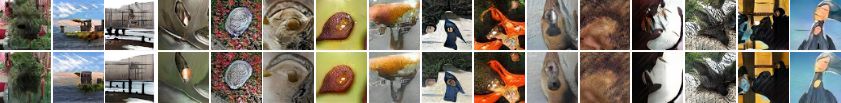}
        \caption{Random model samples (top row) and reconstructions (bottom row).}
    \end{subfigure}%
    
    \vspace{2em}
    
    \begin{subfigure}{0.5\columnwidth}
        \centering
        \includegraphics[width=\columnwidth]{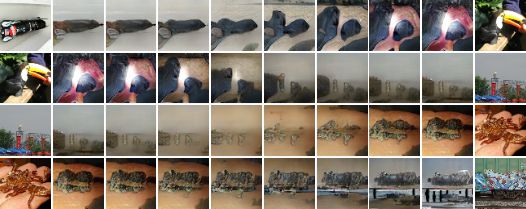}%
        \caption{Real sample interpolations.}
    \end{subfigure}%
    ~
    \begin{subfigure}{0.5\columnwidth}
        \centering
        \includegraphics[width=\columnwidth]{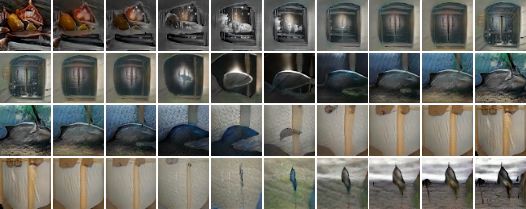}
        \caption{Random model sample interpolations.}
    \end{subfigure}%
    \caption{Qualitative results of training BiGAN on ImageNet. Interpolations are obtain by embedding two real/fake images $x_1$ and $x_2$ into the latent space with $E$, then decoding images uniformly along a line between these two embeddings $G\left(\alpha E(x_1) + (1-\alpha) E(x_2)\right)$. Left- and right-most images in each interpolation row are originals. \newline \newline
Statistics: $FID(P_X, P_G)=53.71$, \ $FID(P_X, P_{G(E(X))})=65.02$, \ Inception $L_2(P_X, P_{G(E(X))}) = 19.33$. \newline
Hyper-parameters: Adam learning rate 0.0003, $\beta_1=0.5$, $\beta_2=0.999$, 1 discriminator update(s) per generator update, WGAN-GP penalty weight $1.0$.} 
    \label{fig:qualitative_gan_x_ae_imagenet}
\end{figure}


\begin{figure}
    \centering
    \textbf{Model: GAN + Z Auto-encoder. \  Dataset: Cifar10.}
    
    \vspace{2em}
    
    \begin{subfigure}{0.5\columnwidth}
        \centering
        \includegraphics[width=\columnwidth]{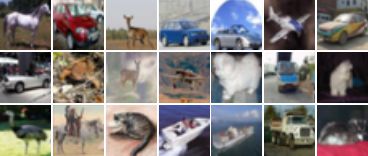}
        \caption{Real images.}
    \end{subfigure}%
    ~
    \begin{subfigure}{0.5\columnwidth}
        \centering
        \includegraphics[width=\columnwidth]{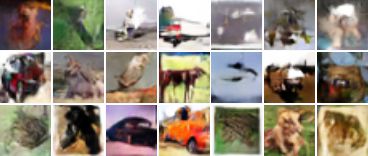}
        \caption{Random model samples.}
    \end{subfigure}%
    
    \vspace{2em}
    
    \begin{subfigure}{\columnwidth}
        \centering
        \includegraphics[width=\columnwidth]{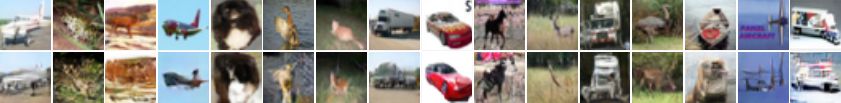}
        \caption{Real images (top row) and reconstructions (bottom row).}
    \end{subfigure}%
    
    \vspace{2em}
    
    \begin{subfigure}{\columnwidth}
        \centering
        \includegraphics[width=\columnwidth]{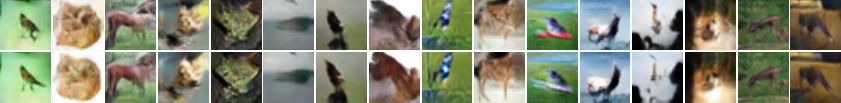}
        \caption{Random model samples (top row) and reconstructions (bottom row).}
    \end{subfigure}%
    
    \vspace{2em}
    
    \begin{subfigure}{0.5\columnwidth}
        \centering
        \includegraphics[width=\columnwidth]{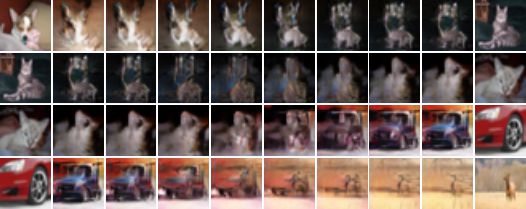}%
        \caption{Real sample interpolations.}
    \end{subfigure}%
    ~
    \begin{subfigure}{0.5\columnwidth}
        \centering
        \includegraphics[width=\columnwidth]{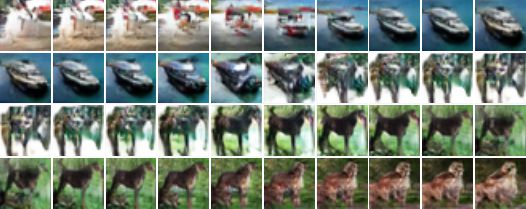}
        \caption{Random model sample interpolations.}
    \end{subfigure}%
    \caption{Qualitative results of training BiGAN on Cifar10. Interpolations are obtain by embedding two real/fake images $x_1$ and $x_2$ into the latent space with $E$, then decoding images uniformly along a line between these two embeddings $G\left(\alpha E(x_1) + (1-\alpha) E(x_2)\right)$. Left- and right-most images in each interpolation row are originals. \newline \newline
Statistics: $FID(P_X, P_G)=31.20$, \ $FID(P_X, P_{G(E(X))})=28.72$, \ Inception $L_2(P_X, P_{G(E(X))}) = 16.66$. \newline
Hyper-parameters: Adam learning rate 0.0001, $\beta_1=0.5$, $\beta_2=0.999$, 1 discriminator update(s) per generator update, WGAN-GP penalty weight $10.0$.
} 
    \label{fig:qualitative_gan_z_ae_cifar10}
\end{figure}

\begin{figure}
    \centering
    \textbf{Model: GAN + Z Auto-encoder. \  Dataset: CelebA.}
    
    \vspace{2em}
    
    \begin{subfigure}{0.5\columnwidth}
        \centering
        \includegraphics[width=\columnwidth]{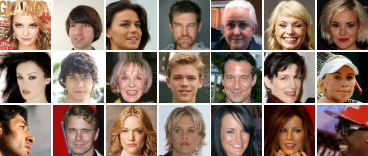}
        \caption{Real images.}
    \end{subfigure}%
    ~
    \begin{subfigure}{0.5\columnwidth}
        \centering
        \includegraphics[width=\columnwidth]{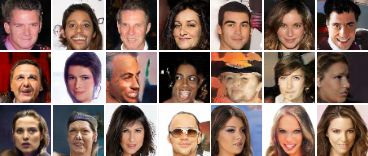}
        \caption{Random model samples.}
    \end{subfigure}%
    
    \vspace{2em}
    
    \begin{subfigure}{\columnwidth}
        \centering
        \includegraphics[width=\columnwidth]{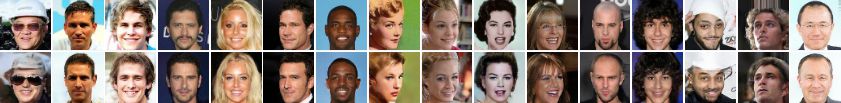}
        \caption{Real images (top row) and reconstructions (bottom row).}
    \end{subfigure}%
    
    \vspace{2em}
    
    \begin{subfigure}{\columnwidth}
        \centering
        \includegraphics[width=\columnwidth]{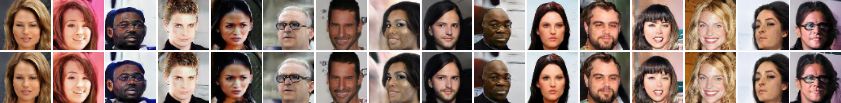}
        \caption{Random model samples (top row) and reconstructions (bottom row).}
    \end{subfigure}%
    
    \vspace{2em}
    
    \begin{subfigure}{0.5\columnwidth}
        \centering
        \includegraphics[width=\columnwidth]{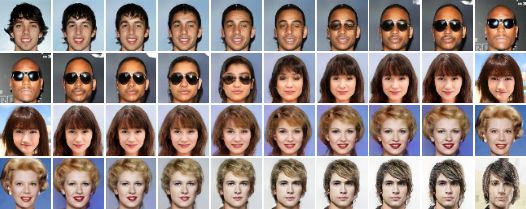}%
        \caption{Real sample interpolations.}
    \end{subfigure}%
    ~
    \begin{subfigure}{0.5\columnwidth}
        \centering
        \includegraphics[width=\columnwidth]{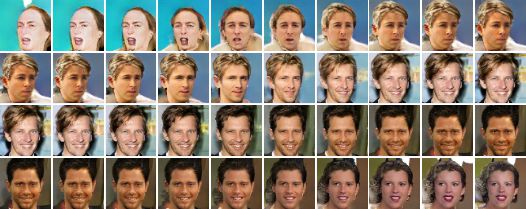}
        \caption{Random model sample interpolations.}
    \end{subfigure}%
    \caption{Qualitative results of training BiGAN on CelebA. Interpolations are obtain by embedding two real/fake images $x_1$ and $x_2$ into the latent space with $E$, then decoding images uniformly along a line between these two embeddings $G\left(\alpha E(x_1) + (1-\alpha) E(x_2)\right)$. Left- and right-most images in each interpolation row are originals. \newline \newline
Statistics: $FID(P_X, P_G)=7.12$, \ $FID(P_X, P_{G(E(X))})=14.21$, \ Inception $L_2(P_X, P_{G(E(X))}) = 10.53$. \newline
Hyper-parameters: Adam learning rate 0.001, $\beta_1=0.5$, $\beta_2=0.999$, 2 discriminator update(s) per generator update, WGAN-GP penalty weight $3.0$.
} 
    \label{fig:qualitative_gan_z_ae_celeba}
\end{figure}

\begin{figure}
    \centering
    \textbf{Model: GAN + Z Auto-encoder. \  Dataset: Flintstones.}
    
    \vspace{2em}
    
    \begin{subfigure}{0.5\columnwidth}
        \centering
        \includegraphics[width=\columnwidth]{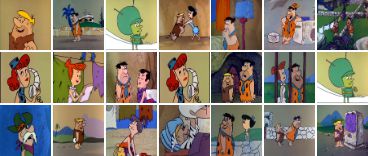}
        \caption{Real images.}
    \end{subfigure}%
    ~
    \begin{subfigure}{0.5\columnwidth}
        \centering
        \includegraphics[width=\columnwidth]{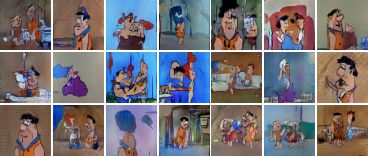}
        \caption{Random model samples.}
    \end{subfigure}%
    
    \vspace{2em}
    
    \begin{subfigure}{\columnwidth}
        \centering
        \includegraphics[width=\columnwidth]{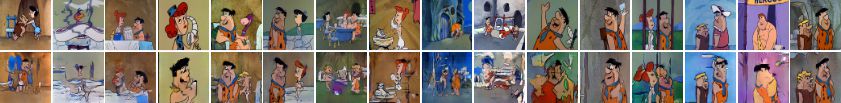}
        \caption{Real images (top row) and reconstructions (bottom row).}
    \end{subfigure}%
    
    \vspace{2em}
    
    \begin{subfigure}{\columnwidth}
        \centering
        \includegraphics[width=\columnwidth]{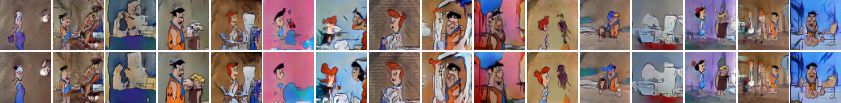}
        \caption{Random model samples (top row) and reconstructions (bottom row).}
    \end{subfigure}%
    
    \vspace{2em}
    
    \begin{subfigure}{0.5\columnwidth}
        \centering
        \includegraphics[width=\columnwidth]{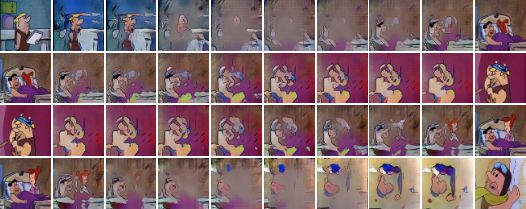}%
        \caption{Real sample interpolations.}
    \end{subfigure}%
    ~
    \begin{subfigure}{0.5\columnwidth}
        \centering
        \includegraphics[width=\columnwidth]{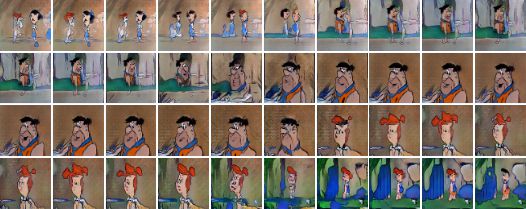}
        \caption{Random model sample interpolations.}
    \end{subfigure}%
    \caption{Qualitative results of training BiGAN on Flintstones. Interpolations are obtain by embedding two real/fake images $x_1$ and $x_2$ into the latent space with $E$, then decoding images uniformly along a line between these two embeddings $G\left(\alpha E(x_1) + (1-\alpha) E(x_2)\right)$. Left- and right-most images in each interpolation row are originals.\newline \newline
Statistics: $FID(P_X, P_G)=80.32$, \ $FID(P_X, P_{G(E(X))})=85.90$, \ Inception $L_2(P_X, P_{G(E(X))}) = 15.54$. \newline
Hyper-parameters: Adam learning rate 0.0003, $\beta_1=0.5$, $\beta_2=0.999$, 1 discriminator update(s) per generator update, WGAN-GP penalty weight $3.0$.} 
    \label{fig:qualitative_gan_z_ae_flintstones}
\end{figure}

\begin{figure}
    \centering
    \textbf{Model: GAN + Z Auto-encoder. \  Dataset: ImageNet.}
    
    \vspace{2em}
    
    \begin{subfigure}{0.5\columnwidth}
        \centering
        \includegraphics[width=\columnwidth]{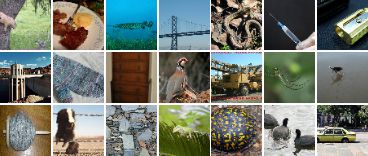}
        \caption{Real images.}
    \end{subfigure}%
    ~
    \begin{subfigure}{0.5\columnwidth}
        \centering
        \includegraphics[width=\columnwidth]{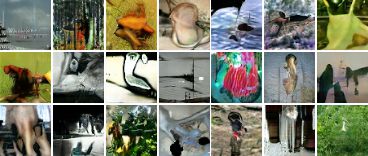}
        \caption{Random model samples.}
    \end{subfigure}%
    
    \vspace{2em}
    
    \begin{subfigure}{\columnwidth}
        \centering
        \includegraphics[width=\columnwidth]{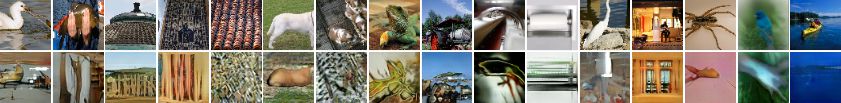}
        \caption{Real images (top row) and reconstructions (bottom row).}
    \end{subfigure}%
    
    \vspace{2em}
    
    \begin{subfigure}{\columnwidth}
        \centering
        \includegraphics[width=\columnwidth]{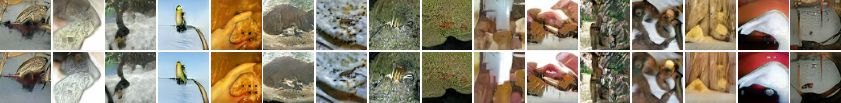}
        \caption{Random model samples (top row) and reconstructions (bottom row).}
    \end{subfigure}%
    
    \vspace{2em}
    
    \begin{subfigure}{0.5\columnwidth}
        \centering
        \includegraphics[width=\columnwidth]{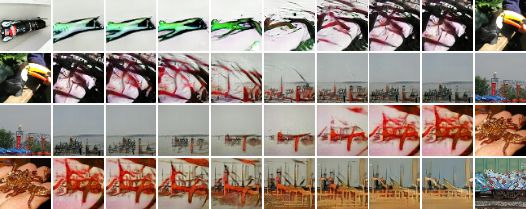}%
        \caption{Real sample interpolations.}
    \end{subfigure}%
    ~
    \begin{subfigure}{0.5\columnwidth}
        \centering
        \includegraphics[width=\columnwidth]{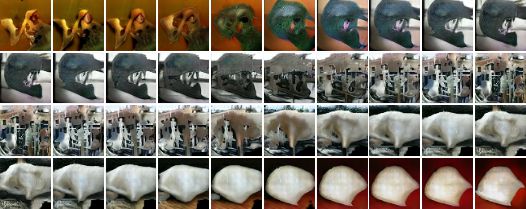}
        \caption{Random model sample interpolations.}
    \end{subfigure}%
    \caption{Qualitative results of training BiGAN on ImageNet. Interpolations are obtain by embedding two real/fake images $x_1$ and $x_2$ into the latent space with $E$, then decoding images uniformly along a line between these two embeddings $G\left(\alpha E(x_1) + (1-\alpha) E(x_2)\right)$. Left- and right-most images in each interpolation row are originals. \newline \newline
Statistics: $FID(P_X, P_G)=61.73$, \ $FID(P_X, P_{G(E(X))})=55.41$, \ Inception $L_2(P_X, P_{G(E(X))}) = 19.24$. \newline
Hyper-parameters: Adam learning rate 0.0003, $\beta_1=0.5$, $\beta_2=0.999$, 2 discriminator update(s) per generator update, WGAN-GP penalty weight $10.0$.} 
    \label{fig:qualitative_gan_z_ae_imagenet}
\end{figure}


\begin{figure}
    \centering
    \textbf{Model: GAN + X adversarial loss. \  Dataset: Cifar10.}
    
    \vspace{2em}
    
    \begin{subfigure}{0.5\columnwidth}
        \centering
        \includegraphics[width=\columnwidth]{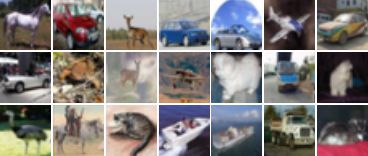}
        \caption{Real images.}
    \end{subfigure}%
    ~
    \begin{subfigure}{0.5\columnwidth}
        \centering
        \includegraphics[width=\columnwidth]{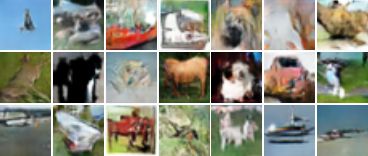}
        \caption{Random model samples.}
    \end{subfigure}%
    
    \vspace{2em}
    
    \begin{subfigure}{\columnwidth}
        \centering
        \includegraphics[width=\columnwidth]{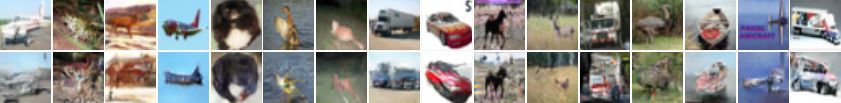}
        \caption{Real images (top row) and reconstructions (bottom row).}
    \end{subfigure}%
    
    \vspace{2em}
    
    \begin{subfigure}{\columnwidth}
        \centering
        \includegraphics[width=\columnwidth]{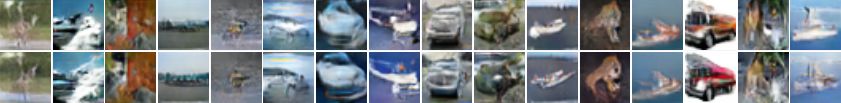}
        \caption{Random model samples (top row) and reconstructions (bottom row).}
    \end{subfigure}%
    
    \vspace{2em}
    
    \begin{subfigure}{0.5\columnwidth}
        \centering
        \includegraphics[width=\columnwidth]{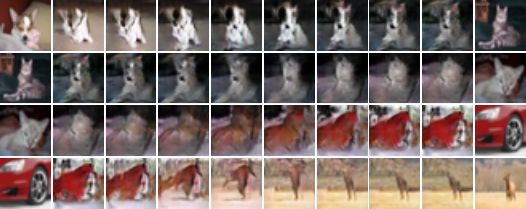}%
        \caption{Real sample interpolations.}
    \end{subfigure}%
    ~
    \begin{subfigure}{0.5\columnwidth}
        \centering
        \includegraphics[width=\columnwidth]{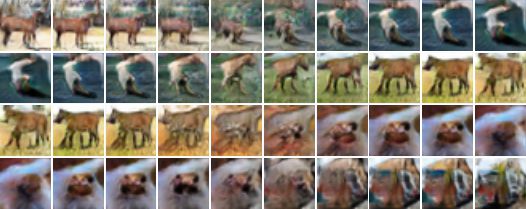}
        \caption{Random model sample interpolations.}
    \end{subfigure}%
    \caption{Qualitative results of training BiGAN on Cifar10. Interpolations are obtain by embedding two real/fake images $x_1$ and $x_2$ into the latent space with $E$, then decoding images uniformly along a line between these two embeddings $G\left(\alpha E(x_1) + (1-\alpha) E(x_2)\right)$. Left- and right-most images in each interpolation row are originals. \newline \newline
Statistics: $FID(P_X, P_G)=29.45$, \ $FID(P_X, P_{G(E(X))})=25.96$, \ Inception $L_2(P_X, P_{G(E(X))}) = 16.56$. \newline
Hyper-parameters: Adam learning rate 0.0003, $\beta_1=0.5$, $\beta_2=0.999$, 2 discriminator update(s) per generator update, WGAN-GP penalty weight $10.0$.
} 
    \label{fig:qualitative_gan_x_adv_cifar10}
\end{figure}

\begin{figure}
    \centering
    \textbf{Model: GAN + X adversarial loss. \  Dataset: CelebA.}
    
    \vspace{2em}
    
    \begin{subfigure}{0.5\columnwidth}
        \centering
        \includegraphics[width=\columnwidth]{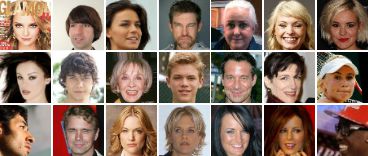}
        \caption{Real images.}
    \end{subfigure}%
    ~
    \begin{subfigure}{0.5\columnwidth}
        \centering
        \includegraphics[width=\columnwidth]{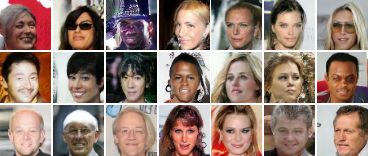}
        \caption{Random model samples.}
    \end{subfigure}%
    
    \vspace{2em}
    
    \begin{subfigure}{\columnwidth}
        \centering
        \includegraphics[width=\columnwidth]{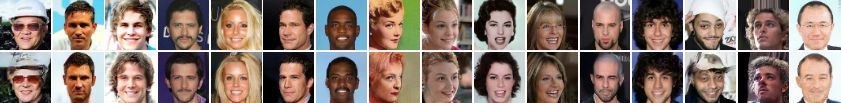}
        \caption{Real images (top row) and reconstructions (bottom row).}
    \end{subfigure}%
    
    \vspace{2em}
    
    \begin{subfigure}{\columnwidth}
        \centering
        \includegraphics[width=\columnwidth]{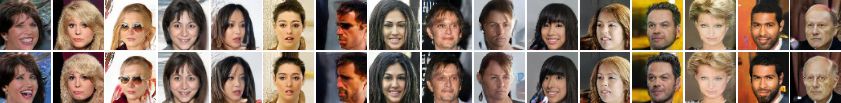}
        \caption{Random model samples (top row) and reconstructions (bottom row).}
    \end{subfigure}%
    
    \vspace{2em}
    
    \begin{subfigure}{0.5\columnwidth}
        \centering
        \includegraphics[width=\columnwidth]{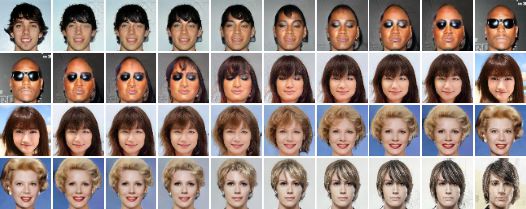}%
        \caption{Real sample interpolations.}
    \end{subfigure}%
    ~
    \begin{subfigure}{0.5\columnwidth}
        \centering
        \includegraphics[width=\columnwidth]{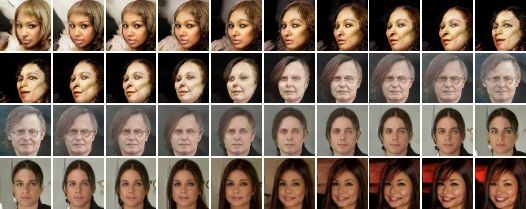}
        \caption{Random model sample interpolations.}
    \end{subfigure}%
    \caption{Qualitative results of training BiGAN on CelebA. Interpolations are obtain by embedding two real/fake images $x_1$ and $x_2$ into the latent space with $E$, then decoding images uniformly along a line between these two embeddings $G\left(\alpha E(x_1) + (1-\alpha) E(x_2)\right)$. Left- and right-most images in each interpolation row are originals. \newline \newline
Statistics: $FID(P_X, P_G)=7.83$, \ $FID(P_X, P_{G(E(X))})=8.82$, \ Inception $L_2(P_X, P_{G(E(X))}) = 10.46$. \newline
Hyper-parameters: Adam learning rate 0.0003, $\beta_1=0.5$, $\beta_2=0.999$, 2 discriminator update(s) per generator update, WGAN-GP penalty weight $1.0$.
} 
    \label{fig:qualitative_gan_x_adv_celeba}
\end{figure}

\begin{figure}
    \centering
    \textbf{Model: GAN + X adversarial loss. \  Dataset: Flintstones.}
    
    \vspace{2em}
    
    \begin{subfigure}{0.5\columnwidth}
        \centering
        \includegraphics[width=\columnwidth]{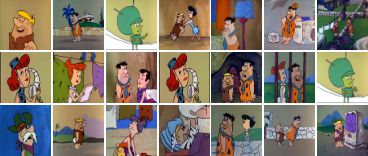}
        \caption{Real images.}
    \end{subfigure}%
    ~
    \begin{subfigure}{0.5\columnwidth}
        \centering
        \includegraphics[width=\columnwidth]{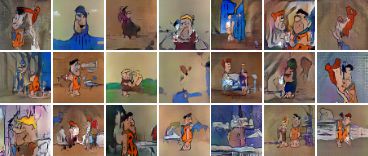}
        \caption{Random model samples.}
    \end{subfigure}%
    
    \vspace{2em}
    
    \begin{subfigure}{\columnwidth}
        \centering
        \includegraphics[width=\columnwidth]{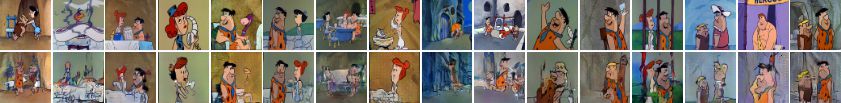}
        \caption{Real images (top row) and reconstructions (bottom row).}
    \end{subfigure}%
    
    \vspace{2em}
    
    \begin{subfigure}{\columnwidth}
        \centering
        \includegraphics[width=\columnwidth]{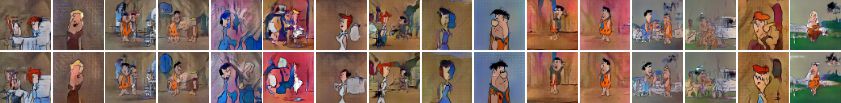}
        \caption{Random model samples (top row) and reconstructions (bottom row).}
    \end{subfigure}%
    
    \vspace{2em}
    
    \begin{subfigure}{0.5\columnwidth}
        \centering
        \includegraphics[width=\columnwidth]{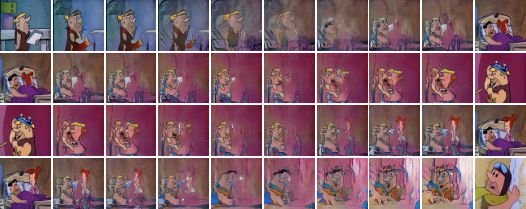}%
        \caption{Real sample interpolations.}
    \end{subfigure}%
    ~
    \begin{subfigure}{0.5\columnwidth}
        \centering
        \includegraphics[width=\columnwidth]{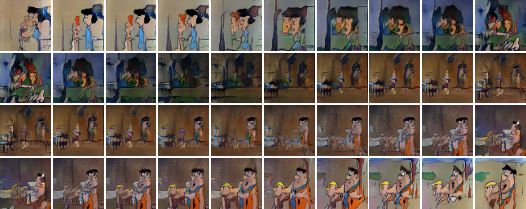}
        \caption{Random model sample interpolations.}
    \end{subfigure}%
    \caption{Qualitative results of training BiGAN on Flintstones. Interpolations are obtain by embedding two real/fake images $x_1$ and $x_2$ into the latent space with $E$, then decoding images uniformly along a line between these two embeddings $G\left(\alpha E(x_1) + (1-\alpha) E(x_2)\right)$. Left- and right-most images in each interpolation row are originals.\newline \newline
Statistics: $FID(P_X, P_G)=83.91$, \ $FID(P_X, P_{G(E(X))})=89.85$, \ Inception $L_2(P_X, P_{G(E(X))}) = 15.48$. \newline
Hyper-parameters: Adam learning rate 0.0003, $\beta_1=0.5$, $\beta_2=0.999$, 2 discriminator update(s) per generator update, WGAN-GP penalty weight $10.0$.
} 
    \label{fig:qualitative_gan_x_adv_flintstones}
\end{figure}

\begin{figure}
    \centering
    \textbf{Model: GAN + X adversarial loss. \  Dataset: ImageNet.}
    
    \vspace{2em}
    
    \begin{subfigure}{0.5\columnwidth}
        \centering
        \includegraphics[width=\columnwidth]{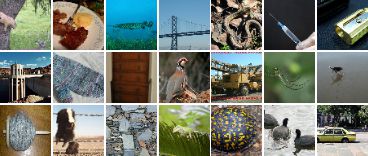}
        \caption{Real images.}
    \end{subfigure}%
    ~
    \begin{subfigure}{0.5\columnwidth}
        \centering
        \includegraphics[width=\columnwidth]{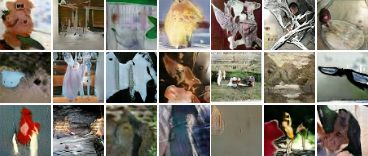}
        \caption{Random model samples.}
    \end{subfigure}%
    
    \vspace{2em}
    
    \begin{subfigure}{\columnwidth}
        \centering
        \includegraphics[width=\columnwidth]{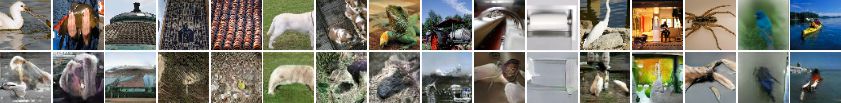}
        \caption{Real images (top row) and reconstructions (bottom row).}
    \end{subfigure}%
    
    \vspace{2em}
    
    \begin{subfigure}{\columnwidth}
        \centering
        \includegraphics[width=\columnwidth]{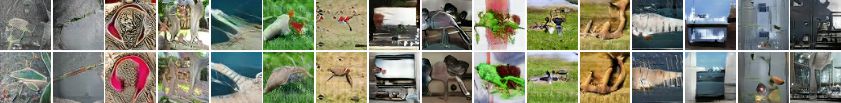}
        \caption{Random model samples (top row) and reconstructions (bottom row).}
    \end{subfigure}%
    
    \vspace{2em}
    
    \begin{subfigure}{0.5\columnwidth}
        \centering
        \includegraphics[width=\columnwidth]{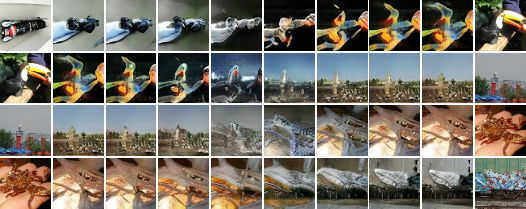}%
        \caption{Real sample interpolations.}
    \end{subfigure}%
    ~
    \begin{subfigure}{0.5\columnwidth}
        \centering
        \includegraphics[width=\columnwidth]{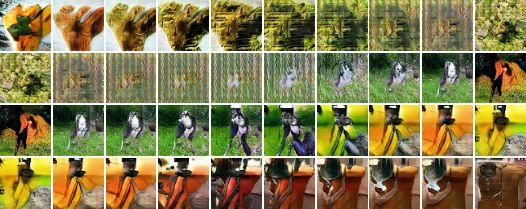}
        \caption{Random model sample interpolations.}
    \end{subfigure}%
    \caption{Qualitative results of training BiGAN on ImageNet. Interpolations are obtain by embedding two real/fake images $x_1$ and $x_2$ into the latent space with $E$, then decoding images uniformly along a line between these two embeddings $G\left(\alpha E(x_1) + (1-\alpha) E(x_2)\right)$. Left- and right-most images in each interpolation row are originals. \newline \newline
Statistics: $FID(P_X, P_G)=58.66$, \ $FID(P_X, P_{G(E(X))})=58.15$, \ Inception $L_2(P_X, P_{G(E(X))}) = 19.68$. \newline
Hyper-parameters: Adam learning rate 0.001, $\beta_1=0.5$, $\beta_2=0.999$, 2 discriminator update(s) per generator update, WGAN-GP penalty weight $10.0$.} 
    \label{fig:qualitative_gan_x_adv_imagenet}
\end{figure}


\begin{figure}
    \centering
    \textbf{Model: GAN + Z adversarial loss. \  Dataset: Cifar10.}
    
    \vspace{2em}
    
    \begin{subfigure}{0.5\columnwidth}
        \centering
        \includegraphics[width=\columnwidth]{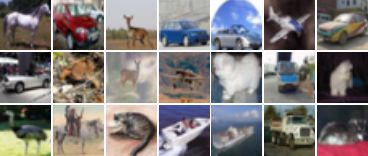}
        \caption{Real images.}
    \end{subfigure}%
    ~
    \begin{subfigure}{0.5\columnwidth}
        \centering
        \includegraphics[width=\columnwidth]{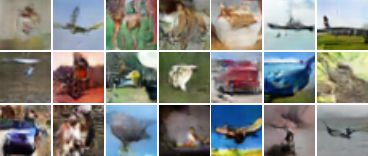}
        \caption{Random model samples.}
    \end{subfigure}%
    
    \vspace{2em}
    
    \begin{subfigure}{\columnwidth}
        \centering
        \includegraphics[width=\columnwidth]{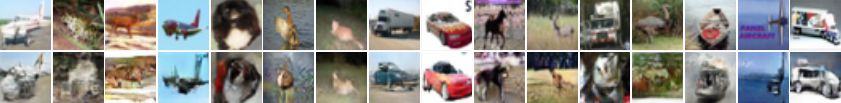}
        \caption{Real images (top row) and reconstructions (bottom row).}
    \end{subfigure}%
    
    \vspace{2em}
    
    \begin{subfigure}{\columnwidth}
        \centering
        \includegraphics[width=\columnwidth]{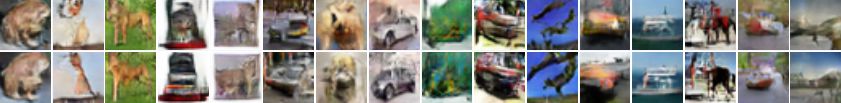}
        \caption{Random model samples (top row) and reconstructions (bottom row).}
    \end{subfigure}%
    
    \vspace{2em}
    
    \begin{subfigure}{0.5\columnwidth}
        \centering
        \includegraphics[width=\columnwidth]{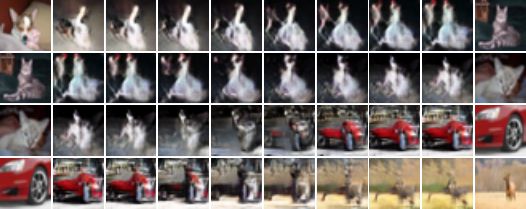}%
        \caption{Real sample interpolations.}
    \end{subfigure}%
    ~
    \begin{subfigure}{0.5\columnwidth}
        \centering
        \includegraphics[width=\columnwidth]{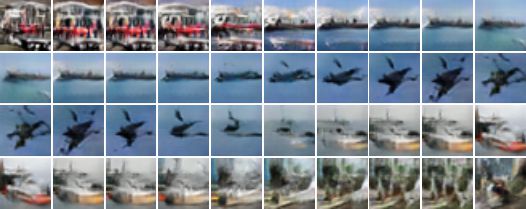}
        \caption{Random model sample interpolations.}
    \end{subfigure}%
    \caption{Qualitative results of training BiGAN on Cifar10. Interpolations are obtain by embedding two real/fake images $x_1$ and $x_2$ into the latent space with $E$, then decoding images uniformly along a line between these two embeddings $G\left(\alpha E(x_1) + (1-\alpha) E(x_2)\right)$. Left- and right-most images in each interpolation row are originals. \newline \newline
Statistics: $FID(P_X, P_G)=31.30$, \ $FID(P_X, P_{G(E(X))})=25.71$, \ Inception $L_2(P_X, P_{G(E(X))}) = 17.01$. \newline
Hyper-parameters: Adam learning rate 0.0001, $\beta_1=0.5$, $\beta_2=0.999$, 2 discriminator update(s) per generator update, WGAN-GP penalty weight $10.0$.
} 
    \label{fig:qualitative_gan_z_adv_cifar10}
\end{figure}

\begin{figure}
    \centering
    \textbf{Model: GAN + Z adversarial loss. \  Dataset: CelebA.}
    
    \vspace{2em}
    
    \begin{subfigure}{0.5\columnwidth}
        \centering
        \includegraphics[width=\columnwidth]{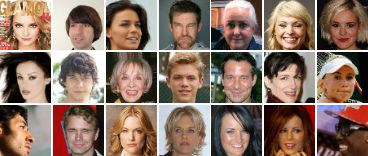}
        \caption{Real images.}
    \end{subfigure}%
    ~
    \begin{subfigure}{0.5\columnwidth}
        \centering
        \includegraphics[width=\columnwidth]{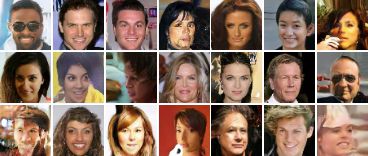}
        \caption{Random model samples.}
    \end{subfigure}%
    
    \vspace{2em}
    
    \begin{subfigure}{\columnwidth}
        \centering
        \includegraphics[width=\columnwidth]{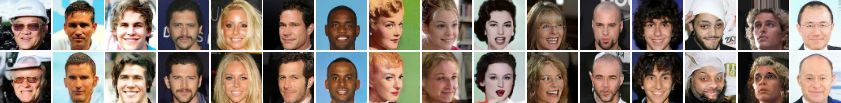}
        \caption{Real images (top row) and reconstructions (bottom row).}
    \end{subfigure}%
    
    \vspace{2em}
    
    \begin{subfigure}{\columnwidth}
        \centering
        \includegraphics[width=\columnwidth]{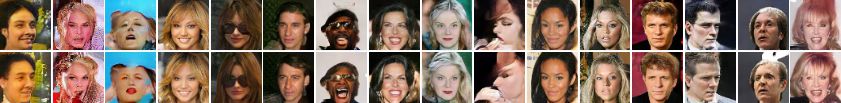}
        \caption{Random model samples (top row) and reconstructions (bottom row).}
    \end{subfigure}%
    
    \vspace{2em}
    
    \begin{subfigure}{0.5\columnwidth}
        \centering
        \includegraphics[width=\columnwidth]{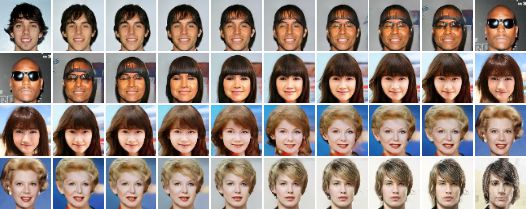}%
        \caption{Real sample interpolations.}
    \end{subfigure}%
    ~
    \begin{subfigure}{0.5\columnwidth}
        \centering
        \includegraphics[width=\columnwidth]{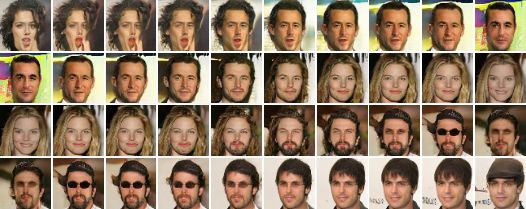}
        \caption{Random model sample interpolations.}
    \end{subfigure}%
    \caption{Qualitative results of training BiGAN on CelebA. Interpolations are obtain by embedding two real/fake images $x_1$ and $x_2$ into the latent space with $E$, then decoding images uniformly along a line between these two embeddings $G\left(\alpha E(x_1) + (1-\alpha) E(x_2)\right)$. Left- and right-most images in each interpolation row are originals. \newline \newline
Statistics: $FID(P_X, P_G)=7.10$, \ $FID(P_X, P_{G(E(X))})=7.76$, \ Inception $L_2(P_X, P_{G(E(X))}) = 10.81$. \newline
Hyper-parameters: Adam learning rate 0.0001, $\beta_1=0.5$, $\beta_2=0.999$, 1 discriminator update(s) per generator update, WGAN-GP penalty weight $1.0$.
} 
    \label{fig:qualitative_gan_z_adv_celeba}
\end{figure}

\begin{figure}
    \centering
    \textbf{Model: GAN + Z adversarial loss. \  Dataset: Flintstones.}
    
    \vspace{2em}
    
    \begin{subfigure}{0.5\columnwidth}
        \centering
        \includegraphics[width=\columnwidth]{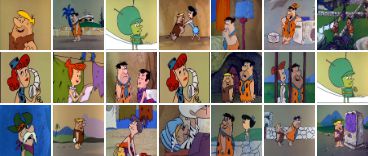}
        \caption{Real images.}
    \end{subfigure}%
    ~
    \begin{subfigure}{0.5\columnwidth}
        \centering
        \includegraphics[width=\columnwidth]{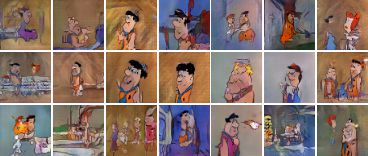}
        \caption{Random model samples.}
    \end{subfigure}%
    
    \vspace{2em}
    
    \begin{subfigure}{\columnwidth}
        \centering
        \includegraphics[width=\columnwidth]{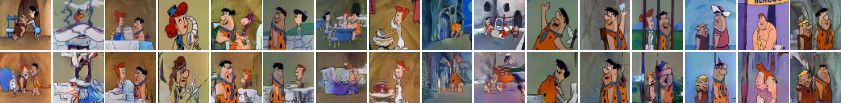}
        \caption{Real images (top row) and reconstructions (bottom row).}
    \end{subfigure}%
    
    \vspace{2em}
    
    \begin{subfigure}{\columnwidth}
        \centering
        \includegraphics[width=\columnwidth]{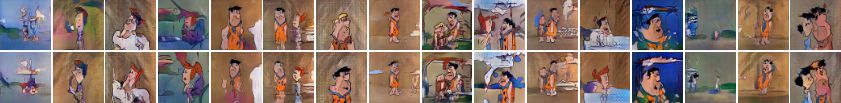}
        \caption{Random model samples (top row) and reconstructions (bottom row).}
    \end{subfigure}%
    
    \vspace{2em}
    
    \begin{subfigure}{0.5\columnwidth}
        \centering
        \includegraphics[width=\columnwidth]{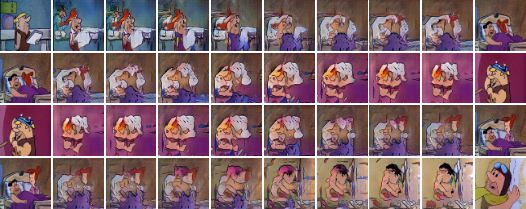}%
        \caption{Real sample interpolations.}
    \end{subfigure}%
    ~
    \begin{subfigure}{0.5\columnwidth}
        \centering
        \includegraphics[width=\columnwidth]{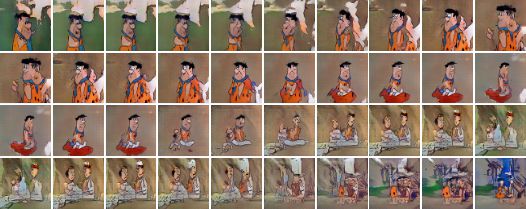}
        \caption{Random model sample interpolations.}
    \end{subfigure}%
    \caption{Qualitative results of training BiGAN on Flintstones. Interpolations are obtain by embedding two real/fake images $x_1$ and $x_2$ into the latent space with $E$, then decoding images uniformly along a line between these two embeddings $G\left(\alpha E(x_1) + (1-\alpha) E(x_2)\right)$. Left- and right-most images in each interpolation row are originals.\newline \newline
Statistics: $FID(P_X, P_G)=85.67$, \ $FID(P_X, P_{G(E(X))})=88.68$, \ Inception $L_2(P_X, P_{G(E(X))}) = 15.51$. \newline
Hyper-parameters: Adam learning rate 0.0001, $\beta_1=0.5$, $\beta_2=0.999$, 2 discriminator update(s) per generator update, WGAN-GP penalty weight $1.0$.
} 
    \label{fig:qualitative_gan_z_adv_flintstones}
\end{figure}

\begin{figure}
    \centering
    \textbf{Model: GAN + Z adversarial loss. \  Dataset: ImageNet.}
    
    \vspace{2em}
    
    \begin{subfigure}{0.5\columnwidth}
        \centering
        \includegraphics[width=\columnwidth]{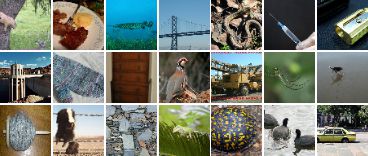}
        \caption{Real images.}
    \end{subfigure}%
    ~
    \begin{subfigure}{0.5\columnwidth}
        \centering
        \includegraphics[width=\columnwidth]{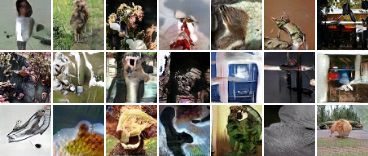}
        \caption{Random model samples.}
    \end{subfigure}%
    
    \vspace{2em}
    
    \begin{subfigure}{\columnwidth}
        \centering
        \includegraphics[width=\columnwidth]{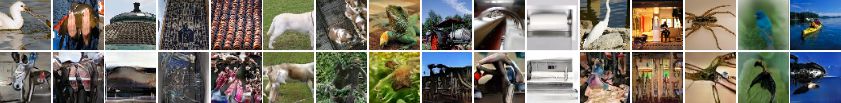}
        \caption{Real images (top row) and reconstructions (bottom row).}
    \end{subfigure}%
    
    \vspace{2em}
    
    \begin{subfigure}{\columnwidth}
        \centering
        \includegraphics[width=\columnwidth]{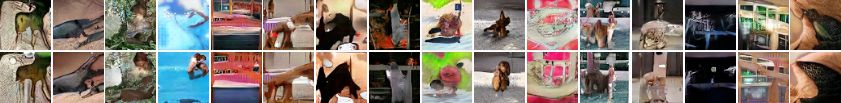}
        \caption{Random model samples (top row) and reconstructions (bottom row).}
    \end{subfigure}%
    
    \vspace{2em}
    
    \begin{subfigure}{0.5\columnwidth}
        \centering
        \includegraphics[width=\columnwidth]{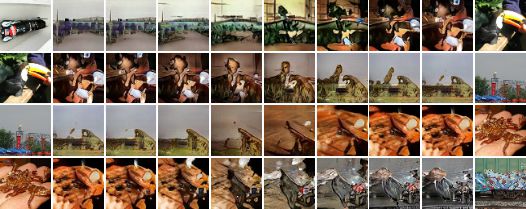}%
        \caption{Real sample interpolations.}
    \end{subfigure}%
    ~
    \begin{subfigure}{0.5\columnwidth}
        \centering
        \includegraphics[width=\columnwidth]{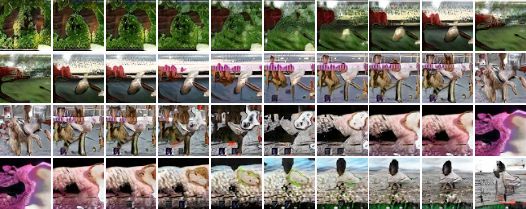}
        \caption{Random model sample interpolations.}
    \end{subfigure}%
    \caption{Qualitative results of training BiGAN on ImageNet. Interpolations are obtain by embedding two real/fake images $x_1$ and $x_2$ into the latent space with $E$, then decoding images uniformly along a line between these two embeddings $G\left(\alpha E(x_1) + (1-\alpha) E(x_2)\right)$. Left- and right-most images in each interpolation row are originals. \newline \newline
Statistics: $FID(P_X, P_G)=59.06$, \ $FID(P_X, P_{G(E(X))})=60.84$, \ Inception $L_2(P_X, P_{G(E(X))}) = 19.63$. \newline
Hyper-parameters: Adam learning rate 0.0003, $\beta_1=0.5$, $\beta_2=0.999$, 2 discriminator update(s) per generator update, WGAN-GP penalty weight $1.0$.
} 
    \label{fig:qualitative_gan_z_adv_imagenet}
\end{figure}


\begin{figure}
    \centering
    \textbf{Model: BiGAN + X Auto-encoder. \  Dataset: Cifar10.}
    
    \vspace{2em}
    
    \begin{subfigure}{0.5\columnwidth}
        \centering
        \includegraphics[width=\columnwidth]{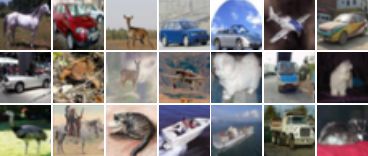}
        \caption{Real images.}
    \end{subfigure}%
    ~
    \begin{subfigure}{0.5\columnwidth}
        \centering
        \includegraphics[width=\columnwidth]{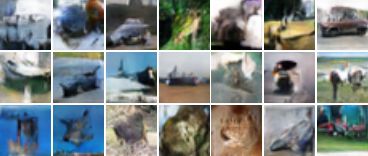}
        \caption{Random model samples.}
    \end{subfigure}%
    
    \vspace{2em}
    
    \begin{subfigure}{\columnwidth}
        \centering
        \includegraphics[width=\columnwidth]{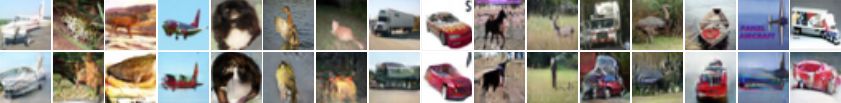}
        \caption{Real images (top row) and reconstructions (bottom row).}
    \end{subfigure}%
    
    \vspace{2em}
    
    \begin{subfigure}{\columnwidth}
        \centering
        \includegraphics[width=\columnwidth]{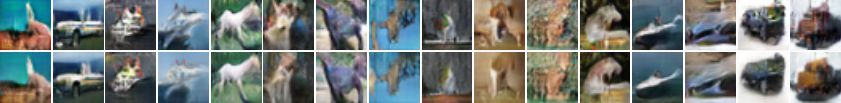}
        \caption{Random model samples (top row) and reconstructions (bottom row).}
    \end{subfigure}%
    
    \vspace{2em}
    
    \begin{subfigure}{0.5\columnwidth}
        \centering
        \includegraphics[width=\columnwidth]{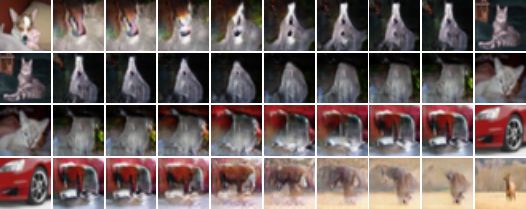}%
        \caption{Real sample interpolations.}
    \end{subfigure}%
    ~
    \begin{subfigure}{0.5\columnwidth}
        \centering
        \includegraphics[width=\columnwidth]{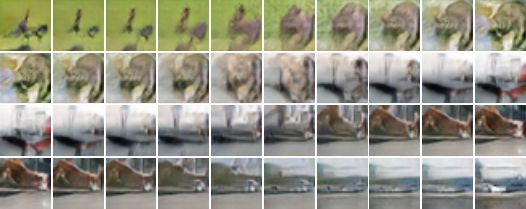}
        \caption{Random model sample interpolations.}
    \end{subfigure}%
    \caption{Qualitative results of training BiGAN on Cifar10. Interpolations are obtain by embedding two real/fake images $x_1$ and $x_2$ into the latent space with $E$, then decoding images uniformly along a line between these two embeddings $G\left(\alpha E(x_1) + (1-\alpha) E(x_2)\right)$. Left- and right-most images in each interpolation row are originals. \newline \newline
Statistics: $FID(P_X, P_G)=31.20$, \ $FID(P_X, P_{G(E(X))})=23.92$, \ Inception $L_2(P_X, P_{G(E(X))}) = 16.77$. \newline
Hyper-parameters: Adam learning rate 0.0001, $\beta_1=0.5$, $\beta_2=0.999$, 2 discriminator update(s) per generator update, WGAN-GP penalty weight $1.0$.
\newline Weighting of additional loss $\lambda=1$.
} 
    \label{fig:qualitative_bigan_x_ae_cifar10}
\end{figure}

\begin{figure}
    \centering
    \textbf{Model: BiGAN + X Auto-encoder. \  Dataset: CelebA.}
    
    \vspace{2em}
    
    \begin{subfigure}{0.5\columnwidth}
        \centering
        \includegraphics[width=\columnwidth]{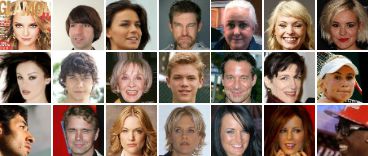}
        \caption{Real images.}
    \end{subfigure}%
    ~
    \begin{subfigure}{0.5\columnwidth}
        \centering
        \includegraphics[width=\columnwidth]{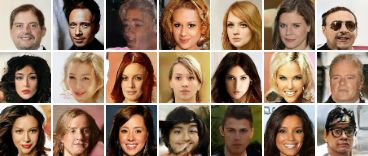}
        \caption{Random model samples.}
    \end{subfigure}%
    
    \vspace{2em}
    
    \begin{subfigure}{\columnwidth}
        \centering
        \includegraphics[width=\columnwidth]{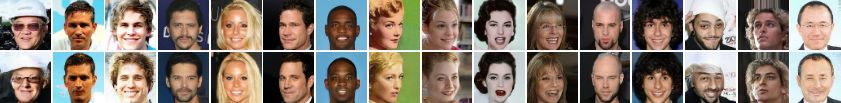}
        \caption{Real images (top row) and reconstructions (bottom row).}
    \end{subfigure}%
    
    \vspace{2em}
    
    \begin{subfigure}{\columnwidth}
        \centering
        \includegraphics[width=\columnwidth]{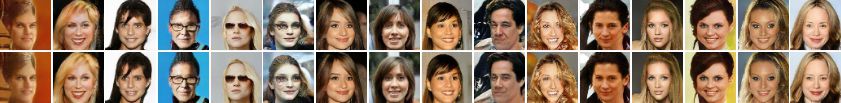}
        \caption{Random model samples (top row) and reconstructions (bottom row).}
    \end{subfigure}%
    
    \vspace{2em}
    
    \begin{subfigure}{0.5\columnwidth}
        \centering
        \includegraphics[width=\columnwidth]{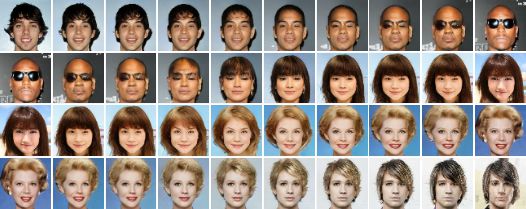}%
        \caption{Real sample interpolations.}
    \end{subfigure}%
    ~
    \begin{subfigure}{0.5\columnwidth}
        \centering
        \includegraphics[width=\columnwidth]{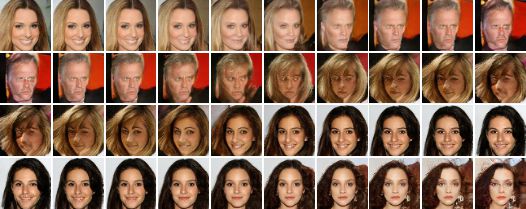}
        \caption{Random model sample interpolations.}
    \end{subfigure}%
    \caption{Qualitative results of training BiGAN on CelebA. Interpolations are obtain by embedding two real/fake images $x_1$ and $x_2$ into the latent space with $E$, then decoding images uniformly along a line between these two embeddings $G\left(\alpha E(x_1) + (1-\alpha) E(x_2)\right)$. Left- and right-most images in each interpolation row are originals. \newline \newline
Statistics: $FID(P_X, P_G)=8.41$, \ $FID(P_X, P_{G(E(X))})=9.06$, \ Inception $L_2(P_X, P_{G(E(X))}) = 9.87$. \newline
Hyper-parameters: Adam learning rate 0.0001, $\beta_1=0.5$, $\beta_2=0.999$, 1 discriminator update(s) per generator update, WGAN-GP penalty weight $1.0$.
\newline Weighting of additional loss $\lambda=0.300000$.
} 
    \label{fig:qualitative_bigan_x_ae_celeba}
\end{figure}

\begin{figure}
    \centering
    \textbf{Model: BiGAN + X Auto-encoder. \  Dataset: Flintstones.}
    
    \vspace{2em}
    
    \begin{subfigure}{0.5\columnwidth}
        \centering
        \includegraphics[width=\columnwidth]{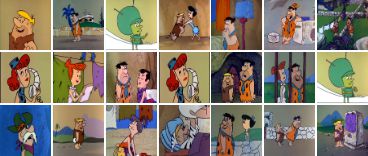}
        \caption{Real images.}
    \end{subfigure}%
    ~
    \begin{subfigure}{0.5\columnwidth}
        \centering
        \includegraphics[width=\columnwidth]{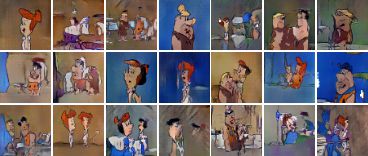}
        \caption{Random model samples.}
    \end{subfigure}%
    
    \vspace{2em}
    
    \begin{subfigure}{\columnwidth}
        \centering
        \includegraphics[width=\columnwidth]{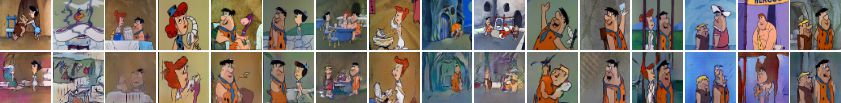}
        \caption{Real images (top row) and reconstructions (bottom row).}
    \end{subfigure}%
    
    \vspace{2em}
    
    \begin{subfigure}{\columnwidth}
        \centering
        \includegraphics[width=\columnwidth]{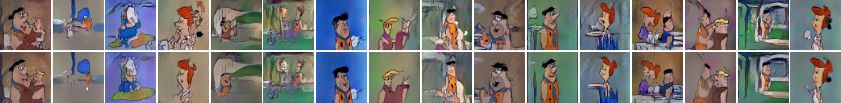}
        \caption{Random model samples (top row) and reconstructions (bottom row).}
    \end{subfigure}%
    
    \vspace{2em}
    
    \begin{subfigure}{0.5\columnwidth}
        \centering
        \includegraphics[width=\columnwidth]{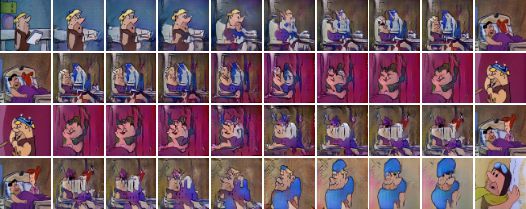}%
        \caption{Real sample interpolations.}
    \end{subfigure}%
    ~
    \begin{subfigure}{0.5\columnwidth}
        \centering
        \includegraphics[width=\columnwidth]{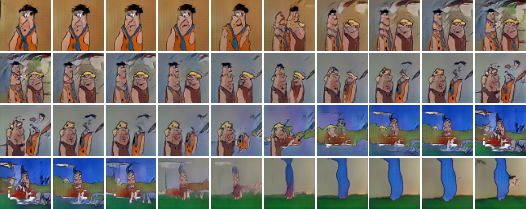}
        \caption{Random model sample interpolations.}
    \end{subfigure}%
    \caption{Qualitative results of training BiGAN on Flintstones. Interpolations are obtain by embedding two real/fake images $x_1$ and $x_2$ into the latent space with $E$, then decoding images uniformly along a line between these two embeddings $G\left(\alpha E(x_1) + (1-\alpha) E(x_2)\right)$. Left- and right-most images in each interpolation row are originals.\newline \newline
Statistics: $FID(P_X, P_G)=82.71$, \ $FID(P_X, P_{G(E(X))})=82.21$, \ Inception $L_2(P_X, P_{G(E(X))}) = 15.85$. \newline
Hyper-parameters: Adam learning rate 0.001, $\beta_1=0.5$, $\beta_2=0.999$, 1 discriminator update(s) per generator update, WGAN-GP penalty weight $10.0$.
\newline Weighting of additional loss $\lambda=10$.
} 
    \label{fig:qualitative_bigan_x_ae_flintstones}
\end{figure}

\begin{figure}
    \centering
    \textbf{Model: BiGAN + X Auto-encoder. \  Dataset: ImageNet.}
    
    \vspace{2em}
    
    \begin{subfigure}{0.5\columnwidth}
        \centering
        \includegraphics[width=\columnwidth]{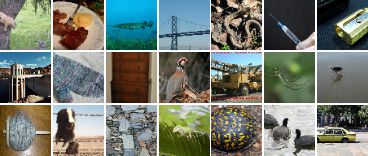}
        \caption{Real images.}
    \end{subfigure}%
    ~
    \begin{subfigure}{0.5\columnwidth}
        \centering
        \includegraphics[width=\columnwidth]{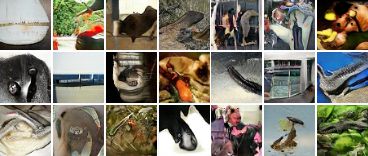}
        \caption{Random model samples.}
    \end{subfigure}%
    
    \vspace{2em}
    
    \begin{subfigure}{\columnwidth}
        \centering
        \includegraphics[width=\columnwidth]{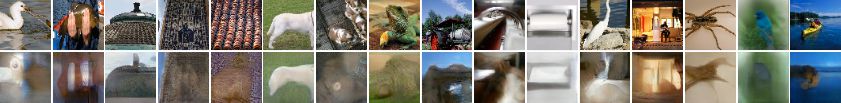}
        \caption{Real images (top row) and reconstructions (bottom row).}
    \end{subfigure}%
    
    \vspace{2em}
    
    \begin{subfigure}{\columnwidth}
        \centering
        \includegraphics[width=\columnwidth]{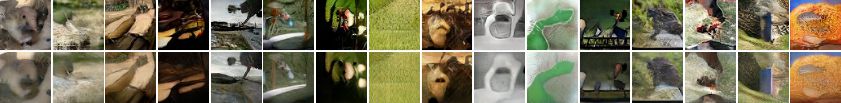}
        \caption{Random model samples (top row) and reconstructions (bottom row).}
    \end{subfigure}%
    
    \vspace{2em}
    
    \begin{subfigure}{0.5\columnwidth}
        \centering
        \includegraphics[width=\columnwidth]{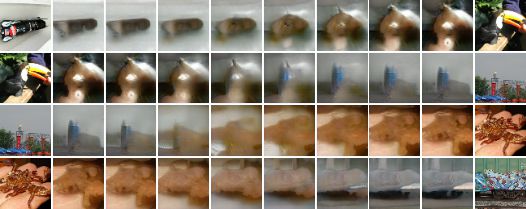}%
        \caption{Real sample interpolations.}
    \end{subfigure}%
    ~
    \begin{subfigure}{0.5\columnwidth}
        \centering
        \includegraphics[width=\columnwidth]{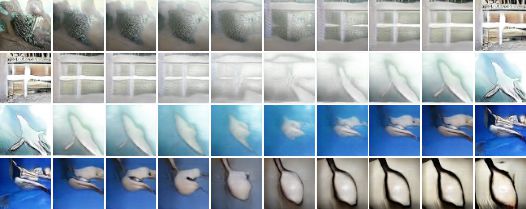}
        \caption{Random model sample interpolations.}
    \end{subfigure}%
    \caption{Qualitative results of training BiGAN on ImageNet. Interpolations are obtain by embedding two real/fake images $x_1$ and $x_2$ into the latent space with $E$, then decoding images uniformly along a line between these two embeddings $G\left(\alpha E(x_1) + (1-\alpha) E(x_2)\right)$. Left- and right-most images in each interpolation row are originals. \newline \newline
Statistics: $FID(P_X, P_G)=58.60$, \ $FID(P_X, P_{G(E(X))})=47.26$, \ Inception $L_2(P_X, P_{G(E(X))}) = 19.13$. \newline
Hyper-parameters: Adam learning rate 0.0001, $\beta_1=0.5$, $\beta_2=0.999$, 1 discriminator update(s) per generator update, WGAN-GP penalty weight $1.0$.
\newline Weighting of additional loss $\lambda=0.100000$.
} 
    \label{fig:qualitative_bigan_x_ae_imagenet}
\end{figure}


\begin{figure}
    \centering
    \textbf{Model: BiGAN + Z Auto-encoder. \  Dataset: Cifar10.}
    
    \vspace{2em}
    
    \begin{subfigure}{0.5\columnwidth}
        \centering
        \includegraphics[width=\columnwidth]{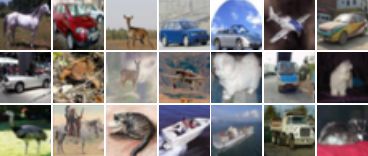}
        \caption{Real images.}
    \end{subfigure}%
    ~
    \begin{subfigure}{0.5\columnwidth}
        \centering
        \includegraphics[width=\columnwidth]{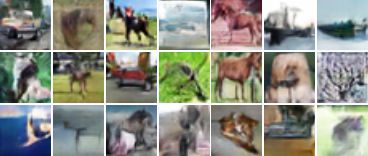}
        \caption{Random model samples.}
    \end{subfigure}%
    
    \vspace{2em}
    
    \begin{subfigure}{\columnwidth}
        \centering
        \includegraphics[width=\columnwidth]{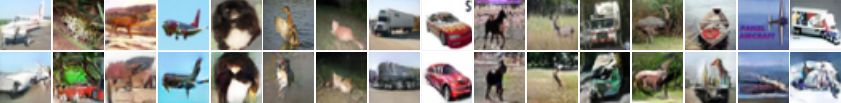}
        \caption{Real images (top row) and reconstructions (bottom row).}
    \end{subfigure}%
    
    \vspace{2em}
    
    \begin{subfigure}{\columnwidth}
        \centering
        \includegraphics[width=\columnwidth]{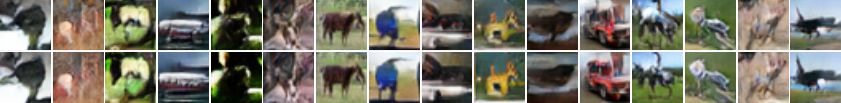}
        \caption{Random model samples (top row) and reconstructions (bottom row).}
    \end{subfigure}%
    
    \vspace{2em}
    
    \begin{subfigure}{0.5\columnwidth}
        \centering
        \includegraphics[width=\columnwidth]{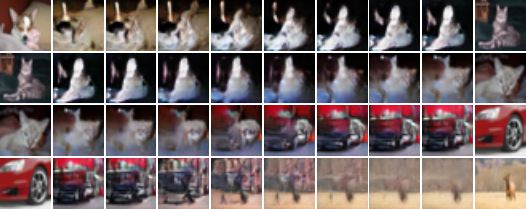}%
        \caption{Real sample interpolations.}
    \end{subfigure}%
    ~
    \begin{subfigure}{0.5\columnwidth}
        \centering
        \includegraphics[width=\columnwidth]{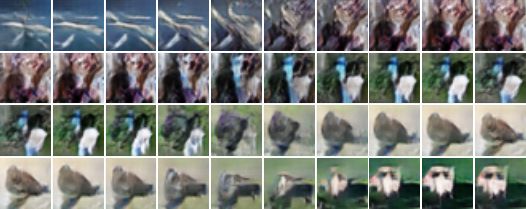}
        \caption{Random model sample interpolations.}
    \end{subfigure}%
    \caption{Qualitative results of training BiGAN on Cifar10. Interpolations are obtain by embedding two real/fake images $x_1$ and $x_2$ into the latent space with $E$, then decoding images uniformly along a line between these two embeddings $G\left(\alpha E(x_1) + (1-\alpha) E(x_2)\right)$. Left- and right-most images in each interpolation row are originals. \newline \newline
Statistics: $FID(P_X, P_G)=33.40$, \ $FID(P_X, P_{G(E(X))})=26.23$, \ Inception $L_2(P_X, P_{G(E(X))}) = 16.96$. \newline
Hyper-parameters: Adam learning rate 0.0003, $\beta_1=0.5$, $\beta_2=0.999$, 2 discriminator update(s) per generator update, WGAN-GP penalty weight $10.0$.
\newline Weighting of additional loss $\lambda=10$.
} 
    \label{fig:qualitative_bigan_z_ae_cifar10}
\end{figure}

\begin{figure}
    \centering
    \textbf{Model: BiGAN + Z Auto-encoder. \  Dataset: CelebA.}
    
    \vspace{2em}
    
    \begin{subfigure}{0.5\columnwidth}
        \centering
        \includegraphics[width=\columnwidth]{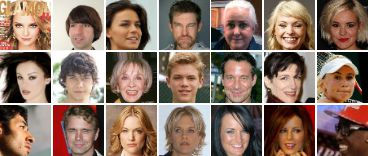}
        \caption{Real images.}
    \end{subfigure}%
    ~
    \begin{subfigure}{0.5\columnwidth}
        \centering
        \includegraphics[width=\columnwidth]{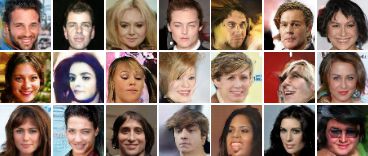}
        \caption{Random model samples.}
    \end{subfigure}%
    
    \vspace{2em}
    
    \begin{subfigure}{\columnwidth}
        \centering
        \includegraphics[width=\columnwidth]{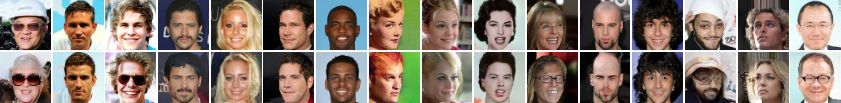}
        \caption{Real images (top row) and reconstructions (bottom row).}
    \end{subfigure}%
    
    \vspace{2em}
    
    \begin{subfigure}{\columnwidth}
        \centering
        \includegraphics[width=\columnwidth]{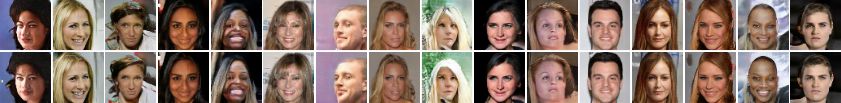}
        \caption{Random model samples (top row) and reconstructions (bottom row).}
    \end{subfigure}%
    
    \vspace{2em}
    
    \begin{subfigure}{0.5\columnwidth}
        \centering
        \includegraphics[width=\columnwidth]{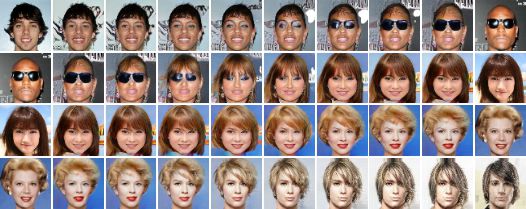}%
        \caption{Real sample interpolations.}
    \end{subfigure}%
    ~
    \begin{subfigure}{0.5\columnwidth}
        \centering
        \includegraphics[width=\columnwidth]{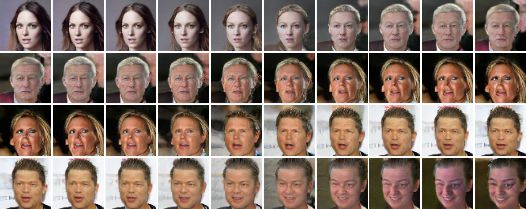}
        \caption{Random model sample interpolations.}
    \end{subfigure}%
    \caption{Qualitative results of training BiGAN on CelebA. Interpolations are obtain by embedding two real/fake images $x_1$ and $x_2$ into the latent space with $E$, then decoding images uniformly along a line between these two embeddings $G\left(\alpha E(x_1) + (1-\alpha) E(x_2)\right)$. Left- and right-most images in each interpolation row are originals. \newline \newline
Statistics: $FID(P_X, P_G)=7.56$, \ $FID(P_X, P_{G(E(X))})=5.91$, \ Inception $L_2(P_X, P_{G(E(X))}) = 10.96$. \newline
Hyper-parameters: Adam learning rate 0.0003, $\beta_1=0.5$, $\beta_2=0.999$, 2 discriminator update(s) per generator update, WGAN-GP penalty weight $10.0$.
\newline Weighting of additional loss $\lambda=0.300000$.
} 
    \label{fig:qualitative_bigan_z_ae_celeba}
\end{figure}

\begin{figure}
    \centering
    \textbf{Model: BiGAN + Z Auto-encoder. \  Dataset: Flintstones.}
    
    \vspace{2em}
    
    \begin{subfigure}{0.5\columnwidth}
        \centering
        \includegraphics[width=\columnwidth]{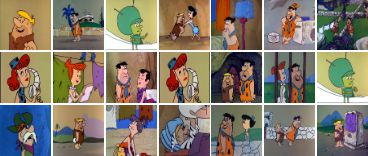}
        \caption{Real images.}
    \end{subfigure}%
    ~
    \begin{subfigure}{0.5\columnwidth}
        \centering
        \includegraphics[width=\columnwidth]{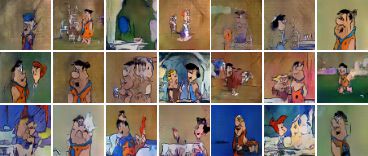}
        \caption{Random model samples.}
    \end{subfigure}%
    
    \vspace{2em}
    
    \begin{subfigure}{\columnwidth}
        \centering
        \includegraphics[width=\columnwidth]{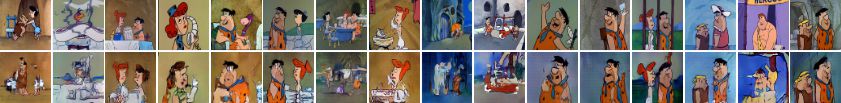}
        \caption{Real images (top row) and reconstructions (bottom row).}
    \end{subfigure}%
    
    \vspace{2em}
    
    \begin{subfigure}{\columnwidth}
        \centering
        \includegraphics[width=\columnwidth]{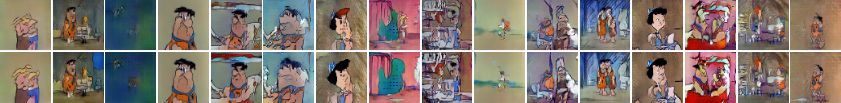}
        \caption{Random model samples (top row) and reconstructions (bottom row).}
    \end{subfigure}%
    
    \vspace{2em}
    
    \begin{subfigure}{0.5\columnwidth}
        \centering
        \includegraphics[width=\columnwidth]{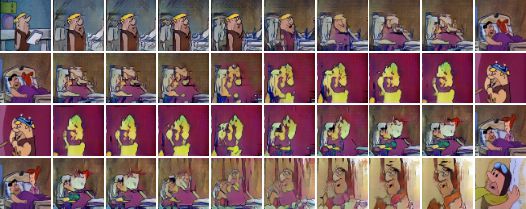}%
        \caption{Real sample interpolations.}
    \end{subfigure}%
    ~
    \begin{subfigure}{0.5\columnwidth}
        \centering
        \includegraphics[width=\columnwidth]{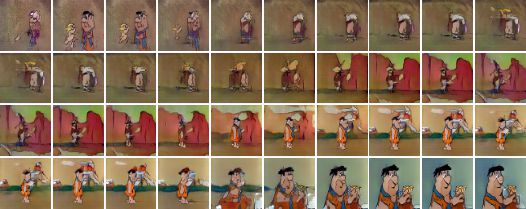}
        \caption{Random model sample interpolations.}
    \end{subfigure}%
    \caption{Qualitative results of training BiGAN on Flintstones. Interpolations are obtain by embedding two real/fake images $x_1$ and $x_2$ into the latent space with $E$, then decoding images uniformly along a line between these two embeddings $G\left(\alpha E(x_1) + (1-\alpha) E(x_2)\right)$. Left- and right-most images in each interpolation row are originals.\newline \newline
Statistics: $FID(P_X, P_G)=90.04$, \ $FID(P_X, P_{G(E(X))})=82.70$, \ Inception $L_2(P_X, P_{G(E(X))}) = 15.84$. \newline
Hyper-parameters: Adam learning rate 0.0001, $\beta_1=0.5$, $\beta_2=0.999$, 1 discriminator update(s) per generator update, WGAN-GP penalty weight $10.0$.
\newline Weighting of additional loss $\lambda=10$.
} 
    \label{fig:qualitative_bigan_z_ae_flintstones}
\end{figure}

\begin{figure}
    \centering
    \textbf{Model: BiGAN + Z Auto-encoder. \  Dataset: ImageNet.}
    
    \vspace{2em}
    
    \begin{subfigure}{0.5\columnwidth}
        \centering
        \includegraphics[width=\columnwidth]{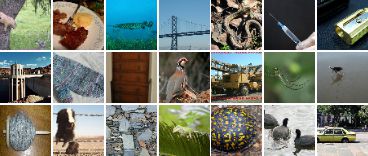}
        \caption{Real images.}
    \end{subfigure}%
    ~
    \begin{subfigure}{0.5\columnwidth}
        \centering
        \includegraphics[width=\columnwidth]{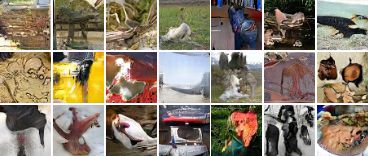}
        \caption{Random model samples.}
    \end{subfigure}%
    
    \vspace{2em}
    
    \begin{subfigure}{\columnwidth}
        \centering
        \includegraphics[width=\columnwidth]{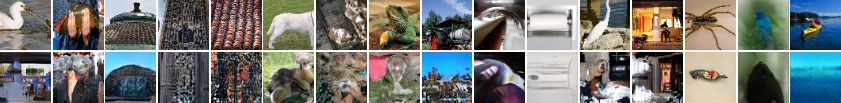}
        \caption{Real images (top row) and reconstructions (bottom row).}
    \end{subfigure}%
    
    \vspace{2em}
    
    \begin{subfigure}{\columnwidth}
        \centering
        \includegraphics[width=\columnwidth]{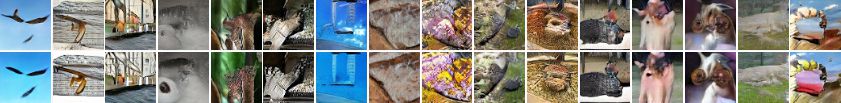}
        \caption{Random model samples (top row) and reconstructions (bottom row).}
    \end{subfigure}%
    
    \vspace{2em}
    
    \begin{subfigure}{0.5\columnwidth}
        \centering
        \includegraphics[width=\columnwidth]{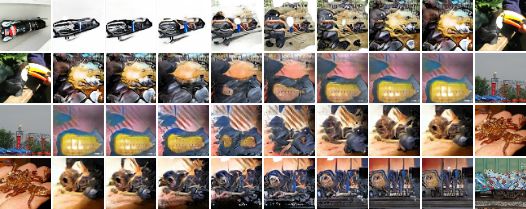}%
        \caption{Real sample interpolations.}
    \end{subfigure}%
    ~
    \begin{subfigure}{0.5\columnwidth}
        \centering
        \includegraphics[width=\columnwidth]{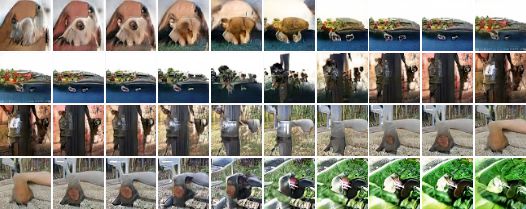}
        \caption{Random model sample interpolations.}
    \end{subfigure}%
    \caption{Qualitative results of training BiGAN on ImageNet. Interpolations are obtain by embedding two real/fake images $x_1$ and $x_2$ into the latent space with $E$, then decoding images uniformly along a line between these two embeddings $G\left(\alpha E(x_1) + (1-\alpha) E(x_2)\right)$. Left- and right-most images in each interpolation row are originals. \newline \newline
Statistics: $FID(P_X, P_G)=56.89$, \ $FID(P_X, P_{G(E(X))})=47.75$, \ Inception $L_2(P_X, P_{G(E(X))}) = 19.08$. \newline
Hyper-parameters: Adam learning rate 0.0001, $\beta_1=0.5$, $\beta_2=0.999$, 1 discriminator update(s) per generator update, WGAN-GP penalty weight $3.0$.
\newline Weighting of additional loss $\lambda=0.300000$.
} 
    \label{fig:qualitative_bigan_z_ae_imagenet}
\end{figure}


\begin{figure}
    \centering
    \textbf{Model: BiGAN + X adversarial loss. \  Dataset: Cifar10.}
    
    \vspace{2em}
    
    \begin{subfigure}{0.5\columnwidth}
        \centering
        \includegraphics[width=\columnwidth]{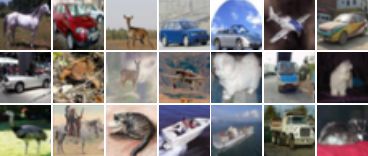}
        \caption{Real images.}
    \end{subfigure}%
    ~
    \begin{subfigure}{0.5\columnwidth}
        \centering
        \includegraphics[width=\columnwidth]{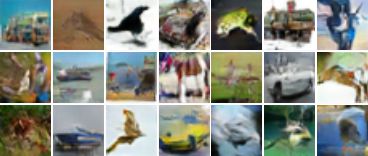}
        \caption{Random model samples.}
    \end{subfigure}%
    
    \vspace{2em}
    
    \begin{subfigure}{\columnwidth}
        \centering
        \includegraphics[width=\columnwidth]{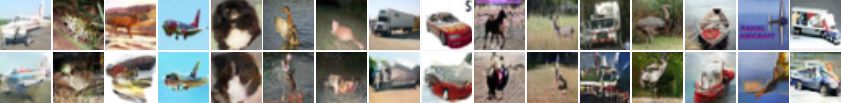}
        \caption{Real images (top row) and reconstructions (bottom row).}
    \end{subfigure}%
    
    \vspace{2em}
    
    \begin{subfigure}{\columnwidth}
        \centering
        \includegraphics[width=\columnwidth]{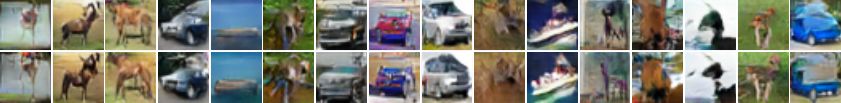}
        \caption{Random model samples (top row) and reconstructions (bottom row).}
    \end{subfigure}%
    
    \vspace{2em}
    
    \begin{subfigure}{0.5\columnwidth}
        \centering
        \includegraphics[width=\columnwidth]{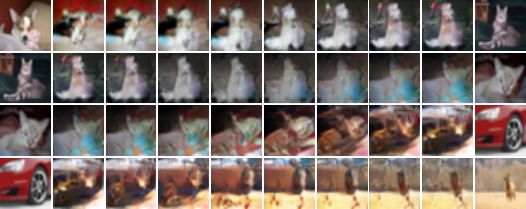}%
        \caption{Real sample interpolations.}
    \end{subfigure}%
    ~
    \begin{subfigure}{0.5\columnwidth}
        \centering
        \includegraphics[width=\columnwidth]{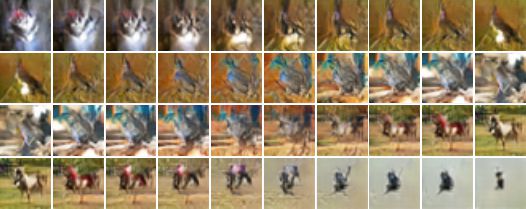}
        \caption{Random model sample interpolations.}
    \end{subfigure}%
    \caption{Qualitative results of training BiGAN on Cifar10. Interpolations are obtain by embedding two real/fake images $x_1$ and $x_2$ into the latent space with $E$, then decoding images uniformly along a line between these two embeddings $G\left(\alpha E(x_1) + (1-\alpha) E(x_2)\right)$. Left- and right-most images in each interpolation row are originals. \newline \newline
Statistics: $FID(P_X, P_G)=30.52$, \ $FID(P_X, P_{G(E(X))})=24.64$, \ Inception $L_2(P_X, P_{G(E(X))}) = 17.14$. \newline
Hyper-parameters: Adam learning rate 0.0001, $\beta_1=0.5$, $\beta_2=0.999$, 1 discriminator update(s) per generator update, WGAN-GP penalty weight $1.0$.
\newline Weighting of additional loss $\lambda=3$.
} 
    \label{fig:qualitative_bigan_x_adv_cifar10}
\end{figure}

\begin{figure}
    \centering
    \textbf{Model: BiGAN + X adversarial loss. \  Dataset: CelebA.}
    
    \vspace{2em}
    
    \begin{subfigure}{0.5\columnwidth}
        \centering
        \includegraphics[width=\columnwidth]{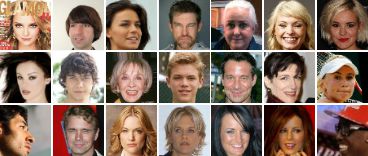}
        \caption{Real images.}
    \end{subfigure}%
    ~
    \begin{subfigure}{0.5\columnwidth}
        \centering
        \includegraphics[width=\columnwidth]{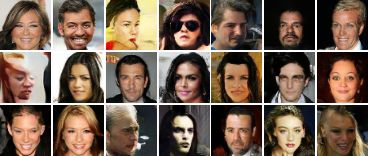}
        \caption{Random model samples.}
    \end{subfigure}%
    
    \vspace{2em}
    
    \begin{subfigure}{\columnwidth}
        \centering
        \includegraphics[width=\columnwidth]{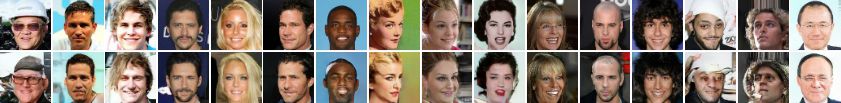}
        \caption{Real images (top row) and reconstructions (bottom row).}
    \end{subfigure}%
    
    \vspace{2em}
    
    \begin{subfigure}{\columnwidth}
        \centering
        \includegraphics[width=\columnwidth]{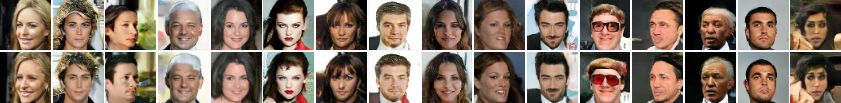}
        \caption{Random model samples (top row) and reconstructions (bottom row).}
    \end{subfigure}%
    
    \vspace{2em}
    
    \begin{subfigure}{0.5\columnwidth}
        \centering
        \includegraphics[width=\columnwidth]{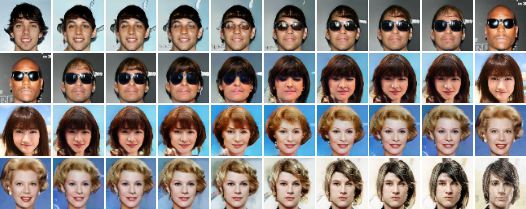}%
        \caption{Real sample interpolations.}
    \end{subfigure}%
    ~
    \begin{subfigure}{0.5\columnwidth}
        \centering
        \includegraphics[width=\columnwidth]{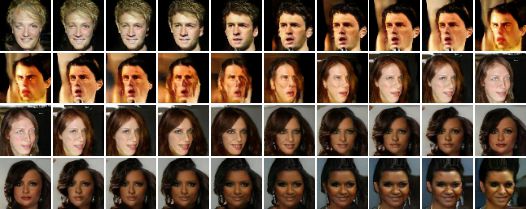}
        \caption{Random model sample interpolations.}
    \end{subfigure}%
    \caption{Qualitative results of training BiGAN on CelebA. Interpolations are obtain by embedding two real/fake images $x_1$ and $x_2$ into the latent space with $E$, then decoding images uniformly along a line between these two embeddings $G\left(\alpha E(x_1) + (1-\alpha) E(x_2)\right)$. Left- and right-most images in each interpolation row are originals. \newline \newline
Statistics: $FID(P_X, P_G)=8.60$, \ $FID(P_X, P_{G(E(X))})=7.08$, \ Inception $L_2(P_X, P_{G(E(X))}) = 10.98$. \newline
Hyper-parameters: Adam learning rate 0.0003, $\beta_1=0.5$, $\beta_2=0.999$, 2 discriminator update(s) per generator update, WGAN-GP penalty weight $3.0$.
\newline Weighting of additional loss $\lambda=0.300000$.
} 
    \label{fig:qualitative_bigan_x_adv_celeba}
\end{figure}

\begin{figure}
    \centering
    \textbf{Model: BiGAN + X adversarial loss. \  Dataset: Flintstones.}
    
    \vspace{2em}
    
    \begin{subfigure}{0.5\columnwidth}
        \centering
        \includegraphics[width=\columnwidth]{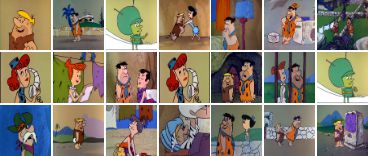}
        \caption{Real images.}
    \end{subfigure}%
    ~
    \begin{subfigure}{0.5\columnwidth}
        \centering
        \includegraphics[width=\columnwidth]{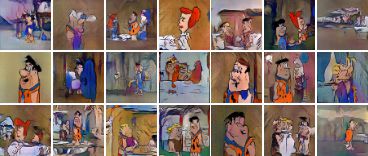}
        \caption{Random model samples.}
    \end{subfigure}%
    
    \vspace{2em}
    
    \begin{subfigure}{\columnwidth}
        \centering
        \includegraphics[width=\columnwidth]{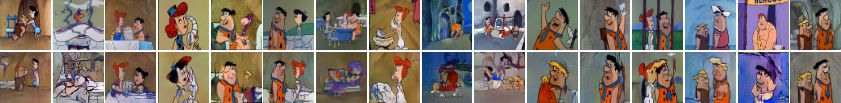}
        \caption{Real images (top row) and reconstructions (bottom row).}
    \end{subfigure}%
    
    \vspace{2em}
    
    \begin{subfigure}{\columnwidth}
        \centering
        \includegraphics[width=\columnwidth]{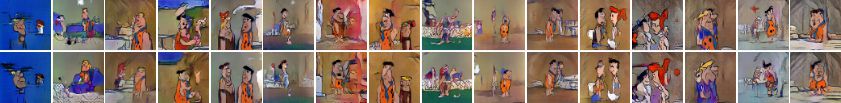}
        \caption{Random model samples (top row) and reconstructions (bottom row).}
    \end{subfigure}%
    
    \vspace{2em}
    
    \begin{subfigure}{0.5\columnwidth}
        \centering
        \includegraphics[width=\columnwidth]{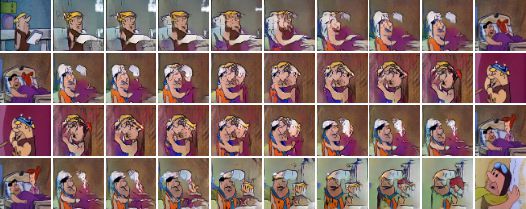}%
        \caption{Real sample interpolations.}
    \end{subfigure}%
    ~
    \begin{subfigure}{0.5\columnwidth}
        \centering
        \includegraphics[width=\columnwidth]{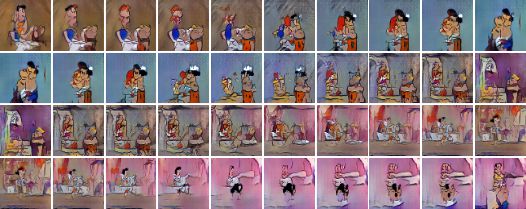}
        \caption{Random model sample interpolations.}
    \end{subfigure}%
    \caption{Qualitative results of training BiGAN on Flintstones. Interpolations are obtain by embedding two real/fake images $x_1$ and $x_2$ into the latent space with $E$, then decoding images uniformly along a line between these two embeddings $G\left(\alpha E(x_1) + (1-\alpha) E(x_2)\right)$. Left- and right-most images in each interpolation row are originals.\newline \newline
Statistics: $FID(P_X, P_G)=84.89$, \ $FID(P_X, P_{G(E(X))})=85.63$, \ Inception $L_2(P_X, P_{G(E(X))}) = 15.66$. \newline
Hyper-parameters: Adam learning rate 0.001, $\beta_1=0.5$, $\beta_2=0.999$, 2 discriminator update(s) per generator update, WGAN-GP penalty weight $10.0$.
\newline Weighting of additional loss $\lambda=10$.
} 
    \label{fig:qualitative_bigan_x_adv_flintstones}
\end{figure}

\begin{figure}
    \centering
    \textbf{Model: BiGAN + X adversarial loss. \  Dataset: ImageNet.}
    
    \vspace{2em}
    
    \begin{subfigure}{0.5\columnwidth}
        \centering
        \includegraphics[width=\columnwidth]{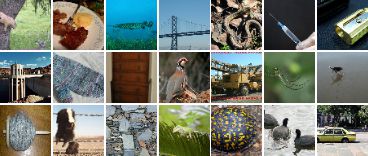}
        \caption{Real images.}
    \end{subfigure}%
    ~
    \begin{subfigure}{0.5\columnwidth}
        \centering
        \includegraphics[width=\columnwidth]{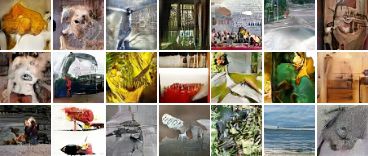}
        \caption{Random model samples.}
    \end{subfigure}%
    
    \vspace{2em}
    
    \begin{subfigure}{\columnwidth}
        \centering
        \includegraphics[width=\columnwidth]{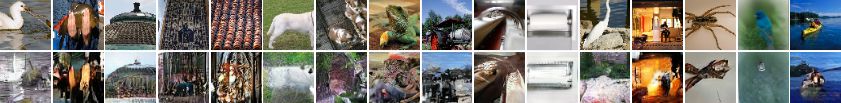}
        \caption{Real images (top row) and reconstructions (bottom row).}
    \end{subfigure}%
    
    \vspace{2em}
    
    \begin{subfigure}{\columnwidth}
        \centering
        \includegraphics[width=\columnwidth]{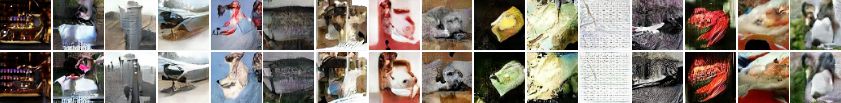}
        \caption{Random model samples (top row) and reconstructions (bottom row).}
    \end{subfigure}%
    
    \vspace{2em}
    
    \begin{subfigure}{0.5\columnwidth}
        \centering
        \includegraphics[width=\columnwidth]{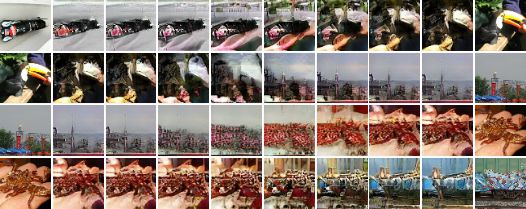}%
        \caption{Real sample interpolations.}
    \end{subfigure}%
    ~
    \begin{subfigure}{0.5\columnwidth}
        \centering
        \includegraphics[width=\columnwidth]{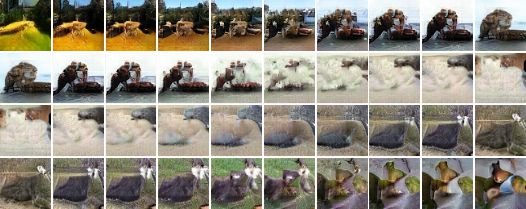}
        \caption{Random model sample interpolations.}
    \end{subfigure}%
    \caption{Qualitative results of training BiGAN on ImageNet. Interpolations are obtain by embedding two real/fake images $x_1$ and $x_2$ into the latent space with $E$, then decoding images uniformly along a line between these two embeddings $G\left(\alpha E(x_1) + (1-\alpha) E(x_2)\right)$. Left- and right-most images in each interpolation row are originals. \newline \newline
Statistics: $FID(P_X, P_G)=62.99$, \ $FID(P_X, P_{G(E(X))})=55.05$, \ Inception $L_2(P_X, P_{G(E(X))}) = 19.26$. \newline
Hyper-parameters: Adam learning rate 0.0003, $\beta_1=0.5$, $\beta_2=0.999$, 2 discriminator update(s) per generator update, WGAN-GP penalty weight $3.0$.
\newline Weighting of additional loss $\lambda=3$.
} 
    \label{fig:qualitative_bigan_x_adv_imagenet}
\end{figure}


\begin{figure}
    \centering
    \textbf{Model: BiGAN + Z adversarial loss. \  Dataset: Cifar10.}
    
    \vspace{2em}
    
    \begin{subfigure}{0.5\columnwidth}
        \centering
        \includegraphics[width=\columnwidth]{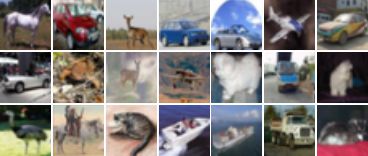}
        \caption{Real images.}
    \end{subfigure}%
    ~
    \begin{subfigure}{0.5\columnwidth}
        \centering
        \includegraphics[width=\columnwidth]{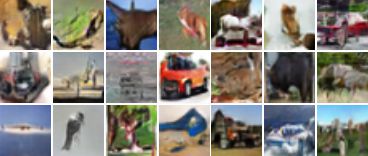}
        \caption{Random model samples.}
    \end{subfigure}%
    
    \vspace{2em}
    
    \begin{subfigure}{\columnwidth}
        \centering
        \includegraphics[width=\columnwidth]{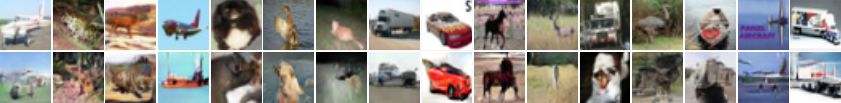}
        \caption{Real images (top row) and reconstructions (bottom row).}
    \end{subfigure}%
    
    \vspace{2em}
    
    \begin{subfigure}{\columnwidth}
        \centering
        \includegraphics[width=\columnwidth]{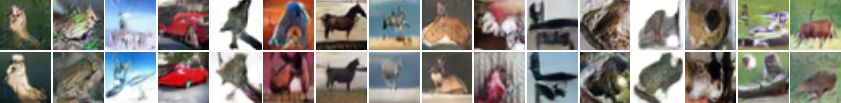}
        \caption{Random model samples (top row) and reconstructions (bottom row).}
    \end{subfigure}%
    
    \vspace{2em}
    
    \begin{subfigure}{0.5\columnwidth}
        \centering
        \includegraphics[width=\columnwidth]{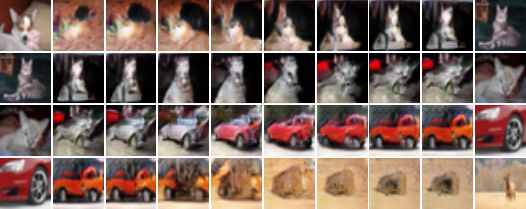}%
        \caption{Real sample interpolations.}
    \end{subfigure}%
    ~
    \begin{subfigure}{0.5\columnwidth}
        \centering
        \includegraphics[width=\columnwidth]{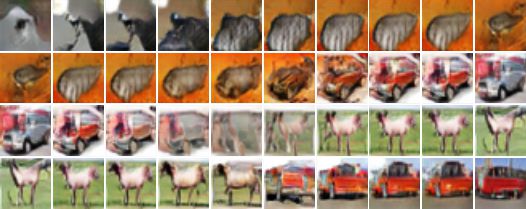}
        \caption{Random model sample interpolations.}
    \end{subfigure}%
    \caption{Qualitative results of training BiGAN on Cifar10. Interpolations are obtain by embedding two real/fake images $x_1$ and $x_2$ into the latent space with $E$, then decoding images uniformly along a line between these two embeddings $G\left(\alpha E(x_1) + (1-\alpha) E(x_2)\right)$. Left- and right-most images in each interpolation row are originals. \newline \newline
Statistics: $FID(P_X, P_G)=27.92$, \ $FID(P_X, P_{G(E(X))})=23.83$, \ Inception $L_2(P_X, P_{G(E(X))}) = 17.74$. \newline
Hyper-parameters: Adam learning rate 0.0001, $\beta_1=0.5$, $\beta_2=0.999$, 1 discriminator update(s) per generator update, WGAN-GP penalty weight $1.0$.
\newline Weighting of additional loss $\lambda=0.100000$.
} 
    \label{fig:qualitative_bigan_z_adv_cifar10}
\end{figure}

\begin{figure}
    \centering
    \textbf{Model: BiGAN + Z adversarial loss. \  Dataset: CelebA.}
    
    \vspace{2em}
    
    \begin{subfigure}{0.5\columnwidth}
        \centering
        \includegraphics[width=\columnwidth]{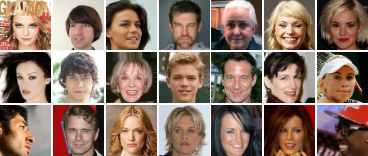}
        \caption{Real images.}
    \end{subfigure}%
    ~
    \begin{subfigure}{0.5\columnwidth}
        \centering
        \includegraphics[width=\columnwidth]{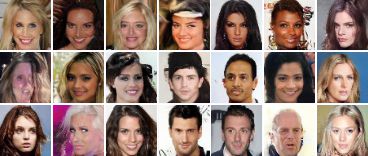}
        \caption{Random model samples.}
    \end{subfigure}%
    
    \vspace{2em}
    
    \begin{subfigure}{\columnwidth}
        \centering
        \includegraphics[width=\columnwidth]{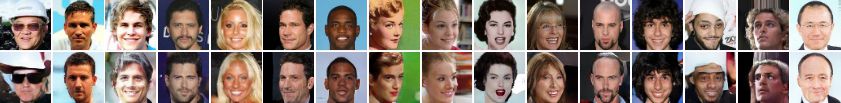}
        \caption{Real images (top row) and reconstructions (bottom row).}
    \end{subfigure}%
    
    \vspace{2em}
    
    \begin{subfigure}{\columnwidth}
        \centering
        \includegraphics[width=\columnwidth]{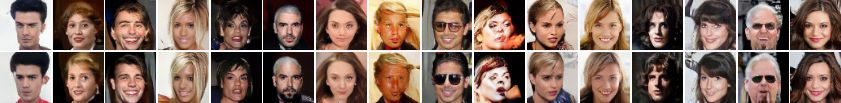}
        \caption{Random model samples (top row) and reconstructions (bottom row).}
    \end{subfigure}%
    
    \vspace{2em}
    
    \begin{subfigure}{0.5\columnwidth}
        \centering
        \includegraphics[width=\columnwidth]{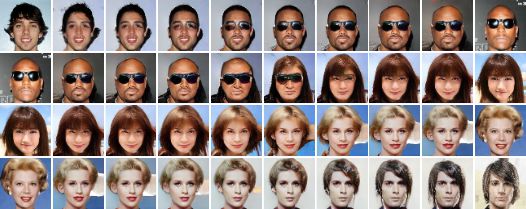}%
        \caption{Real sample interpolations.}
    \end{subfigure}%
    ~
    \begin{subfigure}{0.5\columnwidth}
        \centering
        \includegraphics[width=\columnwidth]{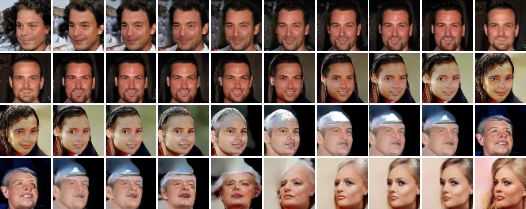}
        \caption{Random model sample interpolations.}
    \end{subfigure}%
    \caption{Qualitative results of training BiGAN on CelebA. Interpolations are obtain by embedding two real/fake images $x_1$ and $x_2$ into the latent space with $E$, then decoding images uniformly along a line between these two embeddings $G\left(\alpha E(x_1) + (1-\alpha) E(x_2)\right)$. Left- and right-most images in each interpolation row are originals. \newline \newline
Statistics: $FID(P_X, P_G)=7.92$, \ $FID(P_X, P_{G(E(X))})=7.16$, \ Inception $L_2(P_X, P_{G(E(X))}) = 11.04$. \newline
Hyper-parameters: Adam learning rate 0.0003, $\beta_1=0.5$, $\beta_2=0.999$, 1 discriminator update(s) per generator update, WGAN-GP penalty weight $10.0$.
\newline Weighting of additional loss $\lambda=3$.
} 
    \label{fig:qualitative_bigan_z_adv_celeba}
\end{figure}

\begin{figure}
    \centering
    \textbf{Model: BiGAN + Z adversarial loss. \  Dataset: Flintstones.}
    
    \vspace{2em}
    
    \begin{subfigure}{0.5\columnwidth}
        \centering
        \includegraphics[width=\columnwidth]{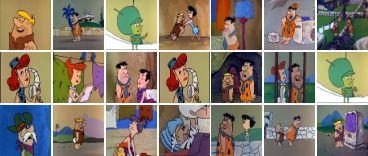}
        \caption{Real images.}
    \end{subfigure}%
    ~
    \begin{subfigure}{0.5\columnwidth}
        \centering
        \includegraphics[width=\columnwidth]{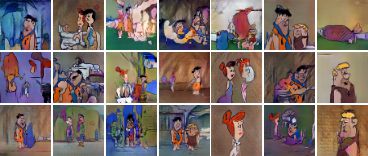}
        \caption{Random model samples.}
    \end{subfigure}%
    
    \vspace{2em}
    
    \begin{subfigure}{\columnwidth}
        \centering
        \includegraphics[width=\columnwidth]{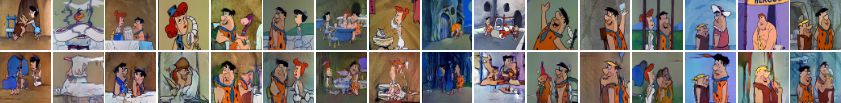}
        \caption{Real images (top row) and reconstructions (bottom row).}
    \end{subfigure}%
    
    \vspace{2em}
    
    \begin{subfigure}{\columnwidth}
        \centering
        \includegraphics[width=\columnwidth]{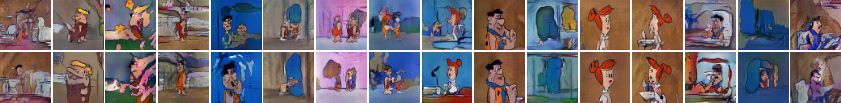}
        \caption{Random model samples (top row) and reconstructions (bottom row).}
    \end{subfigure}%
    
    \vspace{2em}
    
    \begin{subfigure}{0.5\columnwidth}
        \centering
        \includegraphics[width=\columnwidth]{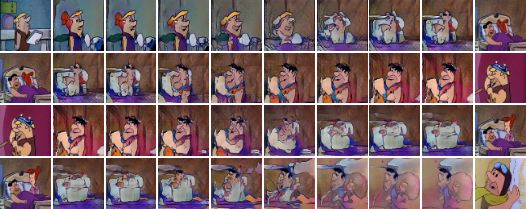}%
        \caption{Real sample interpolations.}
    \end{subfigure}%
    ~
    \begin{subfigure}{0.5\columnwidth}
        \centering
        \includegraphics[width=\columnwidth]{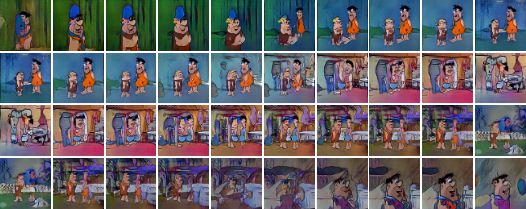}
        \caption{Random model sample interpolations.}
    \end{subfigure}%
    \caption{Qualitative results of training BiGAN on Flintstones. Interpolations are obtain by embedding two real/fake images $x_1$ and $x_2$ into the latent space with $E$, then decoding images uniformly along a line between these two embeddings $G\left(\alpha E(x_1) + (1-\alpha) E(x_2)\right)$. Left- and right-most images in each interpolation row are originals.\newline \newline
Statistics: $FID(P_X, P_G)=79.21$, \ $FID(P_X, P_{G(E(X))})=83.87$, \ Inception $L_2(P_X, P_{G(E(X))}) = 15.63$. \newline
Hyper-parameters: Adam learning rate 0.0001, $\beta_1=0.5$, $\beta_2=0.999$, 2 discriminator update(s) per generator update, WGAN-GP penalty weight $3.0$.
\newline Weighting of additional loss $\lambda=0.010000$.
} 
    \label{fig:qualitative_bigan_z_adv_flintstones}
\end{figure}

\begin{figure}
    \centering
    \textbf{Model: BiGAN + Z adversarial loss. \  Dataset: ImageNet.}
    
    \vspace{2em}
    
    \begin{subfigure}{0.5\columnwidth}
        \centering
        \includegraphics[width=\columnwidth]{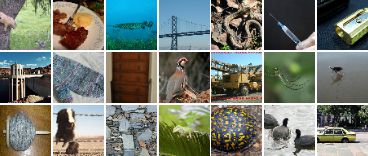}
        \caption{Real images.}
    \end{subfigure}%
    ~
    \begin{subfigure}{0.5\columnwidth}
        \centering
        \includegraphics[width=\columnwidth]{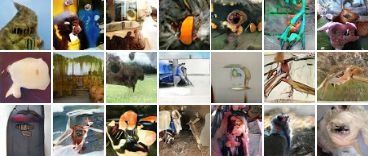}
        \caption{Random model samples.}
    \end{subfigure}%
    
    \vspace{2em}
    
    \begin{subfigure}{\columnwidth}
        \centering
        \includegraphics[width=\columnwidth]{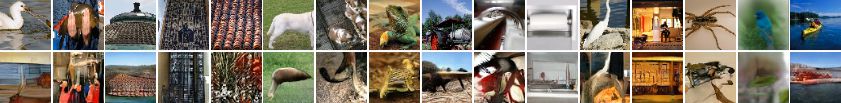}
        \caption{Real images (top row) and reconstructions (bottom row).}
    \end{subfigure}%
    
    \vspace{2em}
    
    \begin{subfigure}{\columnwidth}
        \centering
        \includegraphics[width=\columnwidth]{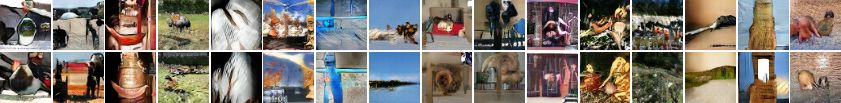}
        \caption{Random model samples (top row) and reconstructions (bottom row).}
    \end{subfigure}%
    
    \vspace{2em}
    
    \begin{subfigure}{0.5\columnwidth}
        \centering
        \includegraphics[width=\columnwidth]{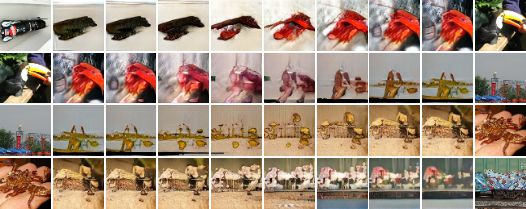}%
        \caption{Real sample interpolations.}
    \end{subfigure}%
    ~
    \begin{subfigure}{0.5\columnwidth}
        \centering
        \includegraphics[width=\columnwidth]{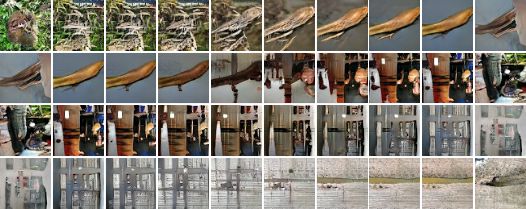}
        \caption{Random model sample interpolations.}
    \end{subfigure}%
    \caption{Qualitative results of training BiGAN on ImageNet. Interpolations are obtain by embedding two real/fake images $x_1$ and $x_2$ into the latent space with $E$, then decoding images uniformly along a line between these two embeddings $G\left(\alpha E(x_1) + (1-\alpha) E(x_2)\right)$. Left- and right-most images in each interpolation row are originals. \newline \newline
Statistics: $FID(P_X, P_G)=58.41$, \ $FID(P_X, P_{G(E(X))})=51.89$, \ Inception $L_2(P_X, P_{G(E(X))}) = 19.50$. \newline
Hyper-parameters: Adam learning rate 0.0001, $\beta_1=0.5$, $\beta_2=0.999$, 1 discriminator update(s) per generator update, WGAN-GP penalty weight $1.0$.
\newline Weighting of additional loss $\lambda=3$.
} 
    \label{fig:qualitative_bigan_z_adv_imagenet}
\end{figure}


\begin{figure}
    \centering
    \textbf{Model: VAE. \  Dataset: Cifar10.}
    
    \vspace{2em}
    
    \begin{subfigure}{0.5\columnwidth}
        \centering
        \includegraphics[width=\columnwidth]{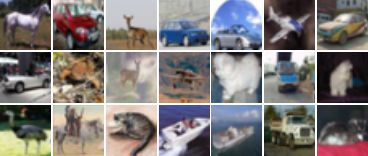}
        \caption{Real images.}
    \end{subfigure}%
    ~
    \begin{subfigure}{0.5\columnwidth}
        \centering
        \includegraphics[width=\columnwidth]{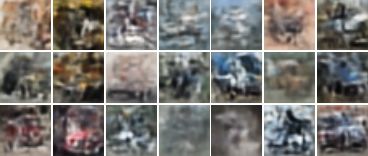}
        \caption{Random model samples.}
    \end{subfigure}%
    
    \vspace{2em}
    
    \begin{subfigure}{\columnwidth}
        \centering
        \includegraphics[width=\columnwidth]{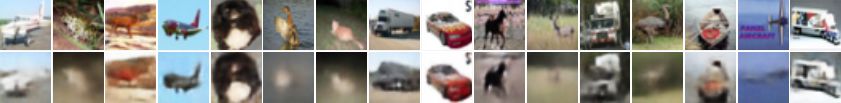}
        \caption{Real images (top row) and reconstructions (bottom row).}
    \end{subfigure}%
    
    \vspace{2em}
    
    \begin{subfigure}{\columnwidth}
        \centering
        \includegraphics[width=\columnwidth]{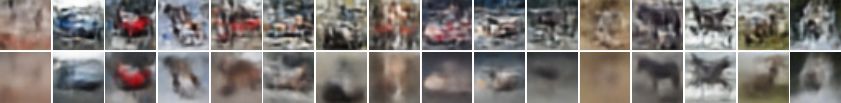}
        \caption{Random model samples (top row) and reconstructions (bottom row).}
    \end{subfigure}%
    
    \vspace{2em}
    
    \begin{subfigure}{0.5\columnwidth}
        \centering
        \includegraphics[width=\columnwidth]{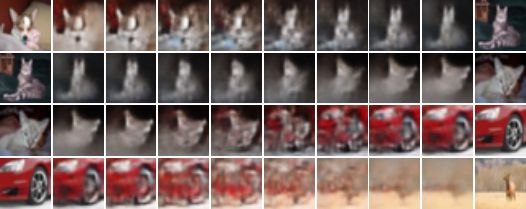}%
        \caption{Real sample interpolations.}
    \end{subfigure}%
    ~
    \begin{subfigure}{0.5\columnwidth}
        \centering
        \includegraphics[width=\columnwidth]{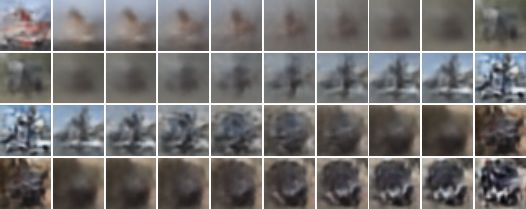}
        \caption{Random model sample interpolations.}
    \end{subfigure}%
    \caption{Qualitative results of training BiGAN on Cifar10. Interpolations are obtain by embedding two real/fake images $x_1$ and $x_2$ into the latent space with $E$, then decoding images uniformly along a line between these two embeddings $G\left(\alpha E(x_1) + (1-\alpha) E(x_2)\right)$. Left- and right-most images in each interpolation row are originals. \newline \newline
Statistics: $FID(P_X, P_G)=101.66$, \ $FID(P_X, P_{G(E(X))})=91.76$, \ Inception $L_2(P_X, P_{G(E(X))}) = 18.64$. \newline
Hyper-parameters: Adam learning rate 0.0003, $\beta_1=0.5$, $\beta_2=0.999$.
} 
    \label{fig:qualitative_vae_cifar10}
\end{figure}

\begin{figure}
    \centering
    \textbf{Model: VAE. \  Dataset: CelebA.}
    
    \vspace{2em}
    
    \begin{subfigure}{0.5\columnwidth}
        \centering
        \includegraphics[width=\columnwidth]{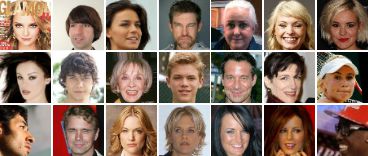}
        \caption{Real images.}
    \end{subfigure}%
    ~
    \begin{subfigure}{0.5\columnwidth}
        \centering
        \includegraphics[width=\columnwidth]{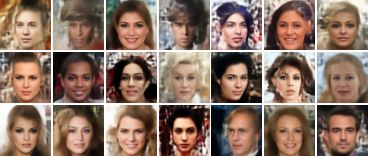}
        \caption{Random model samples.}
    \end{subfigure}%
    
    \vspace{2em}
    
    \begin{subfigure}{\columnwidth}
        \centering
        \includegraphics[width=\columnwidth]{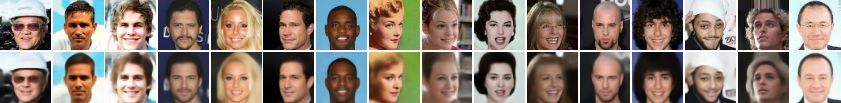}
        \caption{Real images (top row) and reconstructions (bottom row).}
    \end{subfigure}%
    
    \vspace{2em}
    
    \begin{subfigure}{\columnwidth}
        \centering
        \includegraphics[width=\columnwidth]{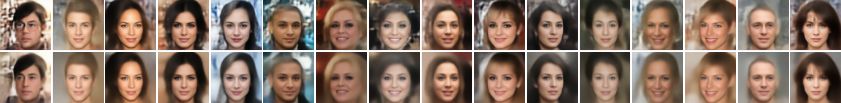}
        \caption{Random model samples (top row) and reconstructions (bottom row).}
    \end{subfigure}%
    
    \vspace{2em}
    
    \begin{subfigure}{0.5\columnwidth}
        \centering
        \includegraphics[width=\columnwidth]{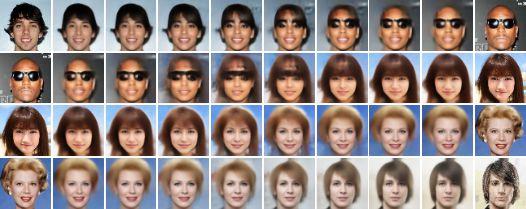}%
        \caption{Real sample interpolations.}
    \end{subfigure}%
    ~
    \begin{subfigure}{0.5\columnwidth}
        \centering
        \includegraphics[width=\columnwidth]{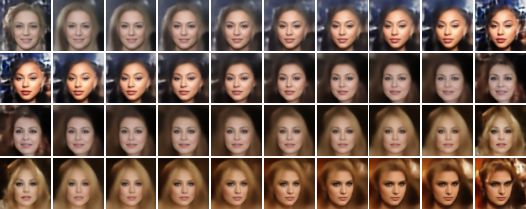}
        \caption{Random model sample interpolations.}
    \end{subfigure}%
    \caption{Qualitative results of training BiGAN on CelebA. Interpolations are obtain by embedding two real/fake images $x_1$ and $x_2$ into the latent space with $E$, then decoding images uniformly along a line between these two embeddings $G\left(\alpha E(x_1) + (1-\alpha) E(x_2)\right)$. Left- and right-most images in each interpolation row are originals. \newline \newline
Statistics: $FID(P_X, P_G)=53.49$, \ $FID(P_X, P_{G(E(X))})=43.30$, \ Inception $L_2(P_X, P_{G(E(X))}) = 11.70$. \newline
Hyper-parameters: Adam learning rate 0.0003, $\beta_1=0.5$, $\beta_2=0.999$.
} 
    \label{fig:qualitative_vae_celeba}
\end{figure}

\begin{figure}
    \centering
    \textbf{Model: VAE. \  Dataset: Flintstones.}
    
    \vspace{2em}
    
    \begin{subfigure}{0.5\columnwidth}
        \centering
        \includegraphics[width=\columnwidth]{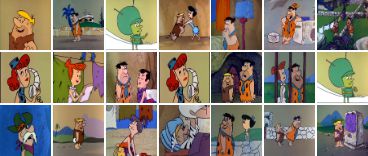}
        \caption{Real images.}
    \end{subfigure}%
    ~
    \begin{subfigure}{0.5\columnwidth}
        \centering
        \includegraphics[width=\columnwidth]{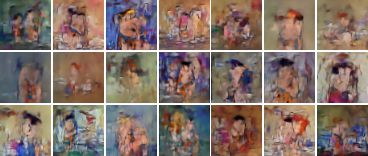}
        \caption{Random model samples.}
    \end{subfigure}%
    
    \vspace{2em}
    
    \begin{subfigure}{\columnwidth}
        \centering
        \includegraphics[width=\columnwidth]{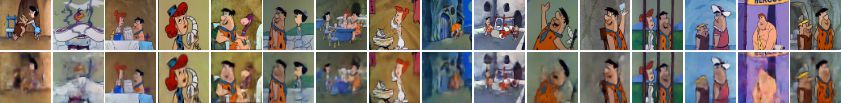}
        \caption{Real images (top row) and reconstructions (bottom row).}
    \end{subfigure}%
    
    \vspace{2em}
    
    \begin{subfigure}{\columnwidth}
        \centering
        \includegraphics[width=\columnwidth]{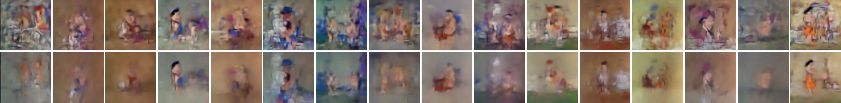}
        \caption{Random model samples (top row) and reconstructions (bottom row).}
    \end{subfigure}%
    
    \vspace{2em}
    
    \begin{subfigure}{0.5\columnwidth}
        \centering
        \includegraphics[width=\columnwidth]{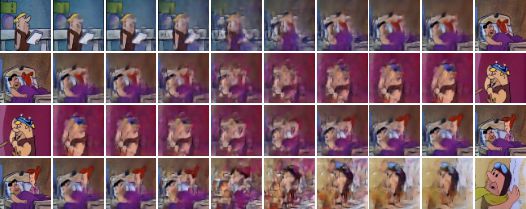}%
        \caption{Real sample interpolations.}
    \end{subfigure}%
    ~
    \begin{subfigure}{0.5\columnwidth}
        \centering
        \includegraphics[width=\columnwidth]{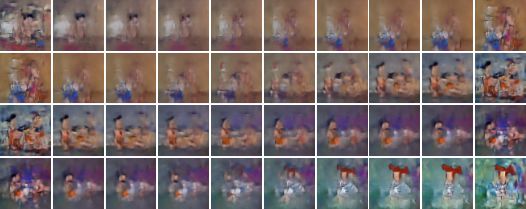}
        \caption{Random model sample interpolations.}
    \end{subfigure}%
    \caption{Qualitative results of training BiGAN on Flintstones. Interpolations are obtain by embedding two real/fake images $x_1$ and $x_2$ into the latent space with $E$, then decoding images uniformly along a line between these two embeddings $G\left(\alpha E(x_1) + (1-\alpha) E(x_2)\right)$. Left- and right-most images in each interpolation row are originals.\newline \newline
Statistics: $FID(P_X, P_G)=242.87$, \ $FID(P_X, P_{G(E(X))})=226.51$, \ Inception $L_2(P_X, P_{G(E(X))}) = 19.42$. \newline
Hyper-parameters: Adam learning rate 0.0001, $\beta_1=0.5$, $\beta_2=0.9900$.
} 
    \label{fig:qualitative_vae_flintstones}
\end{figure}

\begin{figure}
    \centering
    \textbf{Model: VAE. \  Dataset: ImageNet.}
    
    \vspace{2em}
    
    \begin{subfigure}{0.5\columnwidth}
        \centering
        \includegraphics[width=\columnwidth]{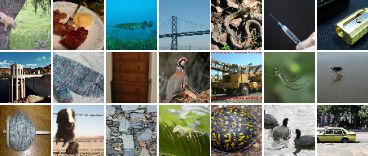}
        \caption{Real images.}
    \end{subfigure}%
    ~
    \begin{subfigure}{0.5\columnwidth}
        \centering
        \includegraphics[width=\columnwidth]{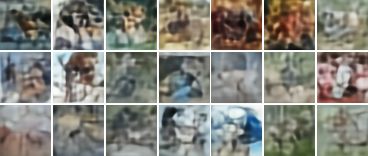}
        \caption{Random model samples.}
    \end{subfigure}%
    
    \vspace{2em}
    
    \begin{subfigure}{\columnwidth}
        \centering
        \includegraphics[width=\columnwidth]{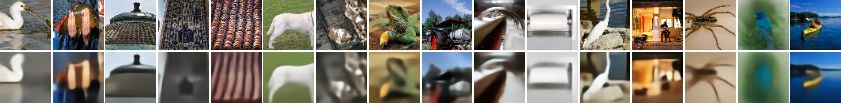}
        \caption{Real images (top row) and reconstructions (bottom row).}
    \end{subfigure}%
    
    \vspace{2em}
    
    \begin{subfigure}{\columnwidth}
        \centering
        \includegraphics[width=\columnwidth]{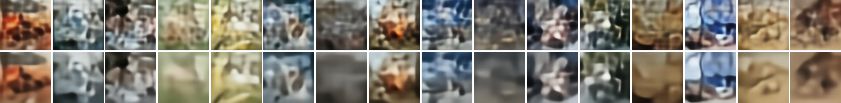}
        \caption{Random model samples (top row) and reconstructions (bottom row).}
    \end{subfigure}%
    
    \vspace{2em}
    
    \begin{subfigure}{0.5\columnwidth}
        \centering
        \includegraphics[width=\columnwidth]{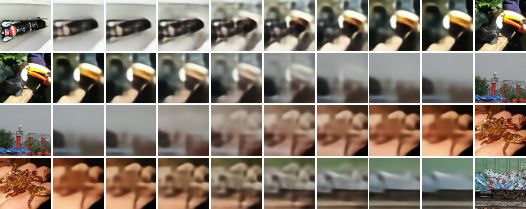}%
        \caption{Real sample interpolations.}
    \end{subfigure}%
    ~
    \begin{subfigure}{0.5\columnwidth}
        \centering
        \includegraphics[width=\columnwidth]{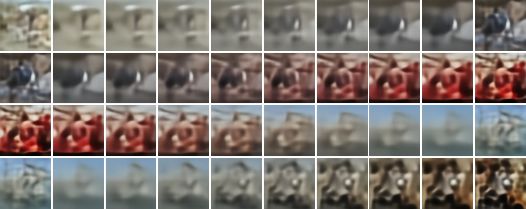}
        \caption{Random model sample interpolations.}
    \end{subfigure}%
    \caption{Qualitative results of training BiGAN on ImageNet. Interpolations are obtain by embedding two real/fake images $x_1$ and $x_2$ into the latent space with $E$, then decoding images uniformly along a line between these two embeddings $G\left(\alpha E(x_1) + (1-\alpha) E(x_2)\right)$. Left- and right-most images in each interpolation row are originals. \newline \newline
Statistics: $FID(P_X, P_G)=152.46$, \ $FID(P_X, P_{G(E(X))})=107.17$, \ Inception $L_2(P_X, P_{G(E(X))}) = 19.07$. \newline
Hyper-parameters: Adam learning rate 0.0003, $\beta_1=0.5$, $\beta_2=0.9900$.
} 
    \label{fig:qualitative_vae_imagenet}
\end{figure}